\documentclass{article}
\usepackage{caption} 
\usepackage{subcaption}

\usepackage{arxiv}
\usepackage[utf8]{inputenc} 
\usepackage[T1]{fontenc}    
\usepackage{hyperref}       
\usepackage{url}            
\usepackage{booktabs}       
\usepackage{amsfonts}       
\usepackage{amsmath}
\usepackage{nicefrac}       
\usepackage{blindtext}
\usepackage{titlesec}
\usepackage{microtype}      
\usepackage{fancyhdr}
\usepackage{lipsum}
\usepackage{graphicx}
\usepackage{enumitem}
\usepackage{soul}
\usepackage{listings}
\usepackage{multirow} 
\usepackage[table,xcdraw]{xcolor}
\usepackage{float}
\usepackage[none]{hyphenat}
\usepackage{fancyvrb,fvextra}
\usepackage{csquotes}

\graphicspath{ {./images/} }
\usepackage{authblk}
\author[1]{Pawitsapak Akarajaradwong$^\dag$}
\author[1]{Pirat Pothavorn$^\dag$}
\author[1]{Chompakorn Chaksangchaichot$^\dag$}
\author[2]{Panuthep Tasawong}
\author[1]{Thitiwat Nopparatbundit}
\author[3]{Keerakiat Pratai}
\author[2]{Sarana Nutanong}

\affil[1]{VISAI AI, Thailand \authorcr
  \texttt{\{pawitsapaka\_visai,\,piratp\_visai,thitiwatn\_visai,chompakornc\_pro\}@vistec.ac.th}}
\affil[2]{Vidyasirimedhi Institute of Science and Technology, Thailand \authorcr
  \texttt{\{panuthep.t\_s20,\,sarana.n\}@vistec.ac.th}}
\affil[3]{Faculty of Law, Thammasat University, Thailand \authorcr \texttt{pratai@tu.ac.th}}
\affil[$\dag$]{These authors contributed equally as co-first authors.}

\title{%
NitiBench: A Comprehensive Study of LLM Framework Capabilities for Thai Legal Question Answering
}

\makeatletter
\renewcommand{\maketitle}{%
  \thispagestyle{fancy}%
  \vspace*{0.5em}
  \begin{center}
    \begin{tabular}{@{}m{1.8cm} m{0.75\textwidth}@{}}    
      \raggedright
      \includegraphics[width=1.5cm]{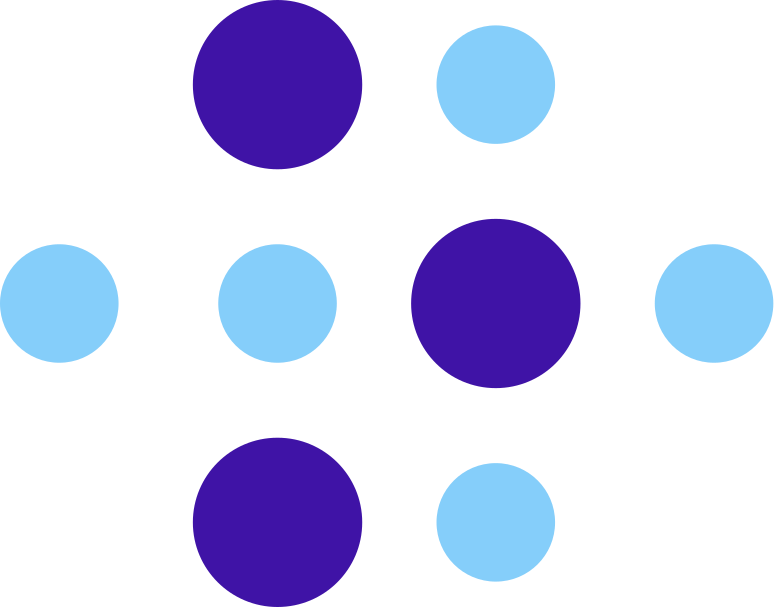} 
      & 
      {\bfseries \Large \@title}
      \\[4.0em]  

      \multicolumn{2}{@{}c@{}}{%
        \begin{minipage}{0.85\textwidth}
          \centering
          \@author \\[2.0em]
          \@date
        \end{minipage}
      } \\[1.0em]
    \end{tabular}
  \end{center}
  \vspace*{2em}
}
\makeatother

\begin{document}
\maketitle

\begin{abstract}
The application of large language models (LLMs) in the legal domain holds significant potential for information retrieval and question answering, yet Thai legal QA systems face challenges due to a lack of standardized evaluation benchmarks and the complexity of Thai legal structures. This paper introduces NitiBench, a benchmark comprising two datasets: the NitiBench-CCL, covering general Thai financial law, and the NitiBench-Tax, which includes real-world tax law cases requiring advanced legal reasoning. We evaluate retrieval-augmented generation (RAG) and long-context LLM-based approaches to address three key research questions: the impact of domain-specific components like section-based chunking and cross-referencing, the comparative performance of different retrievers and LLMs, and the viability of long-context LLMs as an alternative to RAG. Our results show that section-based chunking significantly improves retrieval and end-to-end performance, current retrievers struggle with complex queries, and long-context LLMs still underperform RAG-based systems in Thai legal QA. To support fair evaluation, we propose tailored multi-label retrieval metrics and the use of an LLM-as-judge for coverage and contradiction detection method. These findings highlight the limitations of current Thai legal NLP solutions and provide a foundation for future research in the field. We also open-sourced our codes and dataset to available publicly\footnote{\href{https://huggingface.co/datasets/VISAI-AI/nitibench}{https://huggingface.co/datasets/VISAI-AI/nitibench}}\footnote{\href{https://github.com/vistec-ai/nitibench}{https://github.com/vistec-ai/nitibench}}
\end{abstract}

\definecolor{darkgreen}{RGB}{10,100,10}
\definecolor{lightgray}{gray}{0.9}

\newpage
\tableofcontents
\newpage
\section{Introduction}
\label{sec:introduction}

The application of large language models (LLMs) to legal domains holds immense potential, particularly in information retrieval and question-answering.
One prominent application of LLMs in the legal field is the development of LLM-powered legal research tools. 
These tools assist legal professionals with tasks such as a conversational search of legal documents, document summarization, and citation analysis. 
For example, leveraging the Retrieval-Augmented Generation (RAG) framework, \cite{lexisnexis,thomsonreutersAIpoweredLegal} created a reliable legal assistant chatbot, while \cite{leewayhertzAgentsLegal} proposed agentic workflows to address various legal challenges within a unified framework.

Another valuable LLM application is legal QA systems, which, unlike research tools designed for legal professionals, cater to the general public. 
These systems handle tasks such as providing general legal information and consulting on specific cases. 
To achieve this effectively, this type of system, mostly implemented in the RAG framework, must understand user queries accurately, retrieve or identify relevant legal documents, apply the identified information to the query, and respond in a factually correct and contextually relevant manner. 
Examples of LLM-based legal consulting systems include AI Lawyer \cite{ailawyerLawyerYour} and AskLegal.bot \cite{asklegalAskLegalbotLegal}, which offer LLM-powered legal guidance to consumers.

Despite the proliferation of legal QA systems in resource-rich languages like English, extending and adapting these systems to resource-constrained languages, such as Thai, remains a challenge. 
Thai Legal QA systems require careful design choices to balance trade-offs within the RAG framework and beyond. One example of such a system is Thanoy~\footnote{\href{https://iapp.co.th/thanoy}{https://iapp.co.th/thanoy}}~\cite{thanoy}, a Thai RAG-based legal QA system designed for the general public. 
However, even as a rare marketable example, Thanoy exhibits various errors, including incorrect legislation retrieval and hallucinations.


To shed light on legal QA systems under Thai laws, we investigate the main challenges of implementing such a system. 
One of the most significant challenges in implementing Thai Legal QA systems is the lack of standardized evaluation processes. 
This makes it difficult to compare systems objectively and to identify general performance gaps or bottlenecks. 
The difficulty arises for two primary reasons:  

\begin{itemize}
    \item \textbf{Limited Thai Legal QA Corpora}: The specificity and complexity of the Thai legal domain require substantial human effort to create annotated datasets, which are currently scarce.  
    
    \item \textbf{Inadequate Evaluation Metrics}: Widely used retrieval metrics, such as hit rate, mean reciprocal rank (MRR), and recall, are suitable for single-label retrieval tasks. 
    However, the legal QA task oftentimes requires retrieving and applying multiple relevant documents simultaneously, making these traditional metrics insufficient. 
    Additionally, evaluating end-to-end (E2E) performance poses a challenge due to the multi-faceted nature of evaluating LLM-based systems, as existing studies often use varied tasks and metrics, leading to inconsistencies and making systems difficult to compare comprehensively.
\end{itemize}

To address this challenge, we present a novel benchmark dataset, NitiBench, along with corresponding task and evaluation metrics, specifically designed for assessing Thai Legal QA systems. 
The key features of this benchmark are as follows:  

\begin{itemize}  
    \item The benchmark comprises two datasets: the NitiBench-CCL Dataset, which covers general QA across 21 Thai financial law codes, and the Tax Case Dataset, which focuses on specialized QA involving real-world cases related to Thai tax issues requiring intense legal reasoning.  
    
    \item Each query in both datasets includes a question, an answer, and the relevant documents required to answer the question, enabling detailed evaluation of both the retrieval and E2E aspects.  
    
    \item The proposed retrieval metrics are designed to handle multi-label retrieval tasks, a critical requirement for effective legal QA.  
    
    \item Achieving a high E2E score on this benchmark requires a system to generate correct answers consistent with ground truth, avoid contradictions, and provide appropriate legal citations. 
    These tasks and metrics simulate the legal research process, where accurate citations are as important as the correctness of the answer.  
\end{itemize}

Using the proposed benchmark, we conduct a comprehensive evaluation of legal QA systems built with current industry-standard components within the RAG framework. 
Specifically, we aim to answer three key research questions:

\textbf{(RQ1) What impact do components tailored to the Thai legal system structure have on a Thai legal QA system?}  

A significant challenge in building RAG-based Thai legal QA systems lies in the inherent hierarchical structure of Thai law, where sections\footnote{In this paper, ``section'' refers to a component in legislation, while we use ``\S'' to denote a section, subsection, or subsubsection in this document. For more information on Thai legal terminology, see \S\ref{subsec: thai_legal}.} often reference other sections within the same or different legal codes. 
To address this, we propose a \textbf{hierarchy-aware chunking}, a strategy that segments documents by legal sections, and \textbf{NitiLink}, a component that augments retrieved sections with additional referenced sections.  

Using our benchmark, we demonstrate that this chunking strategy outperforms naive line chunking strategy in both retrieval and E2E metrics, highlighting the importance of incorporating domain-specific knowledge when choosing chunking strategies. 
However, incorporating NitiLink does not improve E2E performance, likely due to the simplistic referencing criteria used in its implementation.

\textbf{(RQ2) How do the choices of retriever and LLM affect performance in a RAG-based Thai legal QA system?}  

This research question explores the impact of different retriever and LLM choices on the performance of a Thai legal QA system. 
Additionally, we examine the current performance gap for both retrievers and LLMs, comparing the RAG-based system with parametric baselines and RAG with a provided golden context.  

Our experiments reveal that all retrievers struggle with the Tax Case dataset, which contains complex queries requiring nuanced understanding. 
Moreover, fine-tuned retrievers perform poorly on out-of-domain queries and do not consistently show improved performance even on in-domain queries. 
For the LLM component, the Claude 3.5 Sonnet model \cite{claude3.5sonnet} achieves the best E2E performance, albeit by a small margin. 
However, all tested LLMs fail to perform well on the Tax Case dataset, even under the RAG with golden context setting, indicating that current LLMs lack sufficient legal reasoning capabilities.  

\textbf{(RQ3) How does a Long-Context LLM-based Thai legal QA system perform compared to RAG-based systems?}  

The advent of Long-Context LLMs (LCLMs) raises questions about their feasibility and effectiveness in the Thai legal domain, potentially challenging the need for traditional RAG systems. 
To investigate this, we evaluate a Thai legal QA system using LCLMs and compare its performance with RAG-based systems on our benchmark. 
Our results show that LCLMs still significantly underperform RAG-based systems, suggesting that LCLMs are not yet viable for implementing effective Thai legal QA systems.
This result contradicted previous findings that LCLM is better than RAG in some tasks \cite{laban2024summaryhaystackchallengelongcontext,selfroute,lee2024longcontextlanguagemodelssubsume} where LCLM is dominant in most setups except for SQL query \cite{lee2024longcontextlanguagemodelssubsume}.

Finally, we summarize our main contributions as follows:
\begin{enumerate}
    \item {We construct a new benchmark corpus for evaluating Thai legal QA systems, consisting of two datasets focused on the Thai financial and tax law domains.}
    
    \item {We design an evaluation process accompanying the constructed dataset, which includes defining the task for legal QA systems, adapting retriever metrics for multi-label scenarios, and establishing concrete end-to-end evaluation metrics. 
    Our goal is to ensure that this evaluation process remains dataset-independent and can be applied to other Legal QA benchmark corpora.}
    
    \item {Using the proposed benchmark, we present our findings addressing the three key research questions outlined above.}
\end{enumerate}

\section{Literature Review}
\label{sec:lit_review}

In this section, we review prior work on benchmarking QA systems, the application of RAG in the legal domain, and efforts to implement Long-Context LLMs (LCLMs) in QA systems. 
Additionally, given the unique characteristics of the Thai legal domain, we outline the structure of the Thai legal system to highlight the specific challenges associated with developing a Thai legal QA system.

\subsection{Benchmarking}
\label{subsec: benchmarking}

Benchmarking plays a critical role in developing legal QA systems by providing standardized tasks and metrics to evaluate performance. 
It facilitates the comparison of different frameworks under consistent conditions and helps identify performance bottlenecks.

Popular benchmarks for legal QA in English include LexGlue~\cite{lexglue}, a collection of datasets for tasks such as court opinion classification and law violation identification. 
Similarly, LegalBench~\cite{legalbench} contains tasks for evaluating legal reasoning in large language models (LLMs), including contract natural language inference (NLI) and determining whether a statement constitutes hearsay.

Dahl et al.~\cite{Dahl_2024} introduce a Hallucination QA dataset to quantify hallucinations in LLMs for legal knowledge. 
Tasks include reference-based questions (e.g., verifying case existence) and reference-free questions (e.g., identifying the central holding of a case). 
The hallucination rate is measured using binary classification metrics for reference-based tasks and contradictions in answers across various temperatures for reference-free tasks.

Although these works provide structured evaluations, many focus on specific sub-tasks rather than free-form QA, the primary goal of legal QA systems. 
Notable efforts for free-form legal QA evaluation include Magesh et al.\cite{magesh2024hallucinationfreeassessingreliabilityleading}, which uses LLM-as-a-judge metrics to assess the correctness and groundedness of responses in commercial AI legal research tools. 
Similarly, RAGAS~\cite{es2023ragasautomatedevaluationretrieval} offers a reference-free retrieval-augmented generation (RAG) evaluation framework, measuring Faithfulness, Answer Relevance, and Context Relevance. 
Laban et al.~\cite{laban2024summaryhaystackchallengelongcontext} propose a framework where LLMs identify relevant documents from context—a critical task in RAG-based QA systems requiring models to digest retrieved information.

We conclude that a comprehensive legal QA benchmark should prioritize free-form QA tasks with clearly defined metrics, including:

\begin{itemize}
    \item{Correctness: The accuracy and relevance of the response.}
    \item{Contradiction: A measure of hallucination, defined by answers contradicting references.}
    \item{LLM Citation Quality: The ability of an LLM to filter and select relevant contexts from retrieved information.}
\end{itemize}

CLERC~\cite{clerc} addresses the lack of publicly available legal datasets by constructing a pipeline built on the Caselaw Access Project (CAP)\footnote{\url{https://case.law/}} corpus for RAG evaluations. 
It introduces realistic tasks, novel queries, and tailored metrics like BLEU, ROUGE, Citation Recall, Precision, and False Positive Rate.
Their work serves as a model for building datasets in other languages that also lack the publicly available legal QA benchmark, such as Thai.

In addition to end-to-end evaluations, assessing the retriever component of RAG-based systems is crucial for identifying system gaps. 
Traditional information retrieval metrics (Precision, Recall, F1-score) fall short in multi-label classifications required by Thai law, where multiple relevant legal sections must be retrieved. 
No prior work has developed multi-label variants of these metrics for Thai legal QA, emphasizing the need for tailored benchmarks in this context. 
Hence, a robust legal QA benchmark should also evaluate retriever performance for multi-label legal texts.

\subsection{Thai Legal System}
\label{subsec: thai_legal}

Thailand's legal framework operates within a democratic, constitutional monarchy, adhering to a hierarchical structure. 
Lower-hierarchical laws must not be contradictory to the higher ones~\cite{chuathai2023introduction}. 
This hierarchy encompasses seven distinct levels, namely: (1) the Constitution, (2) Organic Laws, (3) Acts and Codes, (4) Emergency Decrees, (5) Royal Decrees, (6) Ministerial Regulations, and (7) Local Ordinances.

Within the Thai legal hierarchy, \textbf{Acts} holds a pivotal position, ranking immediately after the Constitution and Organic Laws. 
Acts represent primary legislation passed through a rigorous legislative process, ultimately requiring Royal Assent for enactment. 
While most laws are enacted as individual Acts, some areas of law are governed by comprehensive Codes, such as the \textit{Civil and Commercial Code} or the \textit{Criminal Code}, which are structured compilations of related legal provisions.

Acts and Codes follow a multi-tiered structure, organized into levels including Book, Title, Chapter, Division, Section, Subsection, and Clause. 
Among these, \textbf{Section} level is the most fundamental unit, with each section articulating a specific legal rule or principle.

This granular structure serves a practical purpose, enabling legal professionals to efficiently navigate extensive codes, locate specific provisions, and understand their context within the broader legal framework. 
However, this complexity also raises challenges for the RAG framework. 
Specifically, it prompts the question of how to segment legislative documents in a manner that preserves the hierarchical structure while ensuring that each segment contains a meaningful and complete unit of the law.

Another notable characteristic of Thai law is the frequent use of inter-section references within legal texts. 
Sections often refer to other sections within the same legislation or in different legal documents.

Consider the following excerpt from the Criminal Code:
\begin{quote}
    \textit{Section 260: Whoever uses, sells, offers for sale, exchanges, or offers to exchange a ticket arising from an act as provided for in Section 258 or Section 259 shall be punished with imprisonment...}
    \label{quote: inter_ref}
\end{quote}

In this example, understanding Section 260 requires comprehension of the content of Sections 258 and 259, which are not provided within the immediate text. 
This inter-section referencing presents a significant challenge for traditional RAG systems, raising questions about how to handle referenced sections. 
Specifically, the design decision must consider whether these referenced sections can be automatically retrieved and augmented into the primary retrieved content to ensure a comprehensive understanding of the law.

From this understanding of the unique characteristics of the Thai legal system, we can see how these features contribute to the challenges of implementing LLMs in this domain. 
Specifically, it highlights the need to consider incorporating structural knowledge when segmenting legislation into chunks within the RAG framework, as this could help preserve the original content and enhance comprehensibility for retriever models. 
Moreover, given that Thai legislation frequently contains cross-references between sections, it raises an important question: how significant would the impact be of including nested sections as additional context for LLMs?

\subsection{RAG in legal practice}

{
RAG \cite{originalRAG} is an approach designed to enhance the performance of a large language model (LLM) by using a separate retrieval model to fetch relevant documents, which are then used as a context, providing the necessary knowledge for the LLM to answer the question.

This section highlights several works applying RAG in legal practice.
}

\label{sec:raginlegal}
CBR-RAG~\cite{CBR-RAG} is a framework that integrates Case-Based Reasoning (CBR) into RAG for answering legal questions.
CBR-RAG augments the original LLM query with contextually relevant cases retrieved using CBR’s indexing vocabulary and similarity knowledge containers, creating a richer prompt for the LLM. 
The authors evaluate CBR-RAG with different embeddings (BERT~\cite{bert}, LegalBERT~\cite{legalbert}, AnglEBERT~\cite{anglebert}) and retrieval methods (intra, inter, hybrid) on the legal question-answering task. 
Their results demonstrate that CBR-RAG significantly improves the quality of generated answers, particularly with AnglEBERT and hybrid embedding retrieval, highlighting the effectiveness of this approach for knowledge-intensive tasks.

Ajmi~\cite{Ajmi2024} offers a compelling exploration of how AI can bridge the "justice gap" by providing accessible legal information to those who lack traditional legal assistance. 
Their paper highlights the limitations of LLMs and advocates for the implementation of RAG to enhance accuracy and effectiveness. 
Drawing from a real-world implementation in the Nevada court system, this work demonstrates the feasibility and potential impact of RAG-assisted chatbots while maintaining a balanced perspective by acknowledging the limitations of AI and emphasizing the continuing need for human legal professionals.

Several commercial solutions also use RAG to power their legal assistant product.
For example, Lexis+ AI~\cite{lexisnexis} and Westlaw~\cite{thomsonreutersAIpoweredLegal} use RAG to make sure that the assistant response contains a properly cited legal document. 
Some localized legal-based assistant solutions, like Thanoy~\cite{thanoy}, also use RAG on Thai legal documents to create a chatbot. 
{Even though some of the aforementioned products claimed to mitigate hallucination problems via RAG, \cite{magesh2024hallucinationfreeassessingreliabilityleading} showed that hallucination still persists in some challenging cases.}

\subsection{RAG vs Long-Context LLMs}

{
Apart from using RAG to mitigate LLM hallucination by retrieving relevant documents as context, Long Context LLMs (LCLMs) have been an alternative to this problem as well~\cite{laban2024summaryhaystackchallengelongcontext, loft}. 
Since RAG contains multiple components, the end-to-end performance relies on multiple design choices such as the embedding model used, the document chunking method, the number of documents to be retrieved, and so forth; the question arises as to whether we can solve this tradeoff among these choices. 
Long context is proposed as an alternative to RAG, which removes the necessity for tools such as a retrieval model or reranking model where the whole documents are parsed as a context to the prompt without any need for any retrieval module.
Google Gemini 1.5 \cite{gemini1.5} was the first model that introduced a very long context length support of up to 1M tokens in the flash version and 2M tokens in the pro version. 

}

{
Recent works have tried to conduct a qualitative study about the tradeoff between LCLM and RAG. 
\cite{needleinahaystack} introduced a pressure test of the long context by inserting the context (needle) inside a very long document and then asking the model to answer the question based on the inserted context. 
However, this work is still limited to evaluating LCLM solely under naive assumptions. 
LongBench~\cite{longbench} and L-eval~\cite{leval} provide a more sophisticated long context benchmark compared to Needle-in-the-haystack. 
LOFT~\cite{loft}, Self-route~\cite{selfroute}, NEPAQuAD1.0~\cite{nepaquad}, and Summary of a haystack~\cite{laban2024summaryhaystackchallengelongcontext} measures LCLMs performance compared to RAG as a baseline. 
Findings from these works suggested that although many proprietary models such as GPT4o~\cite{gpt4o} and Claude 3.5 Sonnet~\cite{claude3.5sonnet} are able to process up to over 128K and 200K context length respectively, under long context setup, the performance still can't match that of Gemini-1.5-pro~\cite{leval, selfroute, laban2024summaryhaystackchallengelongcontext}. 
\cite{loft} also found that SQL reasoning is the only task RAG wins LCLM, and \cite{laban2024summaryhaystackchallengelongcontext} points out that the LCLMs' performance is still subpar to humans in terms of both citation correctness and answer coverage. 
}
%
Although the literature extensively explores the comparative performance of RAG and LCLMs across various domains, a crucial area remains unexplored: their application and evaluation within the legal domain. 
Numerous studies have meticulously investigated the strengths and weaknesses of RAG and LCLMs in handling complex tasks like question answering and reasoning. 
However, these investigations typically focus on general-purpose datasets and benchmarks. 
To date, no research has directly compared the efficacy of RAG and LCLMs in tackling the unique challenges inherent to the legal field. 
This gap underscores the need for future research to delve into the comparative performance of these paradigms, specifically within the legal domain, ultimately guiding the development of more effective and specialized legal NLP applications.
\section{Methodology}

In this section, we present NitiBench, a novel benchmark dataset specifically designed to evaluate Thai legal QA systems. 
We also outline an accompanying evaluation process, starting with a description of the end-to-end (E2E) task that the system must perform. 
Additionally, we introduce a set of carefully designed metrics to ensure a thorough and comprehensive assessment of the system's performance.

\subsection{Dataset}
\label{subsec: dataset}

As stated in \S\ref{subsec: benchmarking}, a good benchmark dataset should aim to directly measure a legal QA system's performance in a free-form QA task, which such systems are designed to excel at. 
Furthermore, it should explicitly include relevant legal documents to allow for the isolated evaluation of the retrieval component in a RAG-based legal QA system.
Therefore, we construct our datasets such that each entry consists of a query, corresponding relevant legal sections, and the answer. 
Formally, our dataset is represented as $\mathcal{D} = \{(\textbf{x}_1, y_1), ..., (\textbf{x}_N, y_N)\}$ where $\textbf{x}_i = (q, T)$. $q$ denotes a query or a question relevant to Thai law, $T = \{t_1, t_2, ..., t_K\}$ denotes positive documents relevant to $q$, and $y$ is an answer to the query based on the provided positives $T$.


We develop two datasets for assessing Thai legal QA systems: 
\begin{enumerate}
    \item {\textbf{NitiBench-CCL}}: This dataset represents general Thai Financial Law.
    This test set was derived from WangchanX-Legal-ThaiCCL-RAG's test set where we apply additional postprocessing to the original dataset.
    \item {\textbf{NitiBench-Tax}}: A specialized QA dataset for Thai Tax Law based on official government statement from \url{https://rd.go.th/}{Revenue Department website}.
\end{enumerate}

We also present \textbf{WangchanX Legal ThaiCCL RAG dataset} where its train split was used to finetune our retrieval model.

\subsubsection{WangchanX Legal ThaiCCL RAG Dataset}
\label{subsubsec: wcx_dataset}

We present \textbf{WangchanX Legal ThaiCCL RAG Dataset}\footnote{\url{https://huggingface.co/datasets/airesearch/WangchanX-Legal-ThaiCCL-RAG}} (WCX Dataset), a question-answering dataset specifically designed for RAG tasks in the field of Thai financial law containing 35 legislations. 
The goal of this dataset is to provide comprehensive datasets for finetuning both the retrieval model and LLM, which is a foundation for building a domain-specific Thai legal RAG system. 
The dataset also contains a separate test set annotated by the domain experts.

\begin{figure}[H]
    \centering
    \includegraphics[width=0.8\textwidth]{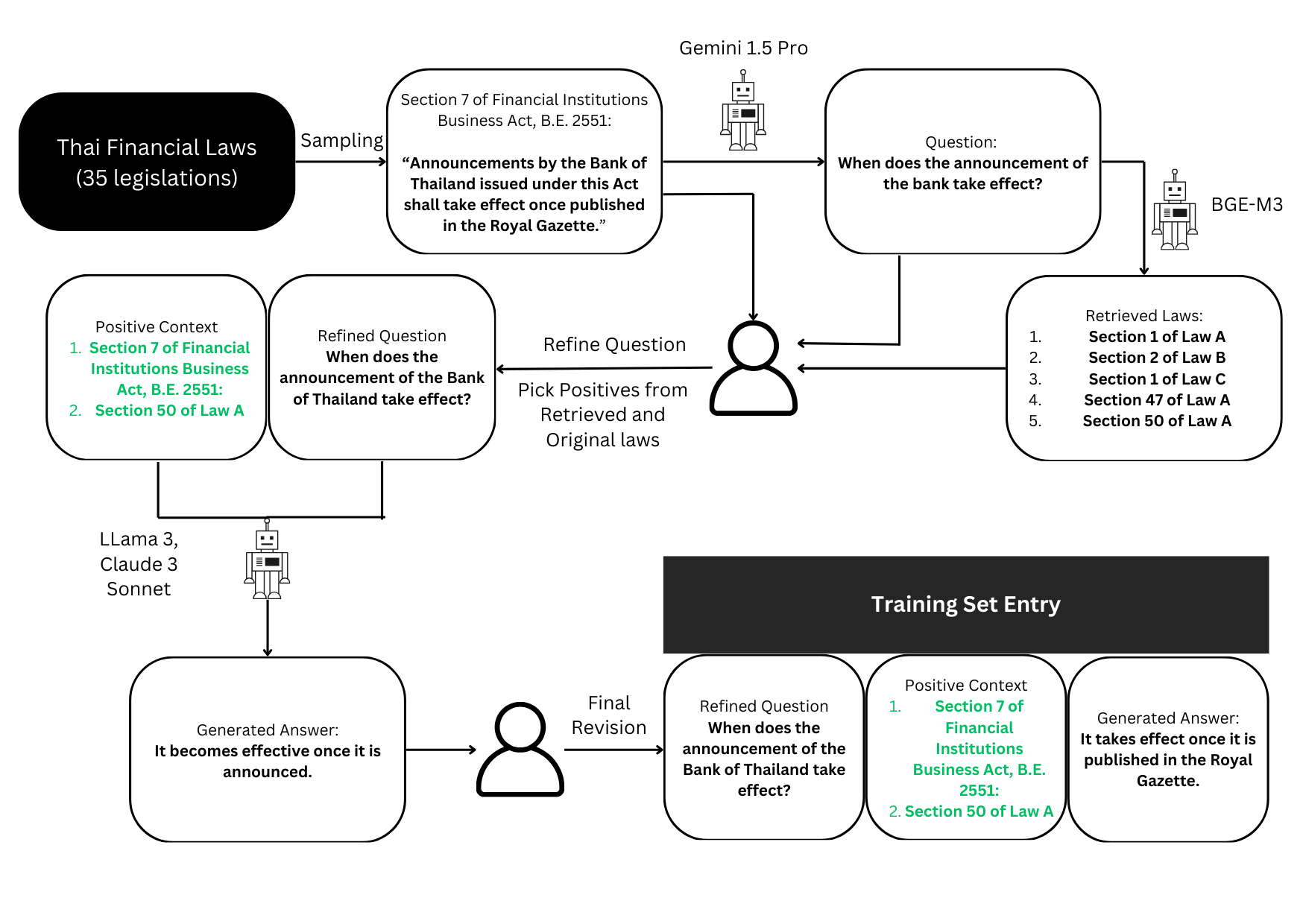}    
    \caption{Overall dataset construction pipeline for training set of \texttt{WangchanX-Legal-ThaiCCL-RAG} dataset}
    \label{fig:wcx_pipeline_train}
\end{figure}

\textbf{Curating Training Data.} Our training dataset is constructed using a semi-automated approach combining synthetic data generation with expert validation. 
First, we automatically generate questions from each legal section from the available 35 legislations within Thai Financial Laws, including the Civil and Commercial Code, Revenue Code, and various other financial laws (see \S\ref{tab:wcxccc}) using Gemini 1.5 Pro \cite{gemini1.5}. 
We then retrieve the top 5 most relevant sections using BGE-M3 embeddings \cite{bge-m3} before asking Legal experts to subsequently refine these retrieved sections alongside the original legal section the question is generated from. 
The Legal experts act as annotators, labeling each retrieved section as positive if relevant. Answers are then generated using \texttt{Meta-Llama-3-70B} \cite{llama3.1}, or \texttt{claude-3-sonnet} \cite{claude3} if Llama3 fails to produce an answer. 
The answer is conditioned on both the labeled positive context and the question. 
Next, another expert will review and can either accept the question-answer pairs, adjust the LLM-generated answer and question again to ensure accuracy and comprehensiveness, or completely reject and discard the entry.
The process of constructing the training dataset is outlined in Figure~\ref{fig:wcx_pipeline_train}.

\textbf{Curating Test Data.} 
Our test set, conversely, is constructed entirely manually and undergoes a rigorous two-tiered expert review. 
Legal experts craft questions and answers based on specific legal sections. 
A separate group of legal experts then reviews these question-answer pairs. 
The experts can either accept the constructed question-answer pair or provide feedback to the original annotators, who will improve the data entry or reject it completely.
The test set covers only 21 out of 35 legislations in the training set.

It is important to note a key difference between the training and test sets: the training set inherently allows for multiple relevant sections (multi-label) per question, reflecting real-world legal scenarios where multiple legal provisions might apply. 
Conversely, due to the complexity of curating fully human-annotated multi-label data, the expert-curated test set primarily focuses on single-label question-answer pairs.
This discrepancy presents a limitation, potentially leading to an overly optimistic evaluation of models trained on the multilabel training set when tested on the single-label test set. 
Future work will explore bridging this gap, potentially through incorporating multi-label annotations in the test set or developing evaluation metrics robust to label multiplicity variations.
Additionally, we ensure that each dataset entry contains only laws within the scope of the 35 Thai financial law codes under consideration and that no duplicate legal sections appear within a single entry.

\begin{figure}[H]
    \centering
    \includegraphics[width=0.8\textwidth]{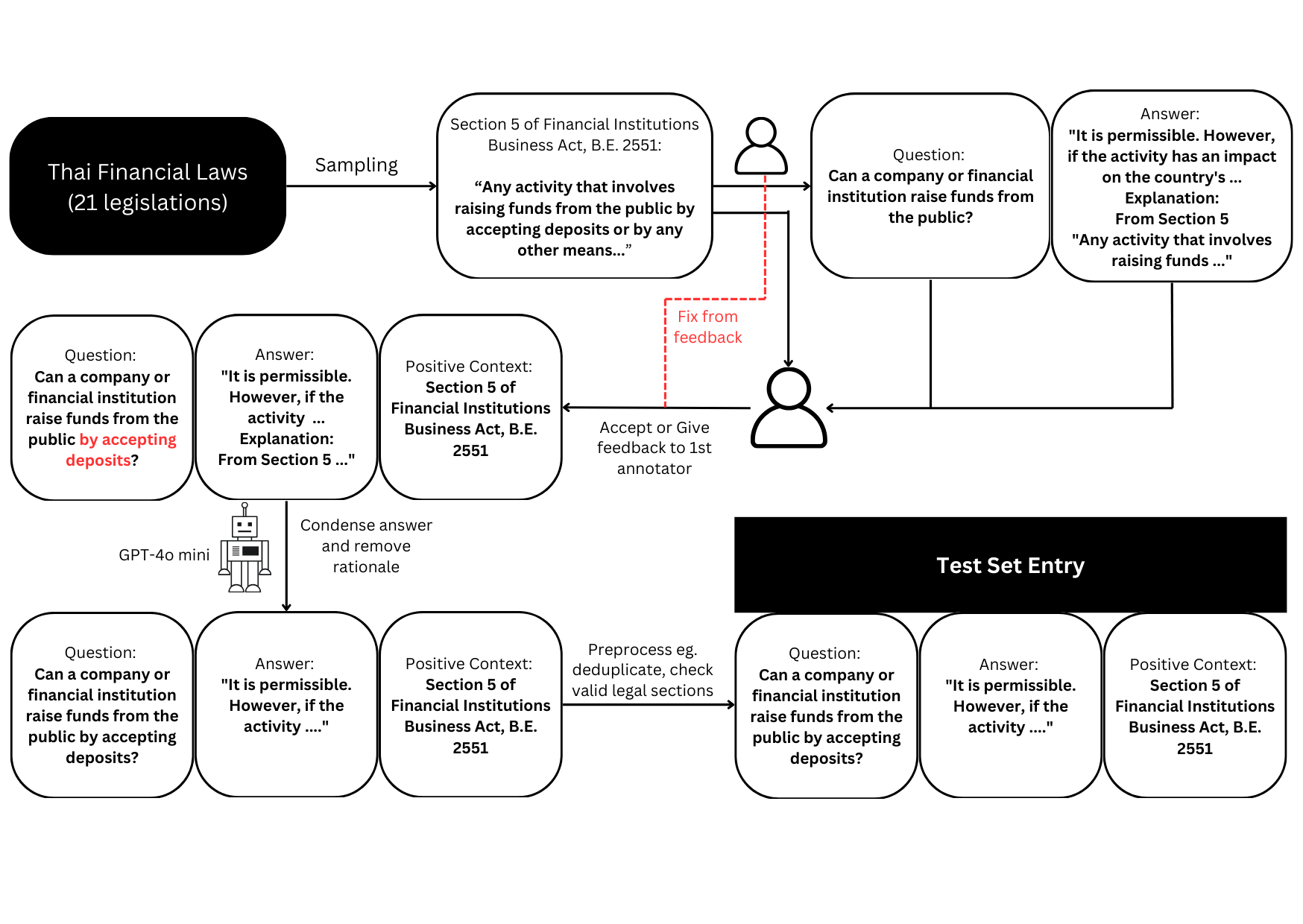}    
    \caption{Overall dataset construction pipeline for test set of \texttt{WangchanX-Legal-ThaiCCL-RAG} dataset and NitiBench-CCL. The original test split of \texttt{WangchanX-Legal-ThaiCCL-RAG} was completed after the annotator accepted all question/answer pairs from another annotator. NitiBench-CCL further augments this test set to condense answers using GPT-4o mini and conduct further postprocessing.}
    \label{fig:wcx_pipeline_test}
\end{figure}

Figure~\ref{fig:wcx_pipeline_test} shows the construction pipeline of the test set.

\textbf{NitiBench-CCL}\footnote{\url{https://huggingface.co/datasets/VISAI-AI/nitibench}} NitiBench-CCL extends the original WangchanX-ThaiLegal-CCL-RAG's test set by applying additioanl postprocessing step. Since the annotated contextual information includes the full content of relevant legal sections, we further preprocess the test set by extracting only the names of the referenced legal sections from the annotations and deduplicate entries with the same questions.

%
As annotated responses often include explanations or reasoning, which are outside the scope of this evaluation, we utilize a LLM\footnote{Specifically, we use \texttt{gpt-4o-mini-2024-07-18}\cite{gpt4o} during September 2024} to extract only the essential answers without the accompanying rationale. 
Note that \textbf{we only apply this preprocessing to our experiment, and the released dataset still contains full answers.} 
Finally, to avoid the data leakage problem with the retriever fine-tuned on the train split of the dataset, only the test split is used for evaluation. 
The test set comprises 3,730 entries, referencing 21 out of 35 law codes present in the training set. 
Notably, 43\% of the queries pertain to the Civil and Commercial Code, and each query corresponds to a single relevant law section.

\textbf{Annotator Profile and Cost} 
Since we are curating a dataset specifically in the Thai legal domain, it is important to ensure that our annotators have a strong background in Thai legal knowledge. 
To achieve this, we recruited legal experts through law school professors via their available channels, such as their personal Facebook networks\footnote{\href{https://www.facebook.com/photo/?fbid=10230736147843884\&set=a.2166246952290}{https://www.facebook.com/photo/?fbid=10230736147843884\&set=a.2166246952290}}. 
We received a total of 97 applications and selected 34 annotators. 
Their occupations include law students, recent law school graduates, and employees at law firms.

We compensate annotators per completed task, which includes curating the training set, conducting quality checks, and curating the test set. 
Tasks are randomly assigned, and we adjust the distribution based on each annotator’s speed of completion. 
Payment is determined per task\footnote{To simplify the calculations, we use a conversion rate of 34 Thai baht per 1 US dollar.}, with each task compensated differently based on its difficulty. The tasks are as follows:

\begin{enumerate}
\item Rerank retrieved documents for the fine-tuning dataset: 5 THB (approximately \$0.15) per task.
\item Validate, correct, and reject the generated answers for both training and test data: 10 THB (approximately \$0.30) per task.
\item Create a question and answer based on a given law section (for the test set): 30 THB (approximately \$0.89) per task.
\end{enumerate}
The total cost spent solely on annotators is approximately 274,240 THB (roughly \$8076).

\begingroup
\renewcommand{\arraystretch}{1.4}
\begin{table}[]
\centering
\begin{tabular}{@{}p{0.6\linewidth}ccc@{}}
\toprule
\textbf{Legislation}                                                                                                                           & \textbf{Legal Terminology} & \textbf{Training} & \textbf{Test} \\ \midrule
Organic Act on Counter Corruption, B.E. 2561                                                                                          & organic law       & \checkmark        &      \\
Civil and Commercial Code                                                                                                             & code              & \checkmark        & \checkmark    \\
Revenue Code                                                                                                                          & code              & \checkmark        & \checkmark    \\
Act on Offenses Relating to Registered Partnerships, Limited Partnerships, Companies Limited, Associations and Foundations, B.E. 2499 & act               & \checkmark        & \checkmark    \\
Chamber of Commerce Act, B.E. 2509                                                                                                    & act               & \checkmark        & \checkmark    \\
Trade Association Act, B.E. 2509                                                                                                      & act               & \checkmark        & \checkmark    \\
Accounting Profession Act, B.E. 2547                                                                                                  & act               & \checkmark        & \checkmark    \\
Business Registration Act, B.E. 2499                                                                                                  & act               & \checkmark        & \checkmark    \\
Public Limited Companies Act, B.E. 2535                                                                                               & act               & \checkmark        & \checkmark    \\
Foreign Business Act, B.E. 2542                                                                                                       & act               & \checkmark        & \checkmark    \\
Accounting Act, B.E. 2543                                                                                                             & act               & \checkmark        & \checkmark    \\
Secured Transactions Act, B.E. 2558                                                                                                   & act               & \checkmark        & \checkmark    \\
Securities and Exchange Act, B.E. 2535                                                                                                & act               & \checkmark        & \checkmark    \\
Derivatives Act, B.E. 2546                                                                                                            & act               & \checkmark        & \checkmark    \\
Provident Fund Act, B.E. 2530                                                                                                         & act               & \checkmark        & \checkmark    \\
Trust for Transactions in Capital Market Act, B.E. 2550                                                                               & act               & \checkmark        & \checkmark    \\
Energy Industry Act, B.E. 2550                                                                                                        & act               & \checkmark        & \checkmark    \\
Energy Conservation Promotion Act, B.E. 2535                                                                                          & act               & \checkmark        & \checkmark    \\
Financial Institutions Business Act, B.E. 2551                                                                                        & act               & \checkmark        & \checkmark    \\
Petroleum Income Tax Act, B.E. 2514                                                                                                   & act               & \checkmark        & \checkmark    \\
Act Repealing the Agricultural Futures Trading Act, B.E. 2542 B.E. 2558                                                               & act               & \checkmark        &      \\
State Enterprise Development and Governance Act, B.E. 2562                                                                            & act               & \checkmark        &      \\
Government Procurement and Supplies Management Act, B.E. 2560                                                                         & act               & \checkmark        &      \\
State Enterprise Committee and Personnel Qualifications Standards Act, B.E. 2518                                                      & act               & \checkmark        &      \\
State Enterprise Labor Relations Act, B.E. 2543                                                                                       & act               & \checkmark        &      \\
State Enterprise Capital Act, B.E. 2542                                                                                               & act               & \checkmark        &      \\
Budget Procedure Act, B.E. 2561                                                                                                       & act               & \checkmark        &      \\
Act on Offences of Officials Working in State Agencies or Organizations, B.E. 2502                                                    & act               & \checkmark        &      \\
Act on the Management of Shares and Stocks of Ministers, B.E. 2543                                                                    & act               & \checkmark        &      \\
Fiscal Discipline Act, B.E. 2561                                                                                                      & act               & \checkmark        &      \\
National Economic and Social Development Act, B.E. 2561                                                                               & act               & \checkmark        &      \\
Act on Disciplinary Offenses of Government Officials Performing Duties in Agencies Other than Government Agencies, B.E. 2534          & act               & \checkmark        &      \\
Act on the Establishment of Government Organizations, B.E. 2496                                                                       & act               & \checkmark        &      \\
Emergency Decree on Special Purpose Juristic Person for Securitization, B.E. 2540                                                     & emergency decree  & \checkmark        & \checkmark    \\
Emergency Decree on Digital Asset Businesses, B.E. 2561                                                                               & emergency decree  & \checkmark        &      \\ \bottomrule
\end{tabular}

\caption{List of legislation distribution in both training and test splits in WangchanX Legal ThaiCCL RAG Dataset (sorted High to Low Legislative Rank, Alphabetical). The test column is the same as NitiBench-CCL since the data was derived from WangchanX Legal ThaiCCL RAG test set}
\label{tab:wcxccc}
\end{table}

\endgroup

\begin{figure}[H]
    \centering
    \begin{minipage}{0.46\textwidth}
        \centering
        \includegraphics[width=\textwidth]{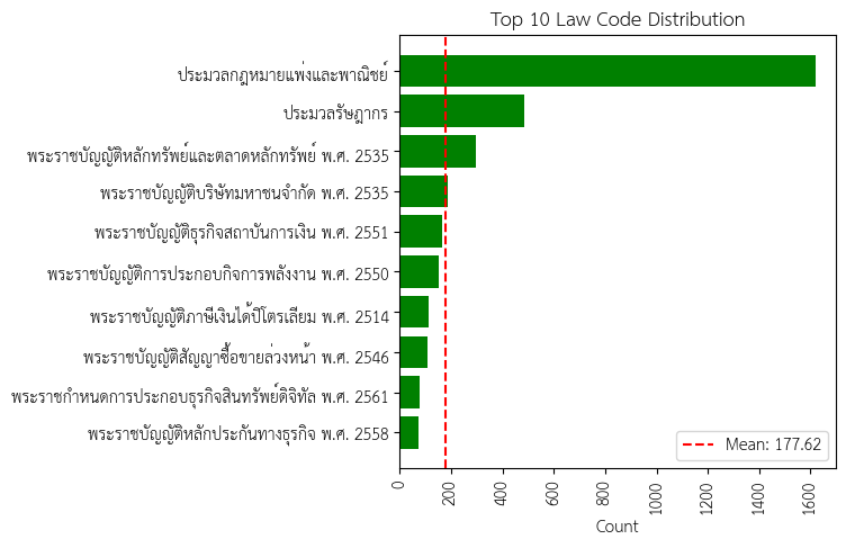}
    \end{minipage}%
    \hfill
    \begin{minipage}{0.26\textwidth}
        \centering
        \includegraphics[width=\textwidth]{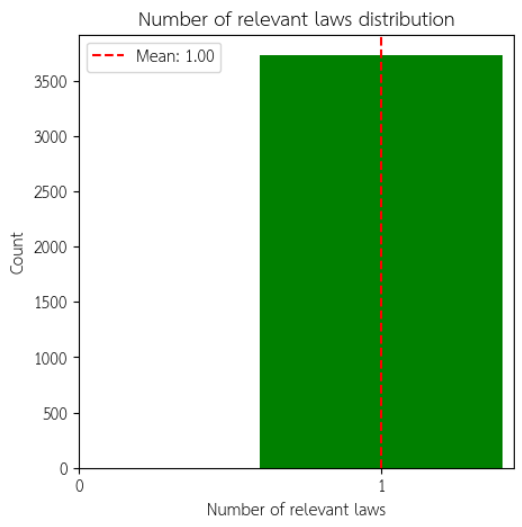}
    \end{minipage}%
    \hfill
    \begin{minipage}{0.26\textwidth}
        \centering
        \includegraphics[width=\textwidth]{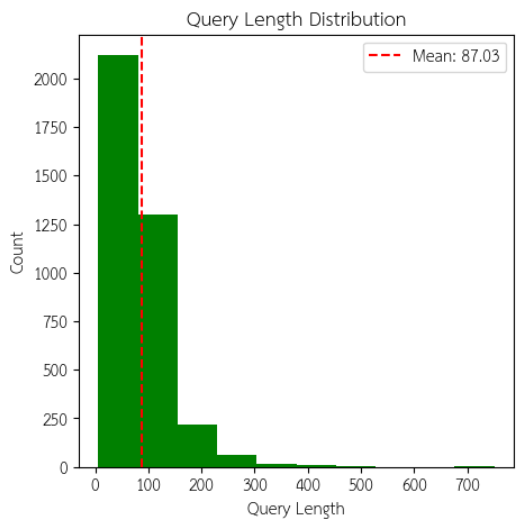}
    \end{minipage}

    \caption{Visualization of the WangchanX Legal ThaiCCL RAG Dataset. From left to right: (1) Number of times each legislation is used as relevant context; (2) Histogram showing the frequency of sections used as relevant context in a single entry; (3) Histogram showing the length of each query in characters.}
    \label{fig:combined}
\end{figure}

\subsubsection{NitiBench-Tax}
\label{subsubsec: tax_dataset}

To evaluate the generalization capability of the system, we curated an additional dataset derived from publicly available resources in the Thai financial legal domain.
Specifically, this dataset was created by scraping tax-related cases from the Revenue Department's official website\footnote{https://www.rd.go.th}. 
These cases represent authentic inquiries or requests (with personally identifiable information removed) submitted to the department. 
Each case includes the original inquiry or request, the official response, and metadata such as the case ID and submission date.

We extracted references to legislative sections mentioned in both the inquiry and the response as case attributes using LLM\footnote{We use only \texttt{gpt-4o-mini-2024-07-18}~\cite{gpt4o}. The API was called during September 2024.} for any preprocessing steps involving the use of LLM used during constructing NitiBench-Tax. 
The dataset was filtered to retain only cases referencing laws within the 35 Thai financial law codes and to eliminate duplicate references within individual entries. 
Some cases, however, involve inquiries requesting discretionary decisions from the department, such as extensions for tax deadlines or tax exemptions, rather than informational responses based on statutory interpretation. 
Since these cases are outside the scope of our work, which focuses on law-based reasoning, they were identified using an LLM and subsequently removed.

Additionally, to align with our evaluation objectives, the department's responses were condensed to essential answers, excluding detailed explanations and rationales.
Finally, we restricted the dataset to cases from 2021 onward, reflecting the most recent legislative updates. 
The resulting NitiBench-Tax consists of 50 cases, predominantly related to the Revenue Code, with an average of three referenced legal sections per case. 
This dataset provides a challenging testbed for evaluating system performance in a specialized domain requiring nuanced legal reasoning and multi-label retrieval.

\begin{figure}[H]
    \centering
    \includegraphics[width=0.8\textwidth]{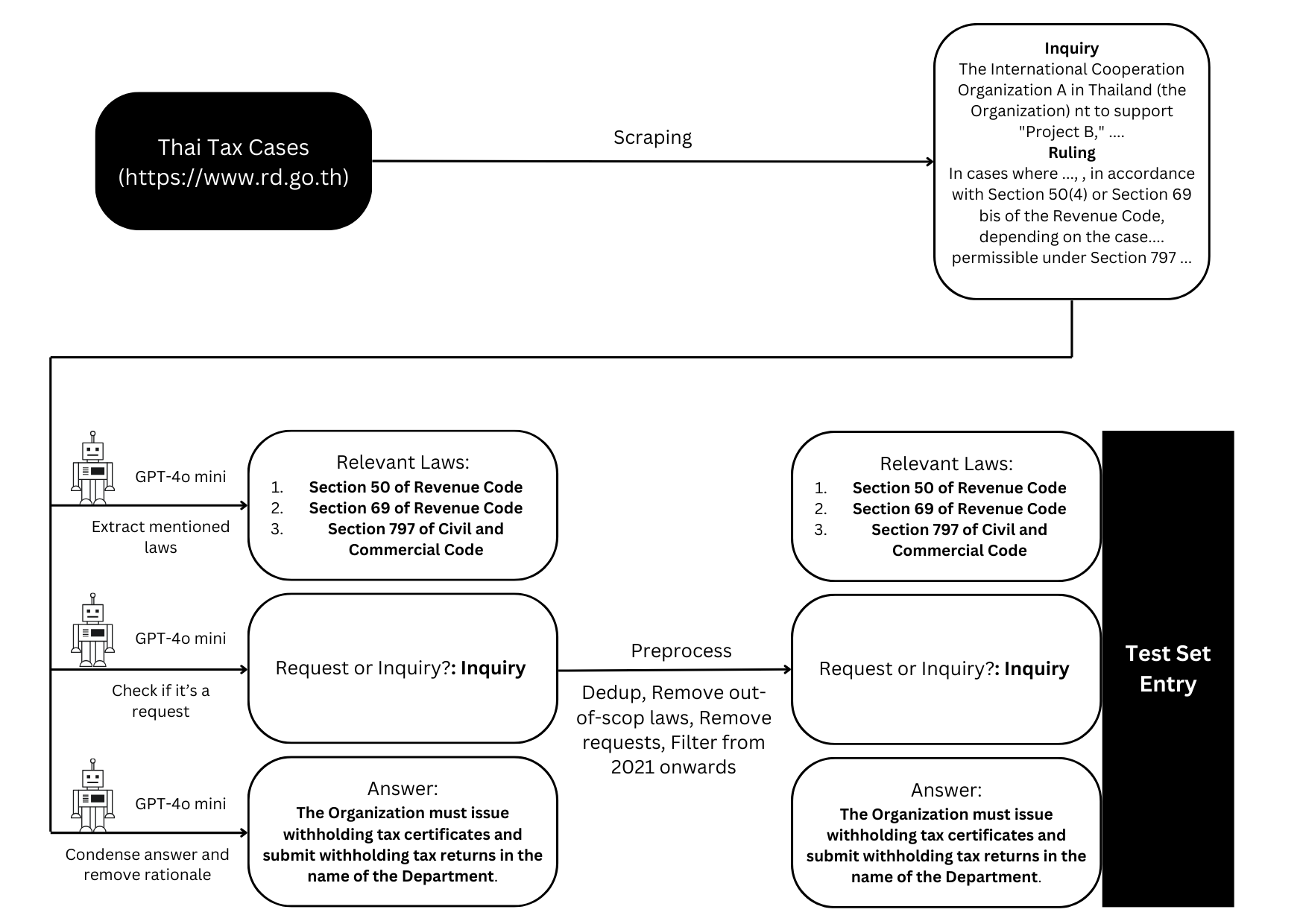}
    
    \caption{Overall dataset construction pipeline for NitiBench-Tax}
    \label{fig:tax_pipeline}
\end{figure}

The complete dataset construction pipeline is outlined in Figure~\ref{fig:tax_pipeline}

\begin{figure}[H]
    \centering
    \begin{minipage}{0.34\textwidth}
        \centering
        \includegraphics[width=\textwidth]{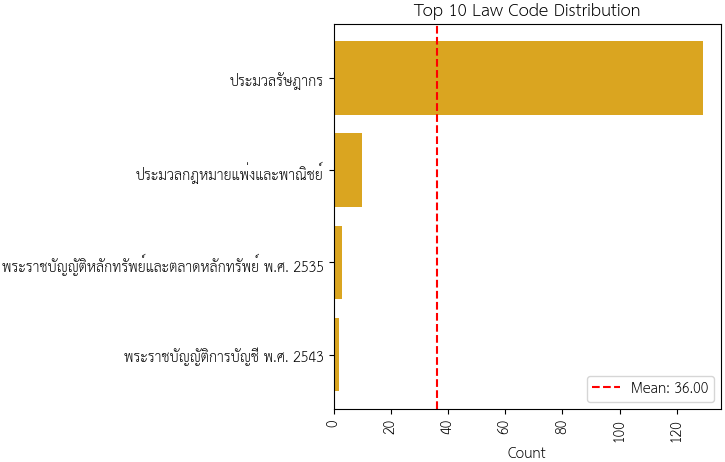}
    \end{minipage}%
    \hfill
    \begin{minipage}{0.31\textwidth}
        \centering
        \includegraphics[width=0.7\textwidth]{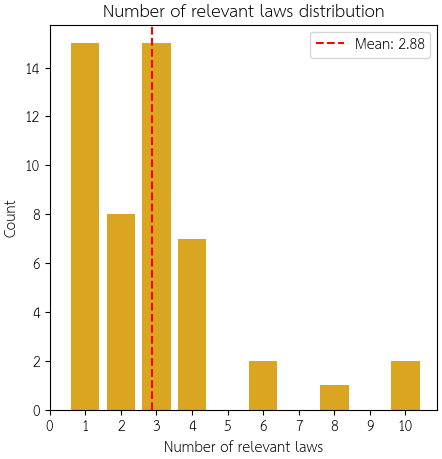}
    \end{minipage}%
    \hfill
    \begin{minipage}{0.31\textwidth}
        \centering
        \includegraphics[width=0.7\textwidth]{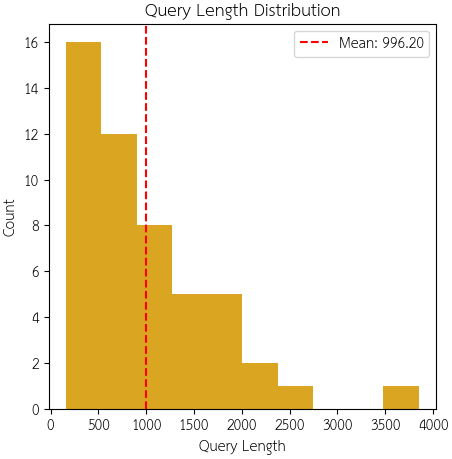}
    \end{minipage}

    \caption{Visualization of the NitiBench-Tax. From left to right: (1) Number of times each legislation is used as relevant context; (2) Histogram showing the number of sections used as relevant context in a single entry; (3) Histogram showing the length of each query in characters.}
    \label{fig:tax_combined}
\end{figure}

\subsection{Task}
\label{subsec: task}

For end-to-end evaluation, we adapt the task proposed by Laban et al.~\cite{laban2024summaryhaystackchallengelongcontext} to suit the Thai legal domain.
While the original task emphasized summarizing long contexts into bullet points with references, our QA-oriented dataset necessitates a different focus. 

Specifically, given a query and an optional context comprising legal sections, the system must generate an accurate, relevant answer and provide a list of cited legal sections used as references. 
This task assesses the system’s ability to select pertinent legal sections and perform legal reasoning to produce a complete, correct response.

\subsection{Metrics}
\label{subsec: metric}
Next, we describe the metrics used in both retrieval and E2E evaluation proposed to measure the performance of the Thai legal QA system in our benchmark.

\subsubsection{Retriever Metrics}
\label{subsubsec: retriever_metric}
As mentioned in \S\ref{sec:lit_review}, a good benchmark should be able to capture the performance of the retrieval component in RAG-based QA systems and should support multiple positive relevant document behaviors commonly observed in the legal domain. 
We modified the traditional information retrieval metrics, including Hit Rate, Recall, and MRR, which are designed for a single positive label to accommodate a multi-label scenario, enabling their use with datasets like ours that feature multiple positive labels.

\textbf{HitRate:} Let $N$ be the number of documents in a dataset and $k$ represents the number of top retrieved documents being evaluated, $T_i$ represents the set of positive relevant documents and $R_i^k$ represents the top-$k$ ranked retrieved documents of the $i^\text{th}$ sample. 
HitRate(HR) can be defined as

\begin{equation}
    \text{HitRate@k}=\frac{1}{N}\sum_{i=1}^{N}\mathbb{I}(R_i^k\subseteq T_i) 
\end{equation}

\textbf{Multi HitRate: } To ensure that hit rate supports multi-positive setup. 
We defined a multi-label version of the hit rate where the metric requires the system to retrieve all true relevant documents to be considered a hit. 
Formally

\begin{equation}
    \text{Multi-HitRate@k}=\frac{1}{N}\sum_{i=1}^{N}\mathbb{I}(T_i\subseteq R_i^k)
\end{equation}

\textbf{Recall: }As with HitRate, recall can be defined as

\begin{equation}
    \text{Recall@k}=\frac{1}{N}\frac{\sum_{i=1}^N |T_i\cap R_i^k|}{\sum_{i=1}^N |T_i|}
\end{equation}

\textbf{MRR and Multi-MRR: }In the traditional calculation of MRR with multi-label ground truth, the metrics can be written as follows:

\begin{equation}
    \text{MRR@k} = \frac{1}{N}\sum_{i=1}^N \frac{1}{\text{argmax}(T_i \cap R_i^k)} \label{eq:single_mrr}
\end{equation}

where $\text{argmax}(T_i\cap R_i^k)$ represents the highest rank number of correctly retrieved documents. 
The metric is zero if $|T_i \cap R_i^k| = 0$ (retrieved document contains no positive.

In some multi-label setup works, MRR was calculated under the assumption that any of the documents in the ground truth set $T$ is a sufficient context for the system to answer the question \cite{DBLP:journals/corr/abs-2006-15498, 10.1145/3397271.3401075}. 
However, this assumption is not true, especially in a legal domain where, sometimes, all relevant laws must be retrieved in order for the system to be able to answer the question. Therefore, the equation~\ref{eq:single_mrr} is augmented to Multi-MRR as follows:

\begin{equation}
    \text{Multi-MRR@k}=\frac{1}{N}\sum_{i=1}^N
    \left[
        \frac{\text{Recall@k}_{i}}{|T_i \cap R_i^k|}\sum_{j=1}^{|T_i\cap R_i^k|}
            \frac{1}{rank(d_j) - j + 1}
    \right]
\end{equation}

where $rank(d_j)$ represents the rank of the $j$-th matched documents, that is $d_j \in T_i \cap R^k_i$. 
This augmented calculation produces the highest scores when all of the relevant documents are retrieved and ranked at the top and, thus, can provide an evaluation metric that closely aligns with our performance goals.

\subsubsection{End-to-End Metrics}
\label{subsubsec: e2e_metric}

Building upon the metric suite employed by Laban et al.~\cite{laban2024summaryhaystackchallengelongcontext}, an additional metric inspired by Dahl et al.~\cite{Dahl_2024} is introduced to quantify hallucination rates. The following metrics are employed for end-to-end evaluation:

\textbf{Coverage:} This metric assesses the degree to which the generated answer semantically encompasses the ground truth response. 
Complete coverage of all points mentioned in the golden answer yields a score of 100, partial coverage results in a score of 50, and complete divergence is assigned a score of 0. 
Automatic evaluation is conducted using an LLM as a judge, following established methodologies (\cite{phan2024ragvslongcontext}, \cite{es2023ragasautomatedevaluationretrieval}).

\textbf{Citation: } This metric assesses the system's ability to ground answers in the provided context and its effectiveness in long-context retrieval. 
Specifically, given the ground truth answer, the generated response, and the relevant legal sections, we extract the cited sections from the generated answer.
This output, produced by the LLM, includes both the textual answer and citations in the form of a list of sections. 
We then compute the Recall, Precision, and F1 against the actual relevant sections. 
To distinguish these from similar retrieval metrics, these scores are often prefixed with "E2E," such as E2E Precision, E2E Recall, and E2E F1.

\textbf{Contradiction: } To assess hallucination, a simplified version of Dahl et al.'s \cite{Dahl_2024} approach is adopted. 
Rather than generating multiple answer trajectories and assessing whether there are contradictions among the generated trajectories, the generated answer is directly compared to the ground truth. 
A contradiction score of 0 indicates alignment between the two, while a score of 1 signifies direct disagreement.

In order to control the quality of judge LLM when assessing coverage and citation score, we iteratively refine the judge prompt to maximize the human agreements in a held-out set. 
In this held-out set, we sampled 150 entries from the NitiBench-Tax (without filtering cases from 2021 onwards) and 200 entries from the training set of WangchanX-Legal-ThaiCCL. 
The judge LLM is then prompted with the sampled queries to answer the inquiries along with cited laws. 
The LLM answer is then parsed to another LLM alongside the ground truth answer to generate the coverage and citation score. 
The generated coverage and citation scores are then compared with the ones assigned by a human annotator, given the queries, generated answers, and ground truth answers.
The agreement between LLM-generated scores and human-annotated scores is measured by precision, recall, and F1. 
The prompt for LLM-as-a-judge is then adjusted until it yields more than 0.8 F1 between human-annotated scores and LLM-generated ones.

The final agreement score between human-annotated scores and LLM-generated ones is displayed in Table~\ref{table: agreement_metric}. 
The LLM-as-a-judge score is generated by \texttt{gpt-4o-2024-08-06} \cite{gpt4o} model with a temperature of 0.3.

\begin{table}[H]
\centering

\begin{tabular}{@{}cccccc@{}}
\toprule
\textbf{Metric}                         & \textbf{Dataset} & \textbf{Precision} & \textbf{Recall} & \textbf{F1-score} & \textbf{Support} \\ \midrule
\multirow{2}{*}{Coverage}      & Tax     & .83       & .83    & .83      & 150     \\
                               & CCL     & .88       & .88    & .88      & 200     \\ \midrule
\multirow{2}{*}{Contradiction} & Tax     & .92       & .91    & .91      & 150     \\
                               & CCL     & .98       & .97    & .98      & 200     \\ \bottomrule
\end{tabular}
\caption{Table displaying the weighted average precision, recall, and F1-score of LLM judge comparing to human experts in annotating coverage and contradiction score}
\label{table: agreement_metric}
\end{table}

The coverage agreement scores for the CCL and Tax are 0.88 and 0.83, respectively, which are lower than the contradiction agreement scores for both datasets. 
To further analyze this, we present the confusion matrices for each score in Tables~\ref{table: agreement_cm_wcx} and~\ref{table: agreement_cm_tax}.

\begin{table}[!ht]
    \centering
    \begin{minipage}{0.45\textwidth}
        \centering

        \resizebox{\textwidth}{!}{%
        \begin{tabular}{@{}lccc@{}}
        \toprule
                         & Predicted 0 & Predicted 50 & Predicted 100 \\ \midrule
        Ground Truth 0   & 8           & 2            & 3             \\
        Ground Truth 50  & 2           & 29           & 7             \\
        Ground Truth 100 & 1           & 9            & 139           \\ \bottomrule
        \end{tabular}
        }
        \caption{Confusion matrix for coverage agreement score on 200 WCX samples}
        \label{table: agreement_cm_wcx}
    \end{minipage}
    \hfill
    \begin{minipage}{0.45\textwidth}
        \centering
        
        \resizebox{\textwidth}{!}{%
        \begin{tabular}{@{}lccc@{}}
        \toprule
                         & Predicted 0 & Predicted 50 & Predicted 100 \\ \midrule
        Ground Truth 0   & 43          & 5            & 1             \\
        Ground Truth 50  & 6           & 35           & 6             \\
        Ground Truth 100 & 2           & 5            & 47            \\ \bottomrule
        \end{tabular}
        }
        \caption{Confusion matrix for coverage agreement score on 150 NitiBench-Tax samples}
        \label{table: agreement_cm_tax}
    \end{minipage}
\end{table}

As observed in the confusion matrices, it is rare for the LLM-as-a-judge to misclassify a ground truth score of 0 as 100 or vice versa. 
Most errors occur in the confusion between 50 and 100, as well as between 0 and 50. 
We consider this acceptable since the boundaries between these scores can sometimes be subjective. 
Therefore, despite the agreement score not being as high as initially expected after multiple iterations, we conclude that it remains reliable, achieving at least 80\% accuracy for the coverage score and at least 90\% accuracy for the contradiction score.

\section{Experimental Setups}
\label{sec:evaluation}
In this section, we outline the experimental setup used to leverage our benchmark dataset and the proposed evaluation process to address the aforementioned three key research questions:
\begin{description}
\item [\textbf{(RQ1)}:] What impact do components tailored to the Thai legal system structure have on a RAG-based Thai legal QA system?

\item [\textbf{(RQ2)}:] How do the choices of retriever and LLM affect performance in a RAG-based Thai legal QA system?

\item [\textbf{(RQ3)}:] How does a Long-Context LLM-based Thai legal QA system perform compared to RAG-based systems?
\end{description}

For each research question, we describe the purpose of the corresponding experiment, its design, and the models or components selected for evaluation. 
Note that for every E2E evaluation of RAG-based systems, we opt to use the top 10 retrieved documents as context for the generator, and the generator is provided with a few shot examples for the task. 
The few-shot examples are ensured to be from the entries not in the final test set for both the NitiBench-CCL (few-shot examples taken from WangchanX-Legal-ThaiCCL's train split) and NitiBench-Tax datasets. We use 3-shot examples in our experiments randomly sampled from our training data.

\subsection{(RQ1) Impact of tailored components}
\label{subsec: setup_rq1}
As outlined in \S\ref{subsec: thai_legal}, implementing LLM-based applications in the Thai legal domain presents two key challenges: 
(1) Thai legislation is organized hierarchically, requiring domain expertise to segment these documents into coherent chunks suitable for a RAG framework and 
(2) Thai legislation frequently includes inter-section references, necessitating the inclusion of nested legal contexts in addition to the standard retrieved context within the RAG framework. 
In this research question, we aim to quantitatively display the effect of components crafted to address these problems on the performance of the RAG-based Thai legal QA system.

\subsubsection{Hierarchy-aware Chunking}
\label{subsubsec: chunk_setup}

To address the limitations of traditional chunking methods, which are largely structure-agnostic (e.g. split documents into multiple chunks by sliding window algorithm) leading to information loss in RAG systems, we introduce \textbf{Hierarchy-aware Chunking}. 
This approach leverages the multi-tiered structure inherent in Thai legislation. 
Rather than relying on arbitrary character counts or generic separators, it utilizes domain knowledge to segment the legislation so that each chunk contains a single complete section of the law. 
This ensures that each chunk comprehensively represents a self-contained legal concept or rule, preserving the integrity of the original document.
The hierarchy-aware chunking was done via exhaustive regular expression and rule-based.

To evaluate the effect of the proposed component, we first quantify the baseline of the naive chunking strategy.
Since the naive chunking method requires a selection of different approaches and parameters, we seek to find the combination that minimized the \textit{information loss} compared to our proposed hierarchy-aware chunking and use the best combination as the naive chunking baseline.
Specifically, we assess information loss among three common naive chunking approaches varying its parameters, such as chunk sizes and overlaps. 
%
%
The chunking strategies considered are as follows:

\begin{enumerate}
    \item \textbf{Character Chunking: } Chunking is based purely on a fixed number of characters.
    \item \textbf{Recursive Chunking: } Chunking using various document structure-related separators.
    \item \textbf{Line Chunking: } Chunking based solely on newline characters.
\end{enumerate}

The chunk sizes in this experiment range from 212 characters (the 25th percentile of section lengths) to 553 characters (the 75th percentile), including intermediate values and one size beyond the 75th percentile. 
For chunk overlaps, the values 50, 100, 150, and 200 are tested. 

For each chunking strategy, we first chunk the legal documents and generate metadata indicating the sections each chunk covers and whether they are fully covered. 
Additionally, since the naive chunking strategy has no awareness of law section boundaries, the chunked text might either contain multiple law sections (if the section is shorter than the chunk size) or be incomplete (if the section is longer than the chunk size). 
This makes it hard to justify whether a retrieved incomplete chunk (partially containing law section content) is considered a correctly retrieved document. 
Therefore, to simply retrieve and enable a fair comparison of top-$k$ retrieval across strategies, chunks that do not fully cover at least one section are discarded based on the chunk metadata.
Next, the chunk's metadata are then used to measure the information loss of each naive chunking strategy.
We define \textit{information loss} of each naive chunking strategy using the following metrics:
%

\begin{enumerate}
    \item \textbf{Sections/Chunk: } The average number of sections (fully or partially covered) in each chunk. For hierarchy-aware chunking, this metric equals 1.0.
    \item \textbf{Chunks/Section: } The average number of chunks that fully or partially cover any given section. For hierarchy-aware chunking, this metric equals 1.0.
    \item \textbf{Fail Chunk Ratio: } The ratio of chunks that do not fully cover any section compared to all chunks. For hierarchy-aware chunking, this metric equals 0.
    \item \textbf{Fail Section Ratio: } The ratio of sections not fully covered by any single chunk, compared to all sections. For hierarchy-aware chunking, this metric equals 0.
    \item \textbf{Uncovered Section Ratio: } The ratio of sections neither fully nor partially covered by any single chunk, compared to all sections. For hierarchy-aware chunking, this metric equals 0.
\end{enumerate}

The naive chunking strategy that yields the best performance based on these metrics is then selected as the optimal naive chunking strategy and represents our naive chunking baseline.
After selecting the optimal naive chunking strategy based on the above metrics, we integrate it into a RAG system and compare it against the hierarchy-aware chunking approach. 
To ensure a fair comparison, we remove sections from the hierarchy-aware chunks that are not covered by the naive chunking strategy due to content length constraints.
Additionally, the dataset is augmented to include only relevant laws within the sections available in the naive chunking strategy.  

After filtering incomplete chunks, only 19 NitiBench-Tax entries and 2,625 NitiBench-CCL entries were left.
Given the limited size of the NitiBench-Tax subset, we perform evaluations solely on the NitiBench-CCL. 
Both retrieval and end-to-end (E2E) performance are assessed to comprehensively analyze the impact of the proposed component.

For retrieval evaluation, retrieved chunks from the RAG system using naive chunking are mapped to sections before computing retrieval metrics. 
The mapping process only associates chunks with sections that are fully contained within them. 
The retriever in this setup is a three-headed, human-finetuned BGE-M3 model, as described in \S\ref{subsubsec: retriever_setup}, while the LLM used is \texttt{gpt-4o-2024-08-06} with few-shot examples, as detailed in \S\ref{subsubsec: llm_setup}.

We note that this E2E experiment does not fully capture the advantages of hierarchy-aware chunking over naive chunking in real-world scenarios. 
In practical applications, the naive chunking strategy often retains chunks that do not fully cover any section, raising questions about the LLM's and retriever's ability to process and utilize fragmented sections effectively.
However, we structured the experiment this way to enable a more direct comparison between chunking strategies under the same top-$k$ setting.

\subsubsection{NitiLink}
\label{subsubsec: referencer_setup}

Another component we propose to address the issue of inter-references between sections in Thai legislation is \textbf{NitiLink}. 
This component is designed to recursively fetch additional sections referenced within the initially retrieved sections in a depth-first manner. 
For example, understanding Section 291 of the Securities and Exchange Act B.E. 2535 requires knowledge of Section 186 from the same legislation. 
Using the proposed NitiLink, the system retrieves Section 186, identified as a child of Section 291, along with any other children within Section 186 and deeper levels, as additional context for the generator in the RAG system.

\begin{quote}
    \textit{Section 291 of Securities and Exchange Act B.E. 2535: Anyone who violates or fails to comply with orders issued under Section 186(2) shall be liable to a penalty of imprisonment for not more than one year, a fine of not more than 300,000 baht, or both.}
\end{quote}

We emphasize that NitiLink is implemented in a depth-first manner. 
For example, if the initially retrieved sections are ranked as [Section A with reference to Section U and Section W, Section B with reference to Section X and Section U, Section C], NitiLink fetches the referenced sections, re-ranking the results as [Section A, Section U, Section W, Section B, Section X, Section C]. 
While this can degrade MRR if the referenced section is not a hit, we argue that referenced sections should rank higher than originally lower-ranked sections due to their direct relevance to the higher-ranked section, thus providing a more complete and continuous context for answering the query, especially in legal tasks requiring precise reasoning.

NitiLink allows for configuring a maximum depth to limit the extent of nested sections retrieved.
In this study, we set the maximum depth to 1 due to computational feasibility and budget constraints since more depth means a longer context is required. 
More studies on the impact of referencing depth are conducted in \S\ref{subsec: ref_depth_vs_retrieval_perf}.

To assess the impact of the NitiLink, we utilized the benchmark dataset and evaluation process to compare the performance of a RAG-based system with NitiLink (configured with a maximum depth of 1) against a system without this component. 
The evaluation was conducted for both retrieval and end-to-end (E2E) performance. 
When calculating the system with NitiLink, we combine both the referenced sections and the retrieved section, as this would allow us to analyze the impact of NitiLink directly on enhancing correctness. 
In other words, we attempt to see if incorporating the referenced sections would result in a more accurate retrieval.

For this experiment, we used hierarchy-aware chunking with three-headed Human-Finetuned BGE-M3 (described in \S\ref{subsubsec: referencer_setup}) as the retriever and \texttt{gpt-4o-2024-08-06} as the LLM with few-shot examples in the RAG system configuration.

\subsection{(RQ2) Impact of Retriever and LLM}
\label{subsec: setup_rq2}
When implementing a RAG-based legal QA system, numerous design choices must be made for its components, with two of the most critical choices being the selection of the \textit{retriever model} and the \textit{LLM}. 
In this research question, we aim to analyze the performance gap and conduct a comparative evaluation of various retrievers and LLMs, focusing on both retrieval and E2E performance.

To answer this research question step-by-step, we decompose the experiment into three steps.
First, in \S\ref{subsubsec: retriever_setup}, we address the question \textit{How well do publicly and commercially available embedding models perform in Thai legal QA, and what are the limitations?}. 
Then, we seek to answer another question \textit{What is the performance of the available LLM suitable to answer Thai legal QA?} in \S\ref{subsubsec: llm_setup}.
Finally, we assess the effectiveness of RAG systems enhanced by the best setups from prior experiments against several baselines, including parametric knowledge QA, naive RAG, and RAG with golden context. 
This comprehensive evaluation provides insights into how current RAG systems perform relative to their theoretical lower and upper performance bound and is covered in \S\ref{subsubsec: e2e_best_setup}.

\subsubsection{Retriever}
\label{subsubsec: retriever_setup}
To quantify the performance gaps of current retrievers in the Thai Legal QA domain, we employ several widely-used retrieval models ranging from simple algorithms to ColBERT-based, hybrid-based, decoder-based, and closed-source retrievers within a RAG system. 
These models are evaluated on top-$k$ retrieval tasks using our benchmark dataset and the custom-designed multi-label retrieval metrics.

Specifically, the retriever models employed in these experiments are as follows:
\begin{enumerate}
    \item \textbf{BM25} \cite{bm25}: A retrieval algorithm that ranks a set of documents based on term frequency and inverse document frequency. 
    It does not consider semantic similarity and serves as a baseline in our experiments.
    \item \textbf{JinaAI Colbert V2\footnote{\url{https://huggingface.co/jinaai/jina-colbert-v2}}} \cite{jina-v2}: A ColBERT-based \cite{colbert} retrieval model that employs token-level embeddings and late interaction mechanisms tailored for information retrieval tasks.
    \item \textbf{JinaAI Embeddings V3\footnote{\url{https://huggingface.co/jinaai/jina-embeddings-v3}}} \cite{jina-v3}: A multilingual, multi-task text embedding model designed for tasks such as retrieval, clustering, classification, and text matching. Its architecture is based on an adapted version of XLM-Roberta, producing dense embeddings.
    \item \textbf{NV-Embed V1\footnote{\url{https://huggingface.co/nvidia/NV-Embed-v1}}} \cite{nvembed}: A decoder-based embedding model trained with an LLM as its base. 
    It incorporates novel techniques such as latent vector attention and two-stage instruction tuning, enhancing performance on both retrieval and non-retrieval tasks.
    \item \textbf{BGE-M3\footnote{\url{https://huggingface.co/BAAI/bge-m3}}} \cite{bge-m3}: A multilingual embedding model capable of dense, multi-vector, and sparse retrieval. 
    For this work, we weight the scores obtained from three embedding types to enhance retrieval accuracy.
    \item \textbf{Human-Finetuned BGE-M3\footnote{\url{https://huggingface.co/airesearch/WangchanX-Legal-ThaiCCL-Retriever}}}: A version of BGE-M3 finetuned on the train split of the WangchanX-Legal-ThaiCCL dataset (see \S\ref{subsubsec: wcx_dataset}). 
    Positive contexts are derived from human reranking of initially retrieved contexts from BGE-M3.
    \item \textbf{Auto-Finetuned BGE-M3}: Another version of BGE-M3 finetuned on the train split of the WangchanX-Legal-ThaiCCL dataset (see \S\ref{subsubsec: wcx_dataset}). 
    Unlike the human-reranked version, positive contexts are determined using BGE-M3 Reranker V2 \cite{li2023making}\cite{bge-m3}, which computes relevance scores between queries and retrieved contexts. 
    Contexts with scores below a threshold of 0.8 are excluded.
    \item \textbf{Cohere Embeddings\footnote{\url{https://cohere.com/blog/introducing-embed-v3}}: } A commercialized text embedding model designed for tasks like retrieval, classification, and clustering. 
    We use the \texttt{embed-multilingual-v3.0} variant in our experiments.
\end{enumerate}

Notably, for all BGE-M3-based models, we use similarity scores from all three heads of the model with weightings of 0.4, 0.4, and 0.2 for dense embeddings, multi-vector, and sparse embeddings, respectively. 

\subsubsection{LLM}
\label{subsubsec: llm_setup}
Next, assuming the effect of LLM choices in a RAG-based system is independent of the retriever choices, we assess the impact of different LLMs as generators on the end-to-end (E2E) performance of a RAG-based Thai legal QA system. 
We fix other design choices, such as the chunking strategy and retriever model, while varying only the LLM selection and inclusion of NitiLink, as experiments in \S\ref{subsec: setup_rq1} showed no clear effect.
Specifically, we implement RAG systems with \textit{hierarchy-aware} chunking and the three-headed \textit{Human-Finetuned BGE-M3} retriever, using the following LLM choices:

\begin{enumerate}
    \item \textbf{GPT 4o}\cite{gpt4o}: OpenAI's autoregressive model, trained in an end-to-end fashion and capable of processing text, audio, images, and video. 
    Specifically, \texttt{gpt-4o-2024-08-06} is used in our experiment.
    \item \textbf{Claude 3.5 Sonnet}\cite{claude3.5sonnet}: The latest iteration of Anthropic's LLM, which improves upon its predecessor in benchmarks ranging from math to coding and reasoning problems. 
    Specifically, \texttt{claude-3-5-sonnet-20240620} is utilized in our experiment.
    \item \textbf{Gemini 1.5 Pro}\cite{gemini1.5}: Google's LLM outperforms its previous version across multiple benchmarks and has a maximum context window of 2M tokens, surpassing many closed-source LLMs. 
    Specifically, \texttt{gemini-1.5-pro-002} is used in our experiment.
    \item \textbf{Typhoon V2 70b}\cite{typhoon2}: An LLM specifically optimized for Thai language, continually pre-trained from the 70b variant of Meta's Llama 3.1\cite{llama3.1}, with post-training techniques to enhance Thai language performance while retaining the model’s original capabilities. 
    Specifically, \texttt{typhoon-v2-70b-instruct}\footnote{\url{https://huggingface.co/scb10x/llama3.1-typhoon2-70b-instruct}} is used in our experiment.
    \item \textbf{Typhoon V2 8b}\cite{typhoon2}: An LLM optimized for Thai language, continually pre-trained from the 8b variant of Meta's Llama 3.1\cite{llama3.1}, with post-training techniques to enhance Thai language performance without sacrificing the original model’s capabilities. 
    Specifically, \texttt{typhoon-v2-8b-instruct}\footnote{\url{https://huggingface.co/scb10x/llama3.1-typhoon2-8b-instruct}} is used in our experiment.
\end{enumerate}

These LLMs were selected because they represent the cutting-edge of both closed-source (GPT-4o, Gemini 1.5 Pro, and Claude 3.5 Sonnet) and open-source (Typhoon v2) models, with a focus on the Thai language for the open-source models. 
All LLMs are configured to generate a maximum of 2048 tokens with a temperature of 0.5, and the seed is set when possible to ensure consistent results. 
All LLMs are also provided with a few-shot examples as well.


\subsubsection{E2E with best setups}
\label{subsubsec: e2e_best_setup}
In addition to assessing the impact of retriever and LLM choices on system performance, we aim to establish the performance of the current RAG system in comparison to both lower and upper bounds. 
Specifically, we evaluate various systems capable of handling the Thai legal QA task, as described in \S\ref{subsec: task}, and measure their retrieval and E2E performance. 
The systems included in our experiments are as follows:

\begin{enumerate}
    \item \textbf{Parametric Knowledge: } A system that uses only instruction-tuned LLMs without any additional context. 
    This serves as the baseline, relying solely on the knowledge contained within the LLM’s training data.
    \item \textbf{Naive RAG: } A basic RAG system with traditional components, consisting of a retrieval module utilizing the best-performing naive chunking method and a generation module without additional enhancements. Note that we did not modify the benchmark dataset to contain only relevant laws within the available sections in the naive chunks as done in Experiments in \S\ref{subsubsec: chunk_setup}. 
    This represents the simplest implementation of a RAG system for the legal domain.
    \item \textbf{Proposed RAG: } A RAG system enhanced with custom components tailored to the structure of Thai law. 
    This includes the hierarchy-aware chunking method for the retrieval module and NitiLink component to fetch all cross-referenced sections as additional context, provided these components demonstrably improve performance based on the experiments in \S\ref{subsec: setup_rq1}. In this setup, we did not remove chunks or sections that are not presented in the naive chunks, and the benchmark datasets are not modified unlike the modifications done in \S\ref{subsubsec: chunk_setup}
    \item \textbf{RAG with Golden Context: } A system where only the generation module is used, provided with the prompt and ground truth context. 
    This represents the upper bound of expected performance for a RAG system.
\end{enumerate}

It is important to note that the same retriever models are used for both the Vanilla RAG and the Proposed RAG systems to ensure fair comparisons. Additionally, the same LLM with few-shot examples used in Vanilla RAG and Proposed RAG is utilized for Parametric Knowledge without the relevant laws in the example.  

\subsection{(RQ3) Performance of Long-Context LLM (LCLM)}
\label{subsec: setup_rq3}
The rise of long-context large language models (LCLMs) presents new opportunities to potentially replace traditional RAG systems in the Thai legal domain. 
With their ability to process extensive context, LCLMs can ingest all 35 Thai financial laws within our benchmark dataset and use them as context to respond to queries. 
This capability could eliminate the need for a RAG framework in Thai legal QA systems. 
Among the LLMs discussed in \S\ref{subsubsec: llm_setup}, the only model capable of processing all 35 legislations is Gemini 1.5 Pro, which supports a maximum context window of 2 million tokens. 
This experiment evaluates the performance of an LCLM-based Thai legal QA system—specifically Gemini 1.5 Pro—when it directly intakes all 35 legislations as context with a few-shot examples. 
The system's E2E performance is then compared to that of other baseline systems described in \S\ref{subsubsec: e2e_best_setup} using our proposed benchmark.

Another noteworthy use case of LCLMs in Thai legal QA systems is their application as \textit{retrievers} within a RAG framework. 
In this approach, the traditional embedding-based retriever is replaced with an LCLM capable of ingesting all available documents (35 legislations) and is conditioned via a few-shot prompt to output the top-$k$ most relevant documents. 
This method leverages the reasoning capabilities of LLMs, allowing them to retrieve relevant documents even for queries requiring additional logical steps. 
We evaluate this setup using Gemini 1.5 Pro as the LCLM retriever and compare its retrieval performance to traditional retrieval models on our proposed benchmark.

Lastly, it is important to note that due to budget constraints, the experiments for this research question are conducted on a 20\% stratified subset of NitiBench-CCL (based on relevant legislations) and the full NitiBench-Tax.

\section{Results}
\subsection{(RQ1) Impact of tailored components}
\label{subsec: rq1_result}

\subsubsection{Hierarchy-aware Chunking}
\label{subsubsec: chunking_result}

From the defined metrics that are used to quantify the information loss on each type of chunking in \S\ref{subsubsec: chunk_setup}, we can conclude that a good chunking strategy should minimize \textit{Fail Chunk Ratio}, \textit{Fail Section Ratio} and \textit{Uncovered Section Ratio}.
Minimizing these metrics will reduce the information loss of some sections or parts of sections that are missing from the chunks.
Additionally, \textit{Sections/Chunk} and \textit{Chunks/Section} should be close to 1 in order for sections \emph{not to be} split into multiple chunks and retain atomicity within each chunk.

We evaluate multiple configurations of chunking strategies, chunk sizes, and overlaps as described in \S\ref{subsubsec: chunk_setup} and present the average metrics for each chunking strategy in Table~\ref{table: chunking_by_type}. 
It is observed that the chunking strategy most closely resembling the output of our hierarchy-aware chunking strategy is line-based chunking. 

However, across all strategies, approximately 30\% of sections are not referenced in any chunks, and at least 41.7\% of sections are not fully contained within a single chunk. 
Further analysis indicates that around 20\% of all sections cannot be fully covered in a single chunk under any naive chunking strategy due to their extended length. 
This necessitates the retrieval model to retrieve multiple chunks to provide sufficient context.

An example of such a section is Section 44 of the \textbf{Emergency Decree on Digital Asset Businesses, B.E. 2561}, which cannot be fully covered in a single chunk across any naive chunking strategy. 
This is attributed to its lengthy content and the presence of multiple subsections separated by newline characters, which are commonly used as delimiters in many naive chunking approaches.

\begin{quote}
    \textbf{Section 44 of Emergency Decree on Digital Asset Businesses, B.E. 2561}
    
    It shall be presumed that the following persons, who exhibit behavior involving the buying or selling of digital tokens or engaging in forward contracts related to digital tokens in an unusual manner for themselves, are persons who possess or are aware of inside information as defined under Section 42:
    
    (1) Holders of digital tokens exceeding 5\% of the total tokens sold in each series by the issuer of digital tokens. This includes digital tokens held by their spouses, cohabiting partners in the manner of husband and wife, and their minor children.
    
    (2) Directors, executives, controlling persons, employees, or staff members of the affiliated entities of the digital token issuer who are in positions or roles responsible for, or with access to, inside information.
    
    (3) Ascendants, descendants, adoptive parents, or adopted children of persons specified under Section 43.
    
    (4) Siblings sharing the same father and mother, or the same father or mother as persons specified under Section 43.
    
    (5) Spouses or cohabiting partners in the manner of husband and wife of persons specified under Section 43 or individuals listed under (3) or (4).
    
    The term \enquote{affiliated entities of the digital token issuer} under (2) refers to parent companies, subsidiaries, or associated companies of the digital token issuer, as defined by the criteria set forth by the SEC Board's announcements.
\end{quote}

\begin{table}[!ht]
\centering

\resizebox{\textwidth}{!}{%
\renewcommand{\arraystretch}{1.3} 
\small 
\begin{tabular}{@{}lccccc@{}}
\toprule
\textbf{Chunking Strategy} & \multicolumn{1}{l}{\textbf{Section/Chunk $\rightarrow$1}} & \multicolumn{1}{l}{\textbf{Chunk/Section $\rightarrow$1}} & \multicolumn{1}{l}{\textbf{Fail Chunk Ratio $\downarrow$}} & \multicolumn{1}{l}{\textbf{Fail Section Ratio $\downarrow$}} & \multicolumn{1}{l}{\textbf{Uncovered Section Ratio $\downarrow$}} \\ \midrule
\cellcolor{lightgray}Hierarchy-aware  & \cellcolor{lightgray}{1.000}   & \cellcolor{lightgray}{1.000}    & \cellcolor{lightgray}{0.000}  & \cellcolor{lightgray}{0.000}  & \cellcolor{lightgray}{0.000}                                       \\
Character  & 3.098                             & 1.710                              & 0.819                                & 0.675                                  & 0.397                                       \\

Line       & \textbf{1.689}                    & \textbf{1.234}                    & \textbf{0.658}                       & \textbf{0.417}                         & \textbf{0.294}                              \\

Recursive  & \underline{1.793}                       &\underline{1.27}                        & \underline{0.741}                          & \underline{0.504}                            & \underline{0.381}                                 \\ \bottomrule
\end{tabular}
}
\caption{Information loss comparison between hierarchy-aware chunking compared to other naive chunking strategies. Since hierarchy-aware chunking consistently parses into a single law section, it was treated as an upper bound because no information loss occurred.}
\label{table: chunking_by_type}
\end{table}

For the specific configuration of line chunking that produces chunks most similar to hierarchy-aware chunking, we fix the chunking strategy while varying the chunk overlap and chunk size parameters. 
Increasing the chunk size results in more text per chunk, leading to higher \textbf{Sections/Chunk} and \textbf{Chunks/Section} values while reducing the \textbf{Fail Chunk Ratio}, \textbf{Fail Section Ratio}, and \textbf{Uncovered Section Ratio}. 
Similarly, increasing the overlap effectively increases the chunk size, producing comparable effects to directly increasing the chunk size.
Based on these observations, we select the optimal configuration for naive chunking as line chunking with a chunk size of 553 characters and a chunk overlap of 50 characters. The detailed results for this configuration are displayed in Appendix~\ref{appendix: chunk_hyper}.

Finally, the metrics for the selected naive chunking configuration are compared against hierarchy-aware chunking in Table~\ref{table: chunking_compare_metric}.

\begin{table}[!ht]
\centering

\resizebox{\textwidth}{!}{%
\renewcommand{\arraystretch}{1.3} 
\small 
\begin{tabular}{@{}lccccc@{}}
\toprule
\textbf{Chunking Strategy} & \multicolumn{1}{l}{\textbf{Section/Chunk $\rightarrow$1}} & \multicolumn{1}{l}{\textbf{Chunk/Section $\rightarrow$1}} & \multicolumn{1}{l}{\textbf{Fail Chunk Ratio $\downarrow$}} & \multicolumn{1}{l}{\textbf{Fail Section Ratio $\downarrow$}} & \multicolumn{1}{l}{\textbf{Uncovered Section Ratio $\downarrow$}} \\ \midrule
Hierarchy-aware chunking  & 1.000   & 1.000    & 0.000  & 0.000  & 0.000                                       \\
Line chunking (553 chunk size and 50 chunk overlap)  & 1.956                             & 1.180                              & 0.521                                & 0.323                                  & 0.156                                       \\ \bottomrule
\end{tabular}
}
\caption{Information loss comparison between perfect chunking strategy (hierarchy-aware chunking) and the best naive chunking setup.}
\label{table: chunking_compare_metric}
\end{table}

Apart from the evaluation of chunking in isolation in terms of information loss, we also present the evaluation results on our benchmark in Table~\ref{table: chunk_e2e_main}.

\begin{table}[ht!]
\centering
\resizebox{\textwidth}{!}{%
\begin{tabular}{@{}lccccccc@{}}
    \toprule
    \textbf{Settings} & \textbf{Retriever Multi MRR ($\uparrow$)} & \textbf{Retriever Recall ($\uparrow$)} & \textbf{Coverage ($\uparrow$)} & \textbf{Contradiction ($\downarrow$)} & \textbf{E2E Recall ($\uparrow$)} & \textbf{E2E Precision ($\uparrow$)} & \textbf{E2E F1 ($\uparrow$)} \\
    \midrule
    Naïve Chunking            & 0.786 & 0.935 & 86.6 & \textbf{0.050} & 0.882 & 0.613 & 0.722 \\
    Hierarchy-aware Chunking  & \textbf{0.834} & \textbf{0.942} & \textbf{86.7} & 0.054 & \textbf{0.894} & \textbf{0.630} & \textbf{0.739} \\
    \bottomrule
\end{tabular}%
}
\caption{Effect of chunking configuration on E2E performance on NitiBench-CCL}
\label{table: chunk_e2e_main}
\end{table}



From Table~\ref{table: chunk_e2e_main}, the naive chunking strategy performs worse than hierarchy-aware chunking in terms of retrieval performance. 
This discrepancy likely arises because naive chunks often contain content from multiple sections, introducing \enquote{noise} that can negatively impact the retrieval model's ranking of relevant documents.  

However, in terms of end-to-end (E2E) performance, the system using hierarchy-aware chunking only slightly outperforms the one using naive chunking. 
We suspect that this is because the LLM can effectively filter out the \enquote{noise} in the retrieved sections during answer generation. 
As a result, the coverage and contradiction scores are not significantly different between the two systems.
Nevertheless, there remains a discrepancy in the E2E citation score.  

In conclusion, \textbf{hierarchy-aware chunking achieves a slight but consistent advantage over the naive chunking strategy.}

\subsubsection{NitiLink}
\label{subsubsec: referencer_result}

The evaluation results of the experiment described in \S\ref{subsubsec: referencer_setup} are presented in Table~\ref{table: augmenter_e2e_main}. 
In the table, ``Ref Depth 1'' denotes a RAG system that incorporates a NitiLink component with a maximum depth of 1, while \enquote{No Ref} represents a RAG system without NitiLink. 
For the metrics, ``NitiLink'' indicates retrieval metrics calculated on the augmented context, which includes both the initially retrieved sections and the additional sections fetched by NitiLink.

\begin{table}[!ht]
\centering
\renewcommand{\arraystretch}{1.3}
\newcommand{\gray}{\cellcolor{gray!15}}
\newcommand{\pos}[1]{\textcolor{darkgreen}{(#1\%)}}
\newcommand{\negv}[1]{\textcolor{red}{(#1\%)}}

\begin{tabular}{lcccccc}
\toprule
\multirow{2}{*}{\textbf{Metric}} & \multicolumn{3}{c}{\textbf{NitiBench-CCL}} & \multicolumn{3}{c}{\textbf{NitiBench-Tax}} \\ 
 & \textbf{No Ref} & \gray \textbf{Ref Depth 1} & $\Delta$ & \textbf{No Ref} & \gray \textbf{Ref Depth 1} & $\Delta$ \\ 
\midrule
\multicolumn{7}{c}{\textbf{Retriever Metrics}} \\ 
\midrule
MRR ($\uparrow$)       & \multicolumn{2}{c}{0.809} & -  & \multicolumn{2}{c}{0.574} & -  \\
Multi MRR ($\uparrow$) & \multicolumn{2}{c}{0.809} & -  & \multicolumn{2}{c}{0.333} & -  \\
Recall ($\uparrow$)    & \multicolumn{2}{c}{0.938} & -  & \multicolumn{2}{c}{0.437} & -  \\
\midrule
\multicolumn{7}{c}{\textbf{NitiLink Metrics}} \\ 
\midrule
MRR ($\uparrow$)             & 0.809  & \gray 0.800  & \negv{-1.11}  & 0.574  & \gray \textbf{0.582}  & \pos{+1.39}  \\
Multi MRR ($\uparrow$)       & 0.809  & \gray 0.800  & \negv{-1.11}  & 0.333  & \gray \textbf{0.345}  & \pos{+3.60}  \\
Recall ($\uparrow$)          & 0.938  & \gray \textbf{0.940}  & \pos{+0.21}  & 0.437  & \gray \textbf{0.535}  & \pos{+22.4}  \\
Coverage ($\uparrow$)        & 85.2   & \gray \textbf{86.3}  & \pos{+1.29}  & \textbf{50.0}   & \gray 45.0   & \negv{-10.0}  \\
Contradiction ($\downarrow$) & 0.055  & \gray \textbf{0.051}  & \pos{-7.27}  & \textbf{0.460}  & \gray 0.520  & \negv{+13.0}  \\
E2E Recall ($\uparrow$)      & 0.880  & \gray \textbf{0.885}  & \pos{+0.57}  & 0.333  & \gray \textbf{0.354}  & \pos{+6.31}  \\
E2E Precision ($\uparrow$)   & 0.601  & \gray 0.579  & \negv{-3.66}  & \textbf{0.640}  & \gray 0.630  & \negv{-1.56}  \\
E2E F1 ($\uparrow$)          & \textbf{0.714}  & \gray 0.700  & \negv{-1.96}  & 0.438  & \gray \textbf{0.453}  & \pos{+3.42}  \\
\bottomrule
\end{tabular}
\caption{Effect of NitiLink augmenter configuration on E2E performance. The $\Delta$ column shows the relative percentage change compared to ``No Ref'', with dark green indicating improvement and red indicating degradation.}
\label{table: augmenter_e2e_main}
\end{table}

The results from Table~\ref{table: augmenter_e2e_main} show that there is no clear significant advantage when employing NitiLink in a RAG system. 
The results also highlight the differing impacts of incorporating NitiLink across datasets.

\textbf{NitiBench-Tax} For this dataset, we can clearly see that the recall was substantially improved from 0.437 to 0.549.
The improvement of recall suggested that NitiLink does provide an additional correct law section to the retrieved documents. 
Despite significant improvement over recall, we only see marginal improvements over MRR and Multi MRR.
Since we're using a depth-first augmented strategy (see \S\ref{subsubsec: referencer_setup}), this suggested that the document that cited more positives by NitiLink is ranked at the bottom of the retrieved documents.
Surprisingly, despite a major improvement in recall, some E2E metrics declined.
This might be due to NitiBench-Tax's query complexity, which often demands advanced reasoning capabilities that the LLM, even with the correct documents, struggles to provide. 
Another reason that might affect the performance decline even with more relevant documents provided to the LLM is the longer context that the LLM needs to process due to the higher amount of content added by NitiLink.

\textbf{NitiBench-CCL} For NitiBench-CCL showed no significant change in retrieval metrics and most E2E metrics.
Incorporating NitiLink yields very little recall gain, while MRR is slightly lower. 
This means that NitiLink often pushed the positive lower in the ranking as we're using a depth-first augmentation strategy (see \S\ref{subsubsec: referencer_setup}).
We highlight several factors that might contribute to the limited recall gain in this dataset:
\begin{enumerate}
    \item \textbf{Binary recall nature:} NitiBench-CCL queries typically involve a single relevant law, making recall binary and thus harder to improve.
    \item \textbf{Simplicity of NitiBench-CCL queries:} Simple, non-specific NitiBench-CCL queries often rely on many relevant law sections that are similar semantically rather than hierarchically. 
    This is opposed to NitiBench-Tax, where referenced law sections are necessary for legal reasoning.
    This simplicity stems from the fact that the dataset was created by letting the annotator craft a question based on a given law section.
    This explicitly provides bias toward the dataset since the question was created without a referenced law section.
    \item \textbf{Hierarchical limiation:} The hierarchical structure itself presents challenges. 
    Although NitiLink augmented the retrieved law section mentioned in the retrieved document (children reference), it lacks a law section that references retrieved law sections (parent reference).  
    Thus, this version of NitiLink that lacks the ability to fetch parent law sections could result in a suboptimal performance.
    %
\end{enumerate}

Despite the limited recall gains in NitiBench-CCL, we can see that there's a slight improvement in coverage score as well as recall.
This suggests that even small recall improvements can enhance the LLM's ability to answer NitiBench-CCL queries effectively.

\subsection{(RQ2) Impact of Retriever and LLM}
\label{subsec: rq2_result}

\subsubsection{Retriever}
\label{subsubsec: retriever_result}

\textbf{NitiBench-CCL} 
Table~\ref{table: retrieval_wangchan} presents the retrieval performance of 8 models as described in \S\ref{subsubsec: retriever_setup} on NitiBench-CCL with hierarchy-aware chunking. 
Because each query has only one positive label (as mentioned in \S\ref{subsubsec: wcx_dataset}), the multi-hit-rate and multi-MRR metrics are equivalent to their single-label counterparts. This is also the case between recall and hit-rate as well. Thus, for this dataset, we only showed Hit-Rate (HR@k) and MRR (MRR@K) since other metrics are considered redundant.

The best-performing model is the human-reranked fine-tuned BGE-M3, achieving an MRR@5 of 0.805. 
Close behind are the auto-reranked fine-tuned BGE-M3 (0.800 MRR@5) and the base BGE-M3 (0.579 MRR@5). 
BGE-M3's strong performance is likely due to its use of three embedding types for relevance calculation, further enhanced by fine-tuning on in-domain data. 
Notably, the auto-reranked version nearly matches human-reranked performance without requiring costly human annotation. 

\textbf{Based on these findings, we recommend a cost-effective in-domain adaptation pipeline, notably Auto-Finetuned BGE-M3, that uses a strong LLM to generate synthetic training pairs, retrieves top-k passages with BGE-M3, and then applies a BGE-M3 Reranker. 
As shown in the results, this approach closely matches human-reranked performance while significantly saving annotation costs in an in-domain setup.
}

The commercially available Cohere embedding model ranks just below the top-performing BGE-M3 models and is followed by the ColBERT-based and dense embedding models, JINA ColBERT v2 and JINA embeddings v3, respectively. 
Among the tested retrievers, NV-Embed v1 shows the lowest performance among non-baseline models (0.713 MRR@5), likely due to its decoder-based architecture and reliance on prefix instruction prompts. 
Overall, retrieval performance on NitiBench-CCL is strong, with most models delivering comparable results, except for NV-Embed v1 and BM25. 
However, despite this strong performance, a gap between hit-rate and MRR when $k=\{5,10\}$ indicates that \textbf{while relevant documents are frequently retrieved, they are not consistently ranked first, potentially impacting end-to-end performance.}

\begin{table}[!ht]
\centering

\small
\begin{tabular}{@{}clcc@{}}
\toprule
\textbf{Top-K} & \textbf{Model} & \textbf{HR/Recall@k} & \textbf{MRR@k} \\ \midrule
\multirow{8}{*}{k=1} 
  & BM25                   & .481 & .321  \\
  & JINA V2                & .681 & .473 \\
  & JINA V3                & .587 & .412 \\
  & NV-Embed V1            & .492 & .334 \\
  & BGE-M3                 & .700 & .491 \\
  & Human-Finetuned BGE-M3 & \textbf{.735} & \textbf{.516} \\
  & Auto-Finetuned BGE-M3  & \underline{.731} & \underline{.513} \\
  & Cohere                 & .676 & .473 \\ \midrule
\multirow{8}{*}{k=5} 
  & BM25                   & .658 & .519 \\
  & JINA V2                & .852 & .753 \\
  & JINA V3                & .821 & .722 \\
  & NV-Embed V1            & .713 & .560 \\
  & BGE-M3                 & .880 & .824 \\
  & Human-Finetuned BGE-M3 & \textbf{.906} & \textbf{.850}  \\
  & Auto-Finetuned BGE-M3  & \underline{.900} & \underline{.840}  \\
  & Cohere                 & .870 & .779  \\ \midrule
\multirow{8}{*}{k=10} 
  & BM25                   & .715 & .600  \\
  & JINA V2                & .889 & .831  \\
  & JINA V3                & .875 & .817  \\
  & NV-Embed V1            & .776 & .640  \\
  & BGE-M3                 & .919 & .899  \\
  & Human-Finetuned BGE-M3 & \textbf{.938} & \textbf{.920}  \\
  & Auto-Finetuned BGE-M3  & \underline{.934} & \underline{.913}  \\
  & Cohere                 & .912 & .862  \\ \bottomrule
\end{tabular}
\caption{Retrieval Evaluation Result on NitiBench-CCL with hierarchy-aware chunking. Since the test split contains a single positive (as mentioned in \S \ref{subsubsec: wcx_dataset}), we collapsed metrics that are duplicated, such as HitRate (HR)/Multi-HitRate/Recall and MultiMRR / MRR.}
\label{table: retrieval_wangchan}
\end{table}

\textbf{NitiBench-Tax} Table~\ref{table: retrieval_tax} presents the retrieval performance of various models on NitiBench-Tax using hierarchy-aware chunking. 
Unlike NitiBench-CCL, this dataset includes multi-label queries, resulting in different values for single-label and multi-label metrics.

Overall performance is significantly lower on this dataset compared to NitiBench-CCL, likely due to the considerably longer and more nuanced queries in NitiBench-Tax.
JINA v3 and BGE-M3 (base, auto-fine-tuned, and human-finetuned) consistently perform among the top, achieving Multi-MRR@10 scores of 0.311, 0.354, 0.345, and 0.333, respectively. 
Conversely, JINA v2 and NV-Embed v1 consistently underperform compared to the baseline, potentially because NitiBench-Tax is out-of-distribution relative to their training data, considering the complexity of the query.
This is particularly evident with JINA v2, whose Multi-MRR@10 drops dramatically from 0.750 on NitiBench-CCL to 0.091 on NitiBench-Tax.

Similarly, the Human-Finetuned BGE-M3 variants are often outperformed by the base BGE-M3, suggesting different data distributions between NitiBench-CCL and NitiBench-Tax, hindering cross-dataset generalization. 
While some models achieve reasonable single-label hit rates, multi-label hit-rate performance is poor across all models. 
This, combined with low recall and significantly lower multi-label MRR compared to single-label MRR, indicates that while models can often retrieve some relevant documents, they struggle to retrieve all relevant documents for a given query. 
This limitation is critical, as comprehensive legal responses require consideration of all relevant legal sections. 
\textbf{Although the proposed pipeline (Human-Finetuned BGE-M3) performs strongly on in-domain data (as seen with NitiBench-CCL), these Tax Case results underscore the critical need for sufficiently diverse in-domain training data, since a narrow domain distribution can lead to inconsistent or contradictory outcomes in real-world settings.}

Despite its lower overall performance, NitiBench-Tax benefits more from increasing the number of top-k retrieved documents compared to NitiBench-CCL. 
Its hit rate and recall improve at a faster rate as more documents are retrieved compared to NitiBench-CCL.

\begin{table}[!ht]
\centering
\small
\begin{tabular}{@{}clccccc@{}}
\toprule
\textbf{Top-K} & \textbf{Model}                  & \textbf{HR@k}          & \textbf{Multi HR@k}    & \textbf{Recall@k}      & \textbf{MRR@k}         & \textbf{Multi MRR@k}   \\ \midrule
k=1   & BM25                   & .220          & .080          & .070          & .220          & .118          \\
      & JINA V2                & .140          & .040          & .035          & .140          & .068          \\
      & JINA V3                & .400          & .100          & .134          & .400          & .203          \\
      & NV-Embed V1            & .100          & .020          & .028          & .100          & .035          \\
      & BGE-M3                 & \underline{.500}    & \underline{.140}    & \underline{.176}    & \underline{.500}    & \underline{.269}    \\
      & Human-Finetuned BGE-M3 & .480          & \underline{.140}    & \underline{.176}          & .480          & .255          \\
      & Auto-Finetuned BGE-M3  & \textbf{.520} & \textbf{.160} & \textbf{.190} & \textbf{.520} & \textbf{.281} \\
      & Cohere                 & .340          & .100          & .127          & .340          & .179          \\ \midrule
k=5   & BM25                   & .480          & .120          & .211          & .318          & .171          \\
      & JINA V2                & .200          & .080          & .070          & .165          & .085          \\
      & JINA V3                & \underline{.720}    & .278 & \textbf{.324} & .508          & .297          \\
      & NV-Embed V1            & .200          & .020          & .077          & .126          & .050          \\
      & BGE-M3                 & \underline{.720}    & \textbf{.294}    & \textbf{.338}    & \underline{.580}    & \textbf{.337} \\
      & Human-Finetuned BGE-M3 & \textbf{.740} & .220          & \underline{.331}          & .565          & .320          \\
      & Auto-Finetuned BGE-M3  & .700          & .200          & .310          & \textbf{.587} & \underline{.329}    \\
      & Cohere                 & .620          & .200          & .268          & .447          & .256          \\ \midrule
k=10  & BM25                   & .540          & .160          & .282          & .327          & .183          \\
      & JINA V2                & .240          & .100          & .099          & .171          & .091          \\
      & JINA V3                & \textbf{.840} & \underline{.340}    & \underline{.444}    & .524          & .311          \\
      & NV-Embed V1            & .220          & .040          & .085          & .128          & .052          \\
      & BGE-M3                 & \underline{.820}    & \textbf{.360} & \textbf{.472} & \underline{.593}    & \textbf{.354} \\
      & Human-Finetuned BGE-M3 & .800          & .280          & .437          & .574          & .333          \\
      & Auto-Finetuned BGE-M3  & .780          & .260          & .423          & \textbf{.600} & \underline{.345}    \\
      & Cohere                 & .680          & .200          & .352          & .454          & .263          \\ \bottomrule
\end{tabular}
\caption{Retrieval Evaluation Result on NitiBench-Tax with hierarchy-aware chunking. This split contains multiple positives per question.}
\label{table: retrieval_tax}
\end{table}

\subsubsection{LLM}
\label{subsubsec: llm_result}

The evaluation results of the experiments described in \S\ref{subsubsec: llm_setup} are presented in Table~\ref{table: llm_e2e_main_wcx} for NitiBench-CCL and Table~\ref{table: llm_e2e_main_tax} for NitiBench-Tax. 
Since experiments in \S\ref{subsubsec: referencer_result} do not provide conclusive results on whether the inclusion of NitiLink is necessary, we also vary the inclusion of NitiLink in this experiment as well.

\begin{table}[!ht]
\centering
\renewcommand{\arraystretch}{1.2}
\resizebox{\textwidth}{!}{%
\begin{tabular}{@{}lcccccccc@{}}
\toprule
\textbf{Setting} & \textbf{NitiLink} & \textbf{Retriever MRR ($\uparrow$)} & \textbf{Retriever Recall ($\uparrow$)} & \textbf{E2E Recall ($\uparrow$)} & \textbf{E2E Precision ($\uparrow$)} & \textbf{E2E F1 ($\uparrow$)} & \textbf{Coverage ($\uparrow$)} & \textbf{Contradiction ($\downarrow$)} \\ \midrule
\multirow{3}{*}{\texttt{gpt-4o-2024-08-06}} 
& No Ref      & \multirow{3}{*}{0.809} & \multirow{3}{*}{0.938} & 0.880  & \textbf{0.601}  & \textbf{0.714}  & 85.2  & 0.055  \\
& \cellcolor{lightgray}Ref Depth 1 &                      &                     & \cellcolor{lightgray}0.885  & \cellcolor{lightgray}\underline{0.579}  & \cellcolor{lightgray}\underline{0.700}  & \cellcolor{lightgray}86.3  & \cellcolor{lightgray}0.051  \\
& $\Delta$    &                      &                     & \textcolor{darkgreen}{+0.6\%}  & \textcolor{red}{-3.7\%}  & \textcolor{red}{-2.0\%}  & \textcolor{darkgreen}{+1.3\%}  & \textcolor{darkgreen}{-7.3\%}  \\ \midrule
\multirow{3}{*}{\texttt{gemini-1.5-pro-002}} 
& No Ref      & \multirow{3}{*}{0.809} & \multirow{3}{*}{0.938} & 0.892  & 0.512  & 0.651  & 86.5  & 0.048  \\
& \cellcolor{lightgray}Ref Depth 1 &                      &                     & \cellcolor{lightgray}\underline{0.895}  & \cellcolor{lightgray}0.491  & \cellcolor{lightgray}0.634  & \cellcolor{lightgray}87.3  & \cellcolor{lightgray}\underline{0.042}  \\
& $\Delta$    &                      &                     & \textcolor{darkgreen}{+0.3\%}  & \textcolor{red}{-4.1\%}  & \textcolor{red}{-2.6\%}  & \textcolor{darkgreen}{+0.9\%}  & \textcolor{darkgreen}{-12.5\%} \\ \midrule
\multirow{3}{*}{\texttt{claude-3-5-sonnet-20240620}} 
& No Ref      & \multirow{3}{*}{0.809} & \multirow{3}{*}{0.938} & \textbf{0.901} & 0.444  & 0.595  & \textbf{89.7} & \textbf{0.040}  \\ 
& \cellcolor{lightgray}Ref Depth 1 &                      &                     & \cellcolor{lightgray}0.894  & \cellcolor{lightgray}0.443  & \cellcolor{lightgray}0.592  & \cellcolor{lightgray}\underline{89.5} & \cellcolor{lightgray}0.044  \\
& $\Delta$    &                      &                     & \textcolor{red}{-0.8\%} & \textcolor{red}{-0.2\%} & \textcolor{red}{-0.5\%}  & \textcolor{red}{-0.2\%}  & \textcolor{red}{+10.0\%}  \\ \midrule
\multirow{3}{*}{\texttt{typhoon-v2-70b-instruct}} 
& No Ref      & \multirow{3}{*}{0.809} & \multirow{3}{*}{0.938} & 0.862  & 0.537  & 0.662  & 81.2  & 0.076  \\
& \cellcolor{lightgray}Ref Depth 1 &                      &                     & \cellcolor{lightgray}0.845  & \cellcolor{lightgray}0.573  & \cellcolor{lightgray}0.683  & \cellcolor{lightgray}79.9  & \cellcolor{lightgray}0.080  \\
& $\Delta$    &                      &                     & \textcolor{red}{-2.0\%} & \textcolor{darkgreen}{+6.7\%} & \textcolor{darkgreen}{+3.2\%}  & \textcolor{red}{-1.6\%}  & \textcolor{red}{+5.3\%}  \\ \midrule
\multirow{3}{*}{\texttt{typhoon-v2-8b-instruct}} 
& No Ref      & \multirow{3}{*}{0.809} & \multirow{3}{*}{0.938} & 0.775  & 0.387  & 0.516  & 70.8  & 0.134  \\
& \cellcolor{lightgray}Ref Depth 1 &                      &                     & \cellcolor{lightgray}0.718  & \cellcolor{lightgray}0.385  & \cellcolor{lightgray}0.501  & \cellcolor{lightgray}68.5  & \cellcolor{lightgray}0.145  \\
& $\Delta$    &                      &                     & \textcolor{red}{-7.4\%} & \textcolor{red}{-0.5\%}  & \textcolor{red}{-2.9\%}  & \textcolor{red}{-3.3\%}  & \textcolor{red}{+8.2\%}  \\ \bottomrule
\end{tabular}
}
\caption{Effect of LLM configuration on E2E performance on NitiBench-CCL. $\Delta$ values are computed relative to No Ref and normalized to percentage change.}
\label{table: llm_e2e_main_wcx}
\end{table}

\begin{table}[!ht]
\centering
\renewcommand{\arraystretch}{1.2}
\resizebox{\textwidth}{!}{%
\begin{tabular}{@{}lccccccccc@{}}
\toprule
{\textbf{Setting}} & {\textbf{NitiLink}} 
& {\textbf{Retriever MRR ($\uparrow$)}} 
& {\textbf{Retriever Multi MRR ($\uparrow$)}} 
& {\textbf{Retriever Recall ($\uparrow$)}} 
& {\textbf{E2E Recall ($\uparrow$)}} 
& {\textbf{E2E Precision ($\uparrow$)}} 
& {\textbf{E2E F1 ($\uparrow$)}} 
& {\textbf{Coverage ($\uparrow$)}} 
& {\textbf{Contradiction ($\downarrow$)}} \\ 
\midrule

\multirow{3}{*}{\texttt{gpt-4o-2024-08-06}} 
& No Ref            
  & \multirow{3}{*}{0.574} 
  & \multirow{3}{*}{0.333}  
  & \multirow{3}{*}{0.437}
  & 0.333 
  & \underline{0.640}
  & 0.438 
  & 50.0  
  & \underline{0.46} \\
& \cellcolor{lightgray}Ref Depth 1  
  & 
  & 
  & 
  & \cellcolor{lightgray}0.354 
  & \cellcolor{lightgray}0.630 
  & \cellcolor{lightgray}0.453 
  & \cellcolor{lightgray}45.0 
  & \cellcolor{lightgray}0.52 \\
& $\Delta$           
  & 
  & 
  & 
  & \textcolor{darkgreen}{+6.3\%}
  & \textcolor{red}{-1.6\%}
  & \textcolor{darkgreen}{+3.4\%}
  & \textcolor{red}{-10.0\%}
  & \textcolor{red}{+13.0\%} \\
\midrule

\multirow{3}{*}{\texttt{gemini-1.5-pro-002}} 
& No Ref            
  & \multirow{3}{*}{0.574} & \multirow{3}{*}{0.333} & \multirow{3}{*}{0.437}
  & 0.361 & 0.308 & 0.332 
  & 44.0  & 0.48 \\
& \cellcolor{lightgray}Ref Depth 1  
  & 
  & 
  & 
  & \cellcolor{lightgray}0.354 
  & \cellcolor{lightgray}0.347 
  & \cellcolor{lightgray}0.351 
  & \cellcolor{lightgray}45.0 
  & \cellcolor{lightgray}0.48 \\
& $\Delta$           
  & 
  & 
  & 
  & \textcolor{red}{-1.9\%}
  & \textcolor{darkgreen}{+12.7\%}
  & \textcolor{darkgreen}{+5.7\%}
  & \textcolor{darkgreen}{+2.3\%}
  & \textcolor{black}{+0.0\%} \\
\midrule

\multirow{3}{*}{\texttt{claude-3-5-sonnet-20240620}} 
& No Ref            
  & \multirow{3}{*}{0.574} & \multirow{3}{*}{0.333} & \multirow{3}{*}{0.437 }
  & \underline{0.389} & 0.554 & \underline{0.457}
  & \underline{51.0}  & \textbf{0.44} \\
& \cellcolor{lightgray}Ref Depth 1  
  & 
  & 
  & 
  & \cellcolor{lightgray}\textbf{0.417} 
  & \cellcolor{lightgray}0.577 
  & \cellcolor{lightgray}\textbf{0.484} 
  & \cellcolor{lightgray}49.0 
  & \cellcolor{lightgray}0.56 \\
& $\Delta$           
  & 
  & 
  & 
  & \textcolor{darkgreen}{+7.2\%}
  & \textcolor{darkgreen}{+4.2\%}
  & \textcolor{darkgreen}{+5.9\%}
  & \textcolor{red}{-3.9\%}
  & \textcolor{red}{+27.3\%} \\
\midrule

\multirow{3}{*}{\texttt{typhoon-v2-70b-instruct}} 
& No Ref            
  & \multirow{3}{*}{0.574} & \multirow{3}{*}{0.333} & \multirow{3}{*}{0.437}
  & 0.326 & \textbf{0.662} & 0.437 
  & 42.0  & 0.58 \\
& \cellcolor{lightgray}Ref Depth 1  
  & 
  & 
  &  
  & \cellcolor{lightgray}0.333 
  & \cellcolor{lightgray}0.453 
  & \cellcolor{lightgray}0.384 
  & \cellcolor{lightgray}\textbf{54.0 }
  & \cellcolor{lightgray}\underline{0.46} \\
& $\Delta$           
  & 
  & 
  & 
  & \textcolor{darkgreen}{+2.1\%}
  & \textcolor{red}{-31.6\%}
  & \textcolor{red}{-12.1\%}
  & \textcolor{darkgreen}{+28.6\%}
  & \textcolor{darkgreen}{-20.7\%} \\
\midrule

\multirow{3}{*}{\texttt{typhoon-v2-8b-instruct}} 
& No Ref            
  & \multirow{3}{*}{0.574} & \multirow{3}{*}{0.333} & \multirow{3}{*}{0.437 }
  & 0.278 & 0.471 & 0.349 
  & 37.0  & 0.60 \\
& \cellcolor{lightgray}Ref Depth 1  
  & 
  & 
  & 
  & \cellcolor{lightgray}0.319 
  & \cellcolor{lightgray}0.561 
  & \cellcolor{lightgray}0.407 
  & \cellcolor{lightgray}35.0 
  & \cellcolor{lightgray}0.54 \\
& $\Delta$           
  & 
  & 
  & 
  & \textcolor{darkgreen}{+14.7\%}
  & \textcolor{darkgreen}{+19.1\%}
  & \textcolor{darkgreen}{+16.6\%}
  & \textcolor{red}{-5.4\%}
  & \textcolor{darkgreen}{-10.0\%} \\
\bottomrule
\end{tabular}
}
\caption{Effect of LLM configuration on E2E performance on NitiBench-Tax. 
\(\Delta\) values are computed relative to No Ref and normalized to percentage change.}
\label{table: llm_e2e_main_tax}
\end{table}

\textbf{NitiBench-CCL} On NitiBench-CCL, \texttt{claude-3.5-sonnet} excels in Coverage, Contradiction, and E2E Recall. 
However, the typhoon family of models struggles to match the performance of the closed-source models in this broader Thai legal QA domain. 
As seen with NitiBench-Tax, both \texttt{claude-3.5-sonnet} and \texttt{gemini-1.5-pro-002} exhibit low E2E Precision, leading to lower F1 scores compared to \texttt{gpt-4o}. 
This underscores a trade-off between recall and precision, particularly for \texttt{claude-3.5-sonnet} and \texttt{gemini-1.5-pro-002}. 
The causes of this precision drop are further analyzed in a later section.

\textbf{NitiBench-Tax} The results on NitiBench-Tax demonstrate that \texttt{claude-3.5-sonnet} also achieves top-2 performance across most end-to-end metrics. 
Interestingly, \texttt{typhoon-v2-70b-instruct}, an open-sourced model, delivers comparable results and outperforms others on NitiBench-Tax with the highest coverage score and E2E precision. 
However, its smaller variant, \texttt{typhoon-v2-8b-instruct}, ranks the lowest among the tested models.f
Despite its limited parameter size, it manages to avoid falling significantly behind, showcasing a reasonable performance given its constraints.

Notably, both \texttt{claude-3.5-sonnet} and \texttt{gemini-1.5-pro-002} exhibit considerably lower E2E precision compared to \texttt{gpt-4o} and \texttt{typhoon-v2-70b-instruct}. This compromises their suitability for precision-critical applications, even though they excel in other areas. Additionally, the NitiLink module fails to consistently enhance performance, with mixed results indicating no definitive advantage in its current configuration.

\subsubsection{E2E results using best setups}
\label{subsubsec: e2e_best_result}

The results from experiments conducted under the four settings described in \S\ref{subsubsec: e2e_best_setup} are presented in Table~\ref{table: main_exp_full}. 
Both the Proposed RAG and Naive RAG settings utilize Human-Finetuned BGE-M3 as the retriever, as it demonstrated the best performance in previous experiments (Section~\ref{subsubsec: retriever_result}). 
Similarly, Claude 3.5 Sonnet is chosen as the generator based on its superior results in \S\ref{subsubsec: llm_result}. 
NitiLink is excluded in both settings due to its inconclusive impact on E2E performance (Section~\ref{subsubsec: referencer_result} and~\ref{subsubsec: llm_result}). 
We opted for this choice instead of the opposite because omitting NitiLink yields similar performance while reducing API costs. 
The primary distinction between Naive RAG and Proposed RAG lies in their chunking strategies: Naive RAG employs a naive chunking approach, whereas Proposed RAG utilizes hierarchy-aware chunking.

\begin{table}[!ht]
\centering
\renewcommand{\arraystretch}{1.2} 
\resizebox{\textwidth}{!}{%
\begin{tabular}{@{}lcccccccc@{}}
\toprule
{\textbf{Setting}} & \textbf{{Retriever MRR} ($\uparrow$)} & \textbf{{Retriever Multi-MRR} ($\uparrow$)} & \textbf{{Retriever Recall} ($\uparrow$)} & \textbf{{Coverage} ($\uparrow$)} & \textbf{{Contradiction} ($\downarrow$)} & \textbf{{E2E Recall} ($\uparrow$)} & \textbf{{E2E Precision} ($\uparrow$)} & \textbf{{E2E F1} ($\uparrow$)} \\ 
\midrule
\multicolumn{9}{c}{\textbf{NitiBench-CCL}} \\ \midrule
Parametric     
 & -     
 & -     
 & -     
 & 60.3     
 & 0.199   
 & 0.188   
 & 0.141   
 & 0.161  
\\
Na\"ive RAG    
 & 0.553 
 & 0.553 
 & 0.658 
 & 77.3     
 & 0.097   
 & 0.745   
 & 0.370   
 & 0.495  
\\
Proposed RAG   
 & 0.809 
 & 0.809 
 & 0.938 
 & \textbf{89.7}     
 & \textbf{0.040}   
 & \textbf{0.901}   
 & \textbf{0.444}   
 & \textbf{0.595}  
\\
\rowcolor{gray!15}
Golden Context 
 & 1.0   
 & 1.0   
 & 1.0   
 & 93.4     
 & 0.034   
 & 0.999   
 & 1.000   
 & 1.000  
\\ 
\midrule
\multicolumn{9}{c}{\textbf{NitiBench-Tax}} \\ \midrule
Parametric     
 & -     
 & -     
 & -     
 & 46.0     
 & 0.480   
 & \textbf{0.458}   
 & \textbf{0.629}   
 & \textbf{0.530}  
\\
Na\"ive RAG    
 & 0.120 
 & 0.048 
 & 0.077 
 & 50.0     
 & 0.460   
 & 0.306   
 & 0.463   
 & 0.368  
\\
Proposed RAG   
 & 0.574 
 & 0.333 
 & 0.437 
 & \textbf{51.0}     
 & \textbf{0.440}   
 & {0.389}   
 & {0.554}   
 & {0.457}  
\\
\rowcolor{gray!15}
Golden Context 
 & 1.0   
 & 1.0   
 & 1.0   
 & 52.0     
 & 0.460   
 & 0.694   
 & 1.000   
 & 0.820  
\\ 
\bottomrule
\end{tabular}
}
\caption{E2E Experiment results on NitiBench-CCL and NitiBench-Tax comparing various RAG setups on Human-Finetuned BGE-M3 retriever and \texttt{claude-3-5-sonnet-20240620} as a LLM.}
\label{table: main_exp_full}
\end{table}

\textbf{NitiBench-Tax} On NitiBench-Tax, the four main settings perform similarly, except for the parametric setting's slightly lower coverage and higher contradiction. Two key observations emerge:

First, the parametric setting achieves the second-highest E2E recall and precision despite lacking a retriever. 
To investigate this further, we inspected the cited law section generated by LLM. 
Surprisingly, we found that out of 105 law sections cited from LLM parametric knowledge, 58 of them aren't even retrieved by the best retriever. 
Among those 58 cited documents, 26 were in the correct law section. 
In contrast, only 5 of 101 sections cited by the proposed RAG system are \emph{not} retrieved. 
This indicates that retriever performance significantly constrains RAG systems, especially with complex queries like those in NitiBench-Tax. 
RAG system generators seem discouraged from using internal knowledge, which might sometimes provide better answers. 
Furthermore, the substantial disparity between retriever and E2E recall shows that the LLM often underutilizes relevant retrieved sections, particularly those containing primarily terminology (as discussed in \S\ref{subsec: retriever_re_error_analysis_tax}).
%
%

Second, Table~\ref{table: main_exp_full} (NitiBench-Tax) shows no clear relationship between E2E citation scores and coverage/contradiction. 
This suggests the LLM struggles to apply cited sections correctly in its reasoning, leading to incorrect or erroneously reasoned answers. 
This resembles the issue in \S\ref{subsubsec: referencer_result}, where improved retriever recall from NitiLink didn't consistently improve E2E metrics. 
Here, increased E2E citations with Claude 3.5 Sonnet don't necessarily improve coverage or contradiction.

\textbf{NitiBench-CCL} For NitiBench-CCL (Table~\ref{table: main_exp_full}), the results are as expected: parametric performs worst, followed by Naive RAG, then Proposed RAG, and finally Golden Context (best).

\subsection{(RQ3) Performance of Long-Context LLM (LCLM)}
\label{subsec: rq3_result}

The evaluation results of the Thai legal QA system based on LCLM are presented in Table~\ref{table: main_exp_sampled} in comparison to the same baselines in Table~\ref{table: main_exp_full}. 
The evaluation is conducted on a stratified 20\% sample of NitiBench-CCL and the full NitiBench-Tax. 
As detailed in \S\ref{subsec: setup_rq3}, the LCLM system processes all 35 Thai financial laws simultaneously as context, with special tokens inserted to serve as section identifiers. 
These tokens enable the LCLM to cite relevant sections explicitly when generating responses to queries.

\begin{table}[!ht]
\centering
\renewcommand{\arraystretch}{1.2} 
\resizebox{\textwidth}{!}{%
\begin{tabular}{@{}lcccccccc@{}}
\toprule
\textbf{Setting} & \textbf{Retriever MRR ($\uparrow$)} & \textbf{Retriever Multi-MRR ($\uparrow$)} & \textbf{Retriever Recall ($\uparrow$)} & \textbf{Coverage ($\uparrow$)} & \textbf{Contradiction ($\downarrow$)} & \textbf{E2E Recall ($\uparrow$)} & \textbf{E2E Precision ($\uparrow$)} & \textbf{E2E F1 ($\uparrow$)} \\ 
\midrule
\multicolumn{9}{c}{\textbf{NitiBench-CCL (20\% subsampled)}} \\ \midrule
Parametric     & -     & -     & -     & 60.6  & 0.198 & 0.197 & 0.147 & 0.169 \\
Na\"ive RAG    & 0.549 & 0.549 & 0.649 & 77.7  & 0.092 & 0.740 & 0.379 & 0.501 \\
Proposed RAG   & 0.825 & 0.825 & 0.945 & \textbf{90.1}  & \textbf{0.028} & \textbf{0.920} & 0.453 & 0.607 \\
LCLM (Gemini)  & -     & -     & -     & 83.2  & 0.063 & 0.765 & \textbf{0.514} & \textbf{0.615} \\
\rowcolor{gray!15}
Golden Context & 1.0   & 1.0   & 1.0   & 94.2  & 0.025 & 0.999 & 1.000 & 0.999 \\
\midrule
\multicolumn{9}{c}{\textbf{NitiBench-Tax}} \\ \midrule
Parametric     & -     & -     & -     & 46.0  & 0.480 & \textbf{0.458} & \textbf{0.629} & \textbf{0.530} \\
Na\"ive RAG    & 0.120 & 0.048 & 0.077 & 50.0  & 0.460 & 0.306 & 0.463 & 0.368 \\
Proposed RAG   & 0.574 & 0.333 & 0.437 & \textbf{51.0}  & \textbf{0.440} & 0.389 & 0.554 & 0.457 \\
LCLM (Gemini)  & -     & -     & -     & 36.0  & 0.620 & 0.410 & 0.484 & 0.444 \\
\rowcolor{gray!15}
Golden Context & 1.0   & 1.0   & 1.0   & 52.0  & 0.460 & 0.694 & 1.000 & 0.820 \\ 
\bottomrule
\end{tabular}
}
\caption{Experiment results on sampled NitiBench-CCL and full NitiBench-Tax. In the LCLM setup, we used the Gemini without retriever section, where full legislation books were parsed as a context.}
\label{table: main_exp_sampled}
\end{table}

From Table~\ref{table: main_exp_sampled}, the LCLM-based system performs comparably to the parametric setting on NitiBench-Tax and to the Naive RAG system on NitiBench-CCL. 
This performance gap may stem from degradation when processing extremely long contexts (1.2 million tokens). 
Regardless of the exact cause, the results suggest that while an LCLM-based Thai legal QA system is feasible, its performance remains significantly behind RAG-based counterparts, highlighting areas for further improvement.

Apart from utilizing LCLM to process the legislations and respond directly to the queries, we also explored using it as a retriever. 
As stated in \S\ref{subsec: setup_rq3}, Gemini 1.5 Pro is provided with all 35 legislations and tasked to retrieve 20 relevant laws given a query. 
This experiment is also conducted on the same sample of NitiBench-CCL as the previous experiment and the full NitiBench-Tax. 
The results are shown in Table~\ref{table: retrieval_wcx_lclm} and~\ref{table: retrieval_tax_lclm}.

\begin{table}[!ht]
\centering
\small
\begin{tabular}{@{}clcc@{}}
\toprule
\textbf{Top-K} & \textbf{Model} & \textbf{HR/Recall@k} & \textbf{MRR@k} \\ \midrule
\multirow{9}{*}{k=1} 
  & BM25                   & .480           & .480 \\
  & JINA V2                & .698           & .698 \\
  & JINA V3                & .601           & .601 \\
  & NV-Embed V1            & .496           & .496 \\
  & BGE-M3                 & .708           & .708 \\
  & Human-Finetuned BGE-M3 & \textbf{.757}  & \textbf{.757} \\
  & Auto-Finetuned BGE-M3  & \underline{.741} & \underline{.741} \\
  & Cohere                 & .707           & .707 \\
  & LCLM-as-a-retriever (Gemini)                   & .590           & .590 \\ \midrule
\multirow{9}{*}{k=5} 
  & BM25                   & .663           & .549 \\
  & JINA V2                & .858           & .761 \\
  & JINA V3                & .828           & .693 \\
  & NV-Embed V1            & .711           & .585 \\
  & BGE-M3                 & \underline{.888} & .779 \\
  & Human-Finetuned BGE-M3 & \textbf{.909}  & \textbf{.819} \\
  & Auto-Finetuned BGE-M3  & \textbf{.909}  & \underline{.807} \\
  & Cohere                 & .867           & .772 \\
  & LCLM-as-a-retriever (Gemini)                   & .776           & .667 \\ \midrule
\multirow{9}{*}{k=10} 
  & BM25                   & .733           & .559 \\
  & JINA V2                & .891           & .766 \\
  & JINA V3                & .878           & .700 \\
  & NV-Embed V1            & .794           & .596 \\
  & BGE-M3                 & .926           & .784 \\
  & Human-Finetuned BGE-M3 & \textbf{.945}  & \textbf{.824} \\
  & Auto-Finetuned BGE-M3  & \underline{.941} & \underline{.812} \\
  & Cohere                 & .913           & .778 \\
  & LCLM-as-a-retriever (Gemini)                   & .807           & .671 \\ \bottomrule
\end{tabular}
\caption{Retrieval Evaluation Result on a 20\% subset of NitiBench-CCL with hierarchy-aware chunking with Long-Context Retriever. Since the test split of NitiBench-CCL is single labeled, duplicated metrics (HR/Recall and MRR) have been collapsed.}
\label{table: retrieval_wcx_lclm}
\end{table}

\begin{table}[!ht]
\centering
\small
\begin{tabular}{@{}clccccc@{}}
\toprule
\textbf{Top-K} & \textbf{Model}                  & \textbf{HR@k}          & \textbf{Multi HR@k}    & \textbf{Recall@k}      & \textbf{MRR@k}         & \textbf{Multi MRR@k}   \\ \midrule
k=1   & BM25                   & .220          & .080          & .070          & .220          & .118          \\
      & JINA V2                & .140          & .040          & .035          & .140          & .068          \\
      & JINA V3                & .400          & .100          & .134          & .400          & .203          \\
      & NV-Embed V1            & .100          & .020          & .028          & .100          & .035          \\
      & BGE-M3                 & \underline{.500}    & \underline{.140}    & \underline{.176}    & \underline{.500}    & \underline{.269}    \\
      & Human-Finetuned BGE-M3 & .480          & \underline{.140}    & \underline{.176}          & .480          & .255          \\
      & Auto-Finetuned BGE-M3  & \textbf{.520} & \textbf{.160} & \textbf{.190} & \textbf{.520} & \textbf{.281} \\
      & Cohere                 & .340          & .100          & .127          & .340          & .179          \\
      & LCLM                   & .480          & .120          & .164          & .480          & .227          \\ \midrule
k=5   & BM25                   & .480          & .120          & .211          & .318          & .171          \\
      & JINA V2                & .200          & .080          & .070          & .165          & .085          \\
      & JINA V3                & .720          & \underline{.260}    & .324    & .508          & .297          \\
      & NV-Embed V1            & .200          & .020          & .077          & .126          & .050          \\
      & BGE-M3                 & .720          & .240          & \underline{.338}          & \underline{.580}    & \underline{.337}    \\
      & Human-Finetuned BGE-M3 & \underline{.740}    & .220          & .331          & .565          & .320          \\
      & Auto-Finetuned BGE-M3  & .700          & .200          & .310          & \textbf{.587} & \underline{.329}    \\
      & Cohere                 & .620          & .200          & .268          & .447          & .256          \\
      & LCLM-as-a-retriever (Gemini)                   & \textbf{.760} & \textbf{.320} & \textbf{.418} & \textbf{.587} & \textbf{.370} \\ \midrule
k=10  & BM25                   & .540          & .160          & .282          & .327          & .183          \\
      & JINA V2                & .240          & .100          & .099          & .171          & .091          \\
      & JINA V3                & \textbf{.840} & \underline{.340}    & .444          & .524          & .311          \\
      & NV-Embed V1            & .220          & .040          & .085          & .128          & .052          \\
      & BGE-M3                 & \underline{.820}    & \textbf{.360} & \underline{.472}    & \underline{.593}    & \underline{.354}    \\
      & Human-Finetuned BGE-M3 & .800          & .280          & .437          & .574          & .333          \\
      & Auto-Finetuned BGE-M3  & .780          & .260          & .423          & \textbf{.600} & \underline{.345}    \\
      & Cohere                 & .680          & .200          & .352          & .454          & .263          \\
      & LCLM-as-a-retriever (Gemini)                   & .780          & \textbf{.360} & \textbf{.477} & .590          & \textbf{.379} \\ \bottomrule
\end{tabular}
\caption{Retrieval Evaluation Result on NitiBench-Tax with hierarchy-aware chunking. This split contains multiple positives per question.}
\label{table: retrieval_tax_lclm}
\end{table}

The results indicate that the LCLM retriever performs comparably to embedding-based retrievers on NitiBench-Tax, likely due to its superior reasoning capabilities. 
However, a noticeable performance gap exists when compared to the best retriever on NitiBench-CCL.
Additionally, increasing the number of retrieved documents for the LCLM yields minimal performance improvements relative to other models. 
We hypothesize that this limited gain is a result of LLMs' next-token prediction mechanism, which may hinder their ability to effectively retrieve and output relevant laws when those laws are distant from the query context or when the model attempts to generate relevant laws ranked lower in the retrieval order.

\section{Discussion}

\subsection{Analysis of Effect of Chunking Strategy on E2E and Retrieval Performance}
\label{subsec: chunk_error_analysis}

\begin{figure}[h!]
\centering

\includegraphics[width=0.6\textwidth]{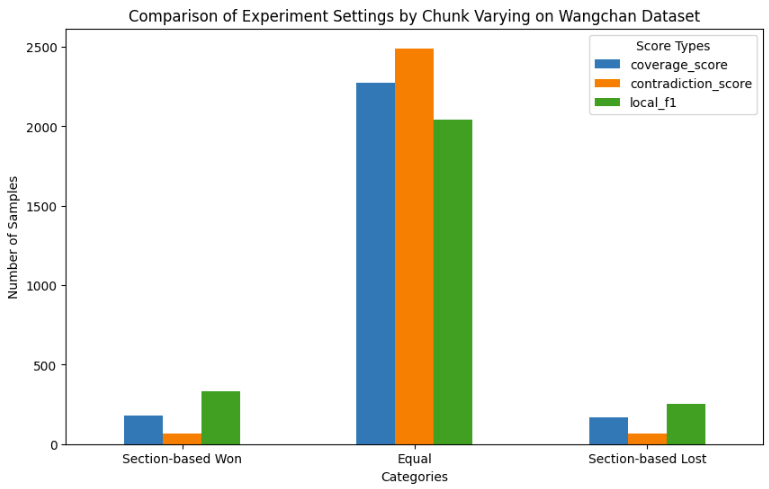}  
\caption{Bar chart showing amount of queries performing better under hierarchical-aware and naive chunking on NitiBench-CCL}
\label{fig:chunk_vary_won_lost_wcx}
\end{figure}

Building on the results in Table~\ref{table: chunk_e2e_main}, we further analyze the number of samples where each chunking strategy outperforms the other, as shown in Figure~\ref{fig:chunk_vary_won_lost_wcx}, where \texttt{local\_f1} represents the E2E F1 score.
The performance difference between the two strategies remains relatively balanced in terms of coverage score.
However, hierarchy-aware chunking achieves better E2E F1 scores, while naive chunking performs slightly better in terms of contradiction scores.

For further investigation, we analyze the subset of queries where the naive chunking method achieves a higher coverage score than hierarchy-aware chunking. 
Within this subset, we further examine queries where the MRR of naive chunking exceeds that of hierarchy-aware chunking. 
This specific focus allows us to identify cases where naive chunking enhances retriever performance and leads to better overall E2E performance despite our assumption that hierarchy-aware chunking inherently provides the most complete and noise-free information for each legislative section.

Upon analyzing this subset of queries, a recurring pattern emerges. 
\textbf{Thai legal QA system benefits from the naive chunking strategy when the target section alone does not provide sufficient context and requires context from the section before or after it.}  

For example, a query regarding the endorsement of a bill of exchange without a date requires the application of Section 933 of the Civil and Commercial Code, which discusses the acceptance of a bill of exchange. 
However, this section does not explicitly mention the term \enquote{bill of exchange}, as it assumes the reader's familiarity with the term from preceding sections.  

By chunking \enquote{imperfectly}, the naive chunking strategy includes Section 932— which explicitly references a bill of exchange— alongside Section 933.
This additional context improves the retriever model's ability to retrieve the correct chunk.

\begin{quote}
\textbf{Question: }If the drawer endorses a bill of exchange but does not include a date, what legal actions can be taken?

\textbf{Relevant Chunk: } 

   \textbf{Section 933 of Civil and Commercial Code:} If an acceptance does not include a date, the last day of the prescribed period for acceptance shall be deemed the date of acceptance.

\textbf{Chunk retrieved under naive chunking setting: }

   \textbf{Section 932 of Civil and Commercial Code:} It is prescribed that if the holder acts in good faith but records an incorrect date due to a material mistake, or if the date is entirely incorrect in any case, the bill of exchange shall not be invalidated solely for this reason, provided that it later comes into the possession of a lawful holder. The bill shall remain valid and enforceable as if the recorded date were correct.

   \textbf{Section 933 of Civil and Commercial Code:} If an acceptance does not include a date, the last day of the prescribed period for acceptance shall be deemed the date of acceptance.

   \textbf{Section 934 of Civil and Commercial Code}
\end{quote}



\subsection{Effect of NitiLink's Reference Depth on Retrieval Performance}
\label{subsec: ref_depth_vs_retrieval_perf}

\textbf{Adding more reference depth improves retrieval performance when the question requires extensive legal reasoning.} 
In Table~\ref{table: augmenter_e2e_main}, we observed that increasing the reference depth by one slightly improved retrieval performance for NitiBench-Tax, while NitiBench-CCL showed no notable difference. 
This suggests that for NitiBench-CCL, which does not require extensive legal reasoning, resolving law section references provides little benefit. 
However, it remains unclear how much retrieval performance—particularly recall—can be improved by further increasing the reference depth.

\begin{figure}[!ht]
\centering
\includegraphics[width=0.7\textwidth]{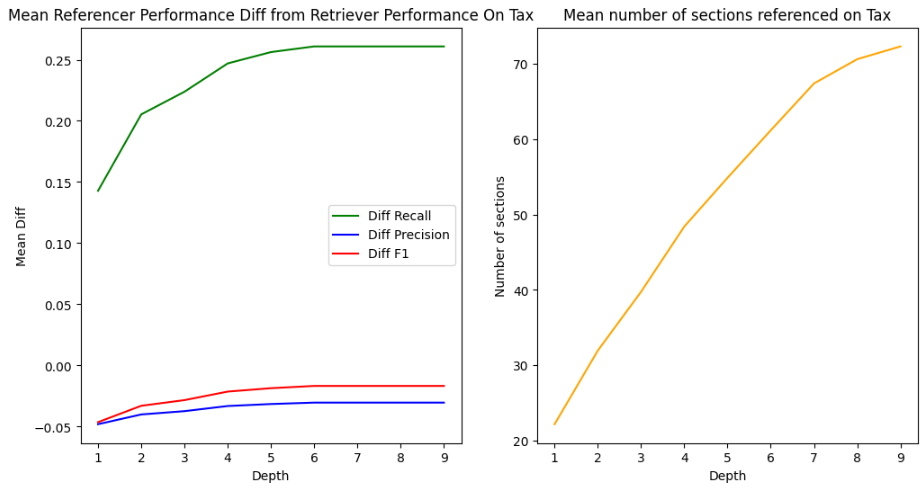}
\caption{Plot of the relationship between reference depth of NitiLink and retrieval performance and number of sections per query on NitiBench-Tax. On the left, \texttt{Mean Diff} shows the average retrieval metric difference when increasing law section depth compared to retrieval performance without NitiLink. The right plot shows the number of sections cited when resolving more reference depth.}
\label{fig: augmenter_vary_retrieval_diff_tax}
\end{figure}

\begin{figure}[!ht]
\centering
\includegraphics[width=0.7\textwidth]{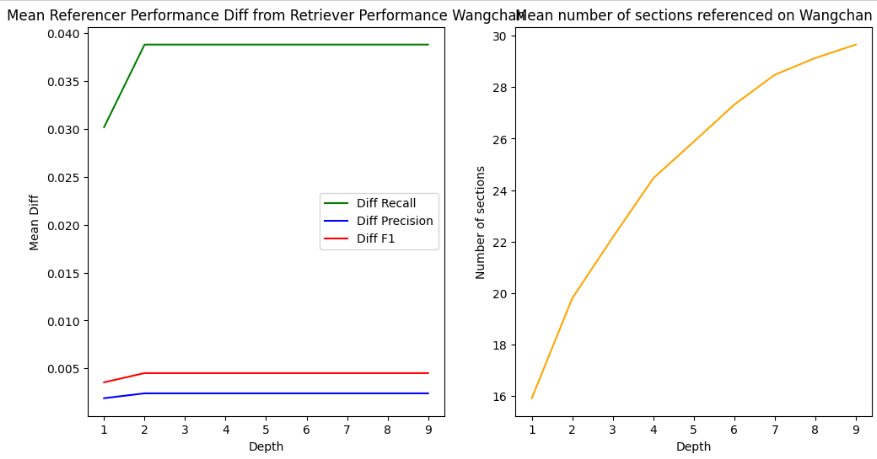}
\caption{Plot of the relationship between reference depth of NitiLink and retrieval performance and number of sections per query on NitiBench-CCL.}
\label{fig: augmenter_vary_retrieval_diff_wcx}
\end{figure}

To investigate this, we examined the relationship between \textit{NitiLink's maximum reference depth depth}, \textit{retrieval performance gains} (\texttt{Mean Diff} on the y-axis), and \textit{the total number of law sections NitiLink resolves} (see Figures~\ref{fig: augmenter_vary_retrieval_diff_tax} and~\ref{fig: augmenter_vary_retrieval_diff_wcx}). 

For NitiBench-Tax, retrieval performance improves as reference depth increases, peaking at a depth of 6.
However, this comes at the cost of increased context length, reaching approximately 60 sections per query.
While the improvement in retrieval performance could be attributed to retrieving more law sections—thereby increasing the hit rate—after extensive recursive reference resolution in NitiBench-Tax, the results for NitiBench-CCL indicate that this is not always the case. 
For the NitiBench-CCL, retrieval gains remain minimal and plateau after a depth of 2 despite resolving up to 30 law sections at a depth of 9. 
We suspect this is due to NitiBench-CCL requiring only one relevant law per entry, eliminating the need for complex legal reasoning during retrieval.

Based on this analysis, we hypothesize that the performance gains from increasing reference depth stem from \textbf{recursively resolving law section references, which primarily benefits retrieval performance when the query requires extensive legal reasoning.}


\subsection{On Generated Answer Performance When Using NitiLink}
\label{subsec: generated_ans_perf_with_ref}

\begin{figure}[!ht]
    \centering
    \begin{subfigure}[b]{0.48\textwidth}
        \centering
        \includegraphics[width=\textwidth]{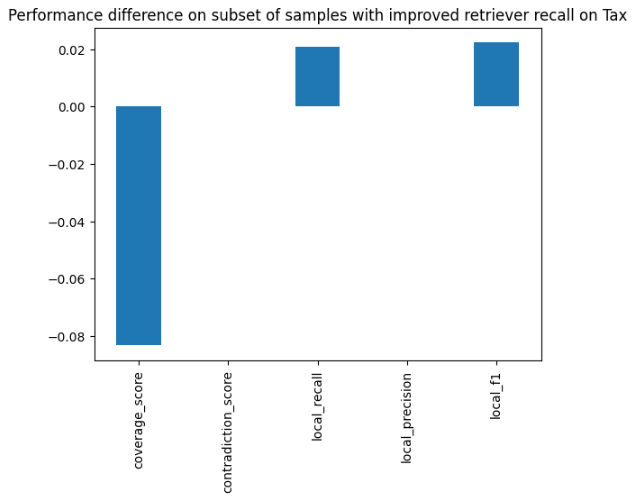}
        \label{fig:augmenter_vary_e2e_gain_tax}
    \end{subfigure}
    \hfill
    \begin{subfigure}[b]{0.48\textwidth}
        \centering
        \includegraphics[width=\textwidth]{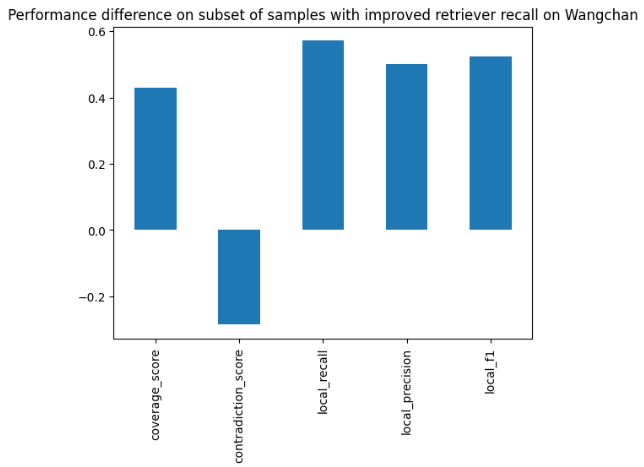}
        \label{fig:augmenter_vary_e2e_gain_wcx}
    \end{subfigure}
    \caption{Difference in E2E metrics on NitiBench-Tax (left) and on NitiBench-CCL (right). The Y-axis of the plot denotes the difference in E2E metrics with NitiLink compared to those without using NitiLink. A positive value means that NitiLink improves that specific metric (specified in x-axis) and negative showed performance degradation. \texttt{local} metric represents the citation score obtained from LLM citation in the generated output.}
    \label{fig:augmenter_vary_e2e_gain}
\end{figure}

\textbf{An increase in retrieval performance from using NitiLink does not always translate to a better generation score.} 
Based on the insights from \S\ref{subsec: ref_depth_vs_retrieval_perf}, we hypothesized that NitiLink could boost retrieval performance—especially in NitiBench-Tax, where extensive legal reasoning is required. 
To further investigate the effect of adding NitiLink on end-to-end (E2E) metrics (e.g., coverage, contradiction), we analyzed the differences in E2E metrics with and without NitiLink on subsets of queries for which NitiLink improved recall. 
This approach isolates our analysis to focus solely on the E2E performance of queries where NitiLink successfully identified the correct law section (unlike the analysis in \S\ref{subsubsec: referencer_result}, which includes all queries). 
The results are presented in Figures~\ref{fig:augmenter_vary_e2e_gain}. 
The local\_recall, local\_precision, and local\_f1 refer to E2E recall, precision, and f1, respectively.

These plots confirm the trend mentioned in \S\ref{subsubsec: referencer_result}: improved retrieval performance in NitiBench-Tax does not consistently translate to better end-to-end performance due to query complexity and increased context length. 
In contrast, retrieval improvements in NitiBench-CCL significantly enhance end-to-end results. 
From these findings, we hypothesize that \textbf{improvements in retrieval recall do not always translate to better E2E performance and may even degrade LLM generation performance. 
We suspect that this degradation is due to two reasons: 
1) NitiLink does not contribute to the LLM reasoning capability necessary for processing queries such as in NitiBench-Tax. 
2) The longer context lengths can have adverse effects on the LLM performance.}

\subsection{Findings from Varying Retrieval Model on Legal Question Answering}
\label{subsec: retriever_e2e_error_analysis}


We measure E2E metrics based on 4 retrievers setups where each retrieval model represents distinct types of embeddings: 
three-head embeddings (\textit{Human-Finetuned BGE-M3}), 
dense embeddings (\textit{JINA V3}), 
sparse embeddings (\textit{BM25}), and 
API-based embeddings for retrieval (\textit{Cohere}). 
We fix our chunking strategy to use a hierarchy-aware chunking method (see \S~\ref{subsubsec: chunk_setup}) without NitiLink and use \texttt{gpt-4o-2024-08-06} as a LLM for text generation. 
The results are summarized in Table~\ref{table: retrieval_e2e_main}.

\begin{table}[!ht]
\centering
\renewcommand{\arraystretch}{1.2} 
\resizebox{\textwidth}{!}{%
\begin{tabular}{@{}lcccccccc@{}}
\toprule
\textbf{Setting} & \textbf{Retriever MRR ($\uparrow$)} & \multicolumn{1}{l}{\textbf{Retriever Multi MRR ($\uparrow$)}} & \textbf{Retriever Recall ($\uparrow$)} & \textbf{Coverage ($\uparrow$)} & \textbf{Contradiction ($\downarrow$)} & \textbf{E2E Recall ($\uparrow$)} & \textbf{E2E Precision ($\uparrow$)} & \textbf{E2E F1 ($\uparrow$)} \\ \midrule

\multicolumn{9}{c}{\textbf{NitiBench-CCL}} \\ \midrule
BM25                   & 0.556                & {\color[HTML]{333333} 0.556} & 0.716            & 73.3          & 0.123          & 0.687          & 0.571          & 0.624          \\
JINA V3                & 0.689                & {\color[HTML]{333333} 0.689} & 0.875            & 82.1          & 0.065          & 0.827          & 0.592          & 0.690          \\
Cohere                 & \underline{\textbf{0.760}} & {\color[HTML]{333333} \underline{0.760}}  & \underline{0.912}      & \underline{84.4}    & \underline{0.060}    & \underline{0.857}    & \underline{0.597}    & \underline{0.704}    \\
Human-Finetuned BGE-M3 & \textbf{0.809}       & \textbf{0.809}         & \textbf{0.938}   & \textbf{85.2} & \textbf{0.055} & \textbf{0.880} & \textbf{0.601} & \textbf{0.714} \\

\midrule \multicolumn{9}{c}{\textbf{NitiBench-Tax}} \\ \midrule
BM25                   & 0.327                & 0.183 & 0.282            & 36.0          & 0.600          & 0.271          & 0.582          & 0.370          \\
JINA V3                & \underline{0.524}     & \textbf{0.444}  & \textbf{0.549}   & \underline{47.0}    & \underline{0.520}    & \underline{0.340}    & \textbf{0.690}   & \underline{0.456}    \\
Cohere                 & 0.454                & 0.263 & 0.352            & \underline{47.0}    & \underline{0.520}    & \textbf{0.361} & 0.642          & \textbf{0.462} \\
Human-Finetuned BGE-M3 & \textbf{0.574}       & \underline{0.333} & \underline{0.437}      & \textbf{50.0} & \textbf{0.460} & 0.333          & \underline{0.640}    & 0.438          \\  \bottomrule
\end{tabular}%
}
\caption{Effect of retrieval configuration on E2E performance. We use \texttt{gpt-4o-2024-08-06} as an LLM and vary different retrieval models to see the difference in E2E metrics.}
\label{table: retrieval_e2e_main}
\end{table}

The results showed that the \textit{Human-Finetuned BGE-M3} performed best overall, achieving results comparable to JinaAI V3 on NitiBench-Tax and Cohere embeddings on NitiBench-CCL. 
This result is consistent with the retrieval result in Table~\ref{table: retrieval_wangchan} and~\ref{table: retrieval_tax}. 
Building upon these results, we further analyze several findings based on varying retrieval models.

\begin{figure}[!ht]
\centering
\includegraphics[width=0.8\textwidth]{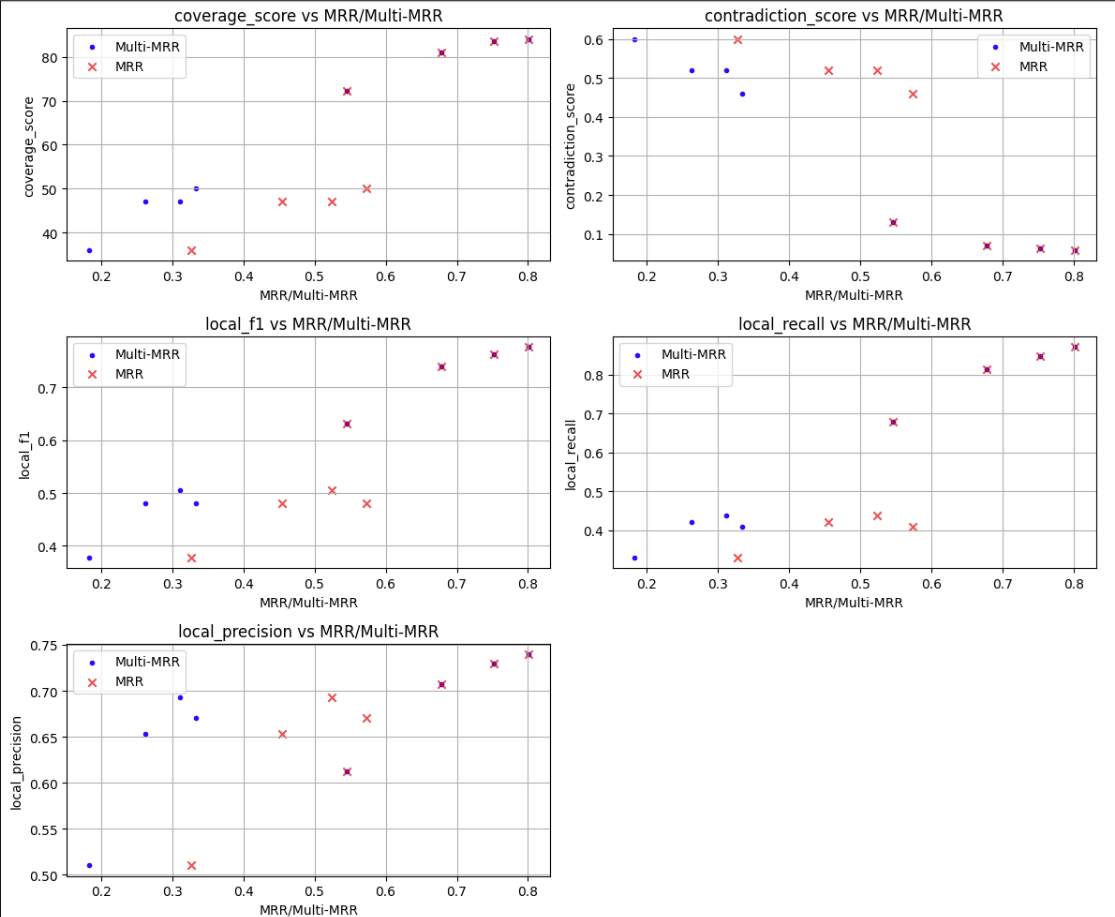}
\caption{Scatter plot of retrieval metrics vs E2E metrics}
\label{fig: retriever_vary_corr}
\end{figure}

\textbf{Multi-MRR do positively correlates with E2E metrics.} Figure~\ref{fig: retriever_vary_corr} visualizes the relationship between Multi-MRR (a retriever metric) and five end-to-end metrics: coverage, contradiction, E2E recall (local\_recall in the figure), E2E precision (local\_precision in the figure), and E2E F1 (local\_f1 in the figure). 
The figure reveals a strong positive correlation between Multi-MRR and all end-to-end metrics except for contradiction, which shows a negative correlation. We also found that the Multi-MRR correlates to the E2E metrics better than the single-label counterpart.
This is expected, as better retrieval leads to better overall system performance. 
These findings suggested that \textbf{the RAG framework can perform best when the relevant law sections are ranked at the top of the retrieved sections.}

\begin{figure}[!ht]
\centering
\begin{minipage}{0.48\textwidth}
    \centering
    \includegraphics[width=\textwidth]{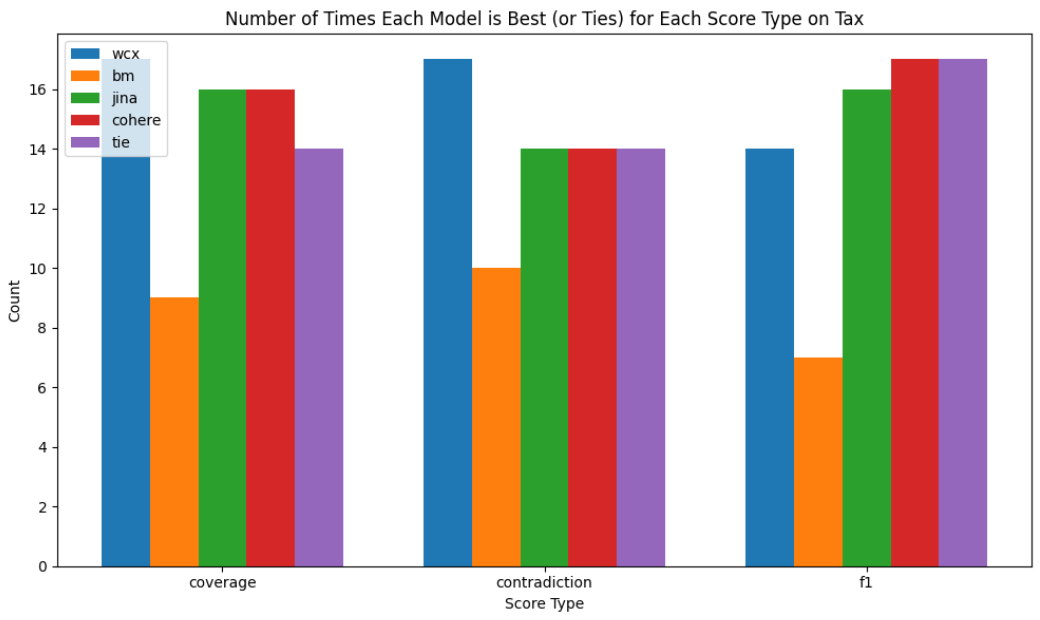}  
    \caption{Bar chart showing amount of times each retriever achieves the highest score on a sample on NitiBench-Tax}
    \label{fig:retriever_vary_win_tax}
\end{minipage}
\hfill
\begin{minipage}{0.48\textwidth}
    \centering
    \includegraphics[width=\textwidth]{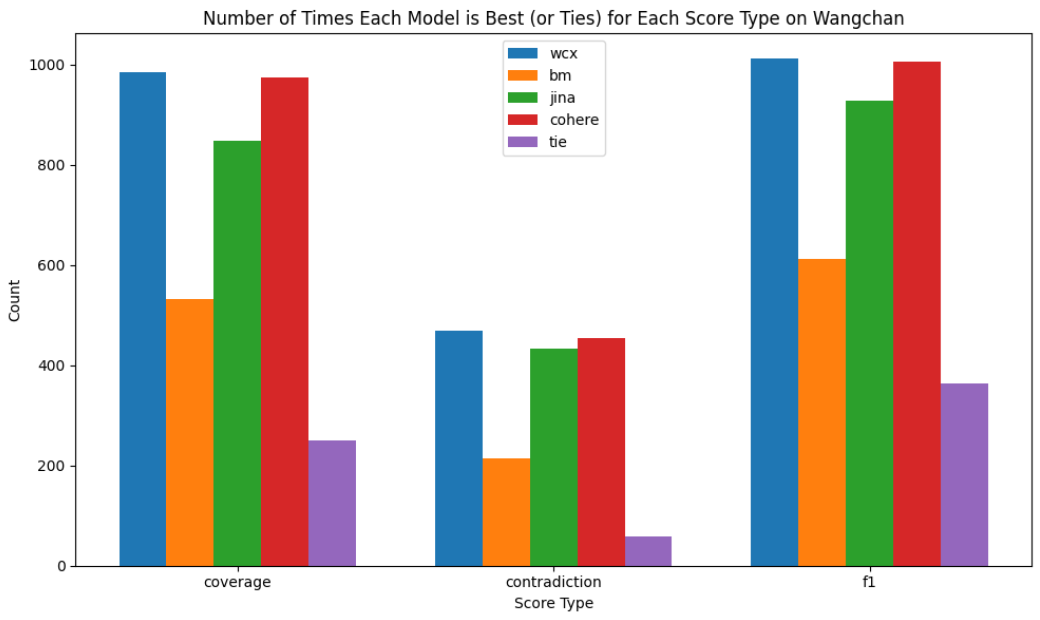}  
    \caption{Bar chart showing amount of times each retriever achieves the highest score on a sample on NitiBench-CCL}
    \label{fig:retriever_vary_win_wcx}
\end{minipage}
\end{figure}

\textbf{There remains a gap in overall retriever performance, indicating room for improvement across all retriever models.} 
To identify the current limitation of the retriever models, we analyzed samples where each retriever achieved the highest score for each end-to-end (E2E) metric.
Figures~\ref{fig:retriever_vary_win_tax} and~\ref{fig:retriever_vary_win_wcx} show the frequency of how each retrieval model achieves the highest score for each E2E metric. 
Note that we omit samples where all retrievers are tied at a maximum score, as the count for the tie column would be too high.
%
These figures demonstrate that the human-finetuned BGE-M3 model (\textbf{wcx} in the figures) most frequently achieved the highest E2E metric scores. 
Notably, a portion of both datasets showed identical E2E metric scores across all four retriever configurations.
Apart from the winning retrieval model for each E2E score, we are specifically interested in the case where all retrieval models are tied (while not getting a perfect score).
We suspect that the identical E2E metric scores result from the general performance gap in retriever models for certain types of queries, as well as the inherent structure of Thai legislation. 
This will be further discussed in more detail in \S~\ref{subsec: retriever_re_error_analysis_wcx} and~\ref{subsec: retriever_re_error_analysis_tax}. 

\subsection{Error Analysis on Retriever Performances on NitiBench-CCL}
\label{subsec: retriever_re_error_analysis_wcx}

Given the retriever results of the hierarchy-aware chunking strategy (showed in \S\ref{subsubsec: retriever_result}), we aim to further conduct an error analysis to answer the following questions: \textbf{What type of queries caused the retriever to fail, and what is the current limitation of dense retriever?}. 
To address these questions, first, we grouped the retrieval model performance of \textit{Human-Finetuned BGE-M3} (from Table~\ref{table: retrieval_e2e_main}) across all legislation on NitiBench-CCL and NitiBench-Tax respectively. 
The \enquote{Support} column indicates the number of queries for which a section within the given law code is relevant, while \enquote{Section Count} represents the total number of sections within that law code. 
The results are presented in Table~\ref{table: retrieval_wangchan_by_code} and~\ref{table: retrieval_tax_by_code}.

\begin{table}[!ht]
\centering

\resizebox{\textwidth}{!}{%
\renewcommand{\arraystretch}{1.4} 

\begin{tabular}{@{}p{0.6\textwidth}cccc@{}}
\toprule
\textbf{Law Code}                                                                                                                              & \multicolumn{1}{c}{\textbf{Multi MRR@10}} & \multicolumn{1}{c}{\textbf{Recall@10}} & \multicolumn{1}{c}{\textbf{Support}} & \multicolumn{1}{c}{\textbf{Section Count}} \\ \midrule
Business Registration Act, B.E. 2499                                                                                                  & \textbf{0.955}                   & \textbf{1.00}                 & 11                          & 22                                \\
Foreign Business Act, B.E. 2542                                                                                                       & \underline{0.900}                      & \textbf{1.00}                 & 36                          & 46                                \\
Trade Association Act, B.E. 2509                                                                                                      & 0.883                            & 0.925                         & 40                          & 56                                \\
Accounting Profession Act, B.E. 2547                                                                                                  & 0.862                            & 0.922                         & 64                          & 78                                \\
Secured Transactions Act, B.E. 2558                                                                                                   & 0.848                            & 0.932                         & 74                          & 91                                \\
Revenue Code                                                                                                                          & 0.836                            & 0.911                         & 484                         & 361                               \\
Provident Fund Act, B.E. 2530                                                                                                         & 0.834                            & \underline{0.972}                   & 36                          & 51                                \\
Emergency Decree on Digital Asset Businesses, B.E. 2561                                                                               & 0.827                            & 0.962                         & 78                          & 100                               \\
Public Limited Companies Act, B.E. 2535                                                                                               & 0.825                            & 0.919                         & 186                         & 231                               \\
Act on Offenses Relating to Registered Partnerships, Limited Partnerships, Companies Limited, Associations and Foundations, B.E. 2499 & 0.819                            & 0.958                         & 48                          & 76                                \\
Financial Institutions Business Act, B.E. 2551                                                                                        & 0.813                            & 0.897                         & 165                         & 168                               \\
Derivatives Act, B.E. 2546                                                                                                            & 0.813                            & 0.927                         & 109                         & 156                               \\
Chamber of Commerce Act, B.E. 2509                                                                                                    & 0.811                            & 0.864                         & 22                          & 62                                \\
Civil and Commercial Code                                                                                                             & 0.807                            & 0.909                         & 1618                        & 1842                              \\
Trust for Transactions in Capital Market Act, B.E. 2550                                                                               & 0.805                            & 0.890                         & 73                          & 94                                \\
Emergency Decree on Special Purpose Juristic Person for Securitization, B.E. 2540                                                     & 0.798                            & 0.879                         & 33                          & 46                                \\
Securities and Exchange Act, B.E. 2535                                                                                                & 0.786                            & 0.898                         & 294                         & 499                               \\
Energy Conservation Promotion Act, B.E. 2535                                                                                          & 0.778                            & 0.948                         & 58                          & 68                                \\
Accounting Act, B.E. 2543                                                                                                             & 0.728                            & 0.895                         & 38                          & 45                                \\
Energy Industry Act, B.E. 2550                                                                                                        & 0.726                            & 0.843                         & 153                         & 155                               \\
Petroleum Income Tax Act, B.E. 2514                                                                                                   & 0.616                            & 0.764                         & 110                         & 115                               \\ \bottomrule

\end{tabular}
}
\caption{Retrieval Evaluation Result by law code on NitiBench-CCL}
\label{table: retrieval_wangchan_by_code}
\end{table}

\begin{table}[!ht]
\centering

\renewcommand{\arraystretch}{1.2} 
\small
\begin{tabular}{@{}lcccc@{}}
\toprule
\textbf{Law Code}                               & \textbf{Multi MRR@10} & \textbf{Recall@10} & \textbf{Support} & \textbf{Section Count} \\ \midrule
Revenue Code                           & 0.333        & 0.437     & 50      & 361           \\
Accounting Act, B.E. 2543              & 0.333        & 0.500       & 1       & 45            \\
Civil and Commercial Code              & 0.173        & 0.370      & 8       & 1842          \\
Securities and Exchange Act, B.E. 2535 & 0.065        & 0.333     & 1       & 499           \\ \bottomrule
\end{tabular}
\caption{Retrieval Evaluation Result by law code on NitiBench-Tax}
\label{table: retrieval_tax_by_code}
\end{table}

Focusing on NitiBench-CCL, we analyze a subset of approximately 780 queries (out of 4000) that showed no improvement in MRR@10 after fine-tuning with in-domain data. 
Within this subset, 540 queries yielded identical MRR@10 scores for both the base and human-finetuned BGE-M3 models. 
A closer examination of these queries reveals that most errors fall into four distinct categories:
\begin{itemize}
    \item Hidden Hierarchical Information
    \item Nested Structure
    \item Missable Details
    \item Complex Queries
\end{itemize}

\subsubsection{Hidden Hierarchical Information}
This error type arises from the redundancy and repetition within Thai Financial Law. 
Multiple sections often convey identical information, differing only in their location within the legal hierarchy or the specific law code. 
For example, Sections 27 and 89/1 of the Revenue Code both address additional tax charges for late payment, but Section 27 resides in Chapter 2: Procedures Related to Tax Assessment, while Section 89/1 belongs to Chapter 4: Value-Added Tax. 
Without providing a proper comprehensive definition of all law hierarchies as a context to the prompt, differentiating these sections becomes extremely difficult.


\begin{quote}
\textbf{Section 27 of Revenue Code}  
Any person who fails to pay or remit taxes within the specified timeframes as stipulated in various chapters of this title concerning assessed taxes shall be subject to an additional charge of 1.5\% per month or a fraction thereof on the tax amount...
\end{quote}

\begin{quote}
\textbf{Section 89/1 of Revenue Code}  
Any person who fails to fully pay or remit taxes within the prescribed period under this chapter shall incur an additional charge of 1.5\% per month or a fraction thereof on the tax amount...
\end{quote}

Two main approaches can potentially address this hierarchical information challenge. 
First, integrate this information directly into the section content, explicitly stating the chapter, division, and law code (or possibly its comprehensive definition of each level). 
This requires retraining the retriever model to incorporate this new information into its embedding generation process. 
Alternatively, a heuristic approach or a separate retrieval model could pre-select relevant legislation, titles, and divisions before section-level retrieval, thus reducing redundancy.

\subsubsection{Nested Structure}
This error type stems from the nested structure of Thai financial law, where sections often refer to other sections within the same or different legislation. 
These referencing sections may lack complete explanations; instead, they should point to other sections for details. 
For instance, section 1409 of the Civil and Commercial Code mandates the analogical application of sections 552–555, 558, 562, and 563 but does not contain their content. 
Consequently, retrievers might struggle to match queries with section 1409 directly, instead retrieving the referenced sections containing the relevant details.

\begin{quote}
\textbf{Section 1409 of Civil and Commercial Code}  
The provisions of this Code regarding the duties and liabilities of a lessee, as stipulated in Sections 552 to 555, Sections 558, 562, and 563, shall apply mutatis mutandis.
\end{quote}

Another common example involves sections specifying penalties. 
Determining the penalty for a specific offense often requires locating the section prohibiting the behavior and a separate section defining the corresponding penalty. 
This separation can cause retrieval issues. 
For example, consider Sections 186 and 291 of the Securities and Exchange Act. 
Section 186 specifies the authority of the Securities and Exchange Commission (SEC), while section 291 outlines penalties for non-compliance with SEC orders. 
A query about penalties for a specific offense might incorrectly retrieve Section 186 (describing the SEC's authority) instead of Section 291 (detailing the actual penalties) because the penalty section itself lacks details about the specific offense.



\begin{quote}
\textbf{Section 186 of Securities and Exchange Act B.E. 2535}  
To prevent potential damage to the public interest or the national economy, the SEC Board has the authority to:  
(1) Temporarily prohibit the trading of all listed securities on the stock exchange...
\end{quote}

\begin{quote}
\textbf{Section 291 of Securities and Exchange Act B.E. 2535}  
Anyone who violates or fails to comply with the order issued under Section 186 (2) shall be subject to imprisonment for a term not exceeding one year, a fine not exceeding 300,000 baht, or both.
\end{quote}

To potentially mitigate these nested structure errors, a knowledge graph representing the relationships between sections could be employed. 
This would allow for augmenting retrieved content with referenced or referencing sections, providing the LLM with a more complete context. 
We implemented a simple knowledge graph, referred to as \textit{NitiLink}, in the pipeline and observed improved retriever performance, as shown in \S~\ref{subsubsec: referencer_result}, particularly on NitiBench-Tax, which requires multiple sections and legal reasoning. 
However, a more advanced implementation of this component should be explored.

\subsubsection{Missable Details}
This type of error occurs when subtle but critical details in a query are overlooked by dense embedding models, which often focus on general relevance instead of fine-grained nuances. 
For instance, consider a legal query about whether one guarantor can guarantee another guarantor. 
The correct answer relies on Section 682 of the Civil and Commercial Code, which explicitly allows such arrangements. 
However, the retriever might incorrectly prioritize Section 680, which only provides a general definition of a guarantor, thus missing the specific nuance required to address the query accurately.





\begin{quote}
\textbf{Question:} Can a person act as a guarantor for another guarantor?\\

\textbf{Retrieved Section:}  
Section 680 of Civil and Commercial Code  
Suretyship is a contract in which a third party, called the guarantor, binds themselves to a creditor to fulfill the obligation if the debtor fails to do so.  

Furthermore, a suretyship contract must be evidenced in writing and signed by the guarantor; otherwise, it cannot be enforced in court.\\

\textbf{Gold Section:}  
Section 682 of Civil and Commercial Code 
A person may act as a secondary guarantor, meaning they guarantee the obligations of the primary guarantor.  

If multiple persons act as guarantors for the same debt, they are jointly liable as co-debtors, even if they did not enter into the suretyship agreement together.
\end{quote}

One approach to address these missable detail errors involves adjusting the weighting in late-interaction embedding models to prioritize keyword matching. 
This increased focus on keywords could improve the model's ability to capture subtle nuances within queries.


\subsubsection{Complex Queries}
This category encompasses complex queries requiring multi-step reasoning. 
These errors are prevalent and difficult to resolve through simple fine-tuning. 
For example, a query about returning a purchased ring to someone claiming ownership (who wasn't the seller) requires multiple reasoning steps. 
The model must first recognize the ring as movable property, then retrieve Section 1303 of the Civil and Commercial Code, which grants superior ownership rights to a good-faith purchaser for value. 
Current retrieval models struggle with this multi-step reasoning, often retrieving more general sections like Section 1332, which addresses returning purchased goods to the seller.


\begin{quote}
\textbf{Question:} If I buy a ring from someone and later another person claims to be the rightful owner, do I have to return the ring? \\

\textbf{Retrieved Section:}  
Section 1332 of Civil and Commercial Code  
A person who purchases property in good faith at a public auction, in a marketplace, or from a merchant dealing in such goods is not required to return it to the rightful owner unless the owner reimburses the purchase price. \\

\textbf{Gold Section:}  
Section 1303 of Civil and Commercial Code 
If multiple individuals claim ownership of the same movable property based on different legal grounds, the person in possession has superior rights, provided they acquired it for value and in good faith.  

This provision does not apply to items specified in the preceding section, lost property, or assets acquired through wrongful acts.
\end{quote}

Addressing these complex queries requires an additional step to decompose the query into multiple sub-queries, each focusing on a specific aspect. 
This allows the model to perform a more nuanced \enquote{thought process}, inferring implicit information and breaking down complex issues into smaller, more manageable parts, ultimately improving retrieval accuracy and comprehensiveness.

\subsection{Error Analysis on Retriever Performances on NitiBench-Tax}
\label{subsec: retriever_re_error_analysis_tax}
Beyond NitiBench-CCL, a detailed error analysis was also performed on the NitiBench-Tax. 

\subsubsection{Generic Section Retrieval Challenge}
This analysis focused on false negatives and false negative rates, particularly within sections of the Revenue Code. 
Table~\ref{table: retrieval_tax_section_fn} highlights the five sections with the highest false negative rates and counts.

\begin{table}[!ht]
\centering
\small
\renewcommand{\arraystretch}{1.2} 

\begin{tabular}{@{}lcc@{}}
\toprule
\multicolumn{1}{l}{\textbf{Section}} & \multicolumn{1}{l}{\textbf{False Negative}} & \multicolumn{1}{l}{\textbf{False Negative Rate}} \\ \midrule
77/2                        & 6                                  & 1.0                                     \\
77/1                        & 5                                  & 1.0                                     \\
2                           & 5                                  & 1.0                                     \\
3 octo                     & 4                                  & 1.0                                     \\
91/5                        & 2                                  & 1.0                                     \\ \bottomrule
\end{tabular}
\caption{Top 5 false negative and false negative ratio of sections in Revenue Code of NitiBench-Tax}
\label{table: retrieval_tax_section_fn}
\end{table}

These sections largely contain either terminology definitions or broad legal overviews. 
Because they lack specific details and conditions, they pose a retrieval challenge: while important, their generality makes them a poor match for the concrete scenarios found in NitiBench-Tax. 
The retriever model, therefore, tends to favor sections with greater contextual relevance. We refer to this problem as the \enquote{Generic Section Retrieval Challenge}.

For example, a query about VAT applicability to animal treatment chemicals and medical supplies might retrieve sections discussing specific VAT exemptions or surcharges rather than the overarching section 77/2 of the Revenue Code, which simply states that all sales, imports, and services are subject to VAT. 
While some retrieved sections may be relevant, the model fails to surface the foundational Section 77/2, hindering comprehensive legal reasoning.

\begin{quote}
    \textbf{Question:}  
    A company operates an animal hospital providing veterinary services. The company purchases medicine for animal treatment and separately charges customers for medicine and treatment services. The company seeks clarification on the following:  
    1. Can the company charge VAT only on treatment services while exempting VAT on medicine used for treatment?  
    2. Should the company charge VAT on chemicals used for treatment, such as disinfectants, alcohol, and saline solution?  
    3. Should the company charge VAT on medical supplies used for treatment, such as cotton, IV tubes, needles, and sutures?  
    
    \textbf{Retrieved Section:}  
    Section 83/10 of Revenue Code – VAT Collection:  
    (1) For imported goods, the Customs Department shall collect VAT on behalf of the Revenue Department. In the case of abandoned goods under Section 78/2 (3), the Customs Department shall deduct VAT, along with penalties and surcharges, according to the criteria set by the Director-General.  
    (2) For the sale of goods or services subject to excise tax, the Excise Department shall collect VAT on behalf of the Revenue Department.  
    
    \textbf{Gold Section:}  
    Section 77/2 of Revenue Code – VAT Liability:  
    The following activities conducted in Thailand are subject to VAT under this section:  
    (1) The sale of goods or the provision of services by an operator.  
    (2) The importation of goods by an importer.  
    
    The provision of services in Thailand refers to services performed in Thailand, regardless of whether they are used domestically or abroad.  
    Services performed abroad but used in Thailand are also considered services rendered in Thailand.
\end{quote}

Two strategies may mitigate this issue. 
First, we should consider whether abstract and general sections constitute foundational knowledge that the LLM requires for all queries within a RAG system, rather than retrieving them selectively. 
Second, keyword matching retrieval could prove effective for terminology sections, as query terms often directly match defined phrases.

Beyond the Generic Section Retrieval Challenge, our analysis of NitiBench-Tax revealed two additional error categories:

\subsubsection{Incorrect Legislation Retrieval} 
Table~\ref{table: retrieval_tax_code_fp} presents the distribution of false positives at the law code level.

\begin{table}[!h]
\centering

\renewcommand{\arraystretch}{1.2} 

\begin{tabular}{@{}lc@{}}
\toprule
\textbf{Law Code}                                                                                                                    & \textbf{False Positive} \\ \midrule
Revenue Code                                                                                                                & 280            \\
Petroleum Income Tax Act, B.E. 2514                                                                                         & 30             \\
Civil and Commercial Code                                                                                                   & 21             \\
Securities and Exchange Act, B.E. 2535                                                                                      & 15             \\
\begin{tabular}[c]{@{}c@{}}Government Procurement and Supplies Management Act, B.E. 2560\end{tabular}                     & 15             \\
Budget Procedure Act, B.E. 2561                                                                                             & 14             \\
Energy Industry Act, B.E. 2550                                                                                              & 12             \\
Business Registration Act, B.E. 2499                                                                                        & 10             \\
Public Limited Companies Act, B.E. 2535                                                                                     & 8              \\
Energy Conservation Promotion Act, B.E. 2535                                                                                & 5              \\
Trust for Transactions in Capital Market Act, B.E. 2550                                                                     & 5              \\
Financial Institutions Business Act, B.E. 2551                                                                              & 4              \\
National Economic and Social Development Act, B.E. 2561                                                                     & 3              \\
Accounting Profession Act, B.E. 2547                                                                                        & 3              \\
\begin{tabular}[c]{@{}c@{}}Act on the Management of Shares and Stocks of Ministers, B.E. 2543\end{tabular}                & 2              \\
State Enterprise Development and Governance Act, B.E. 2562                                                                  & 2              \\
Fiscal Discipline Act, B.E. 2561                                                                                            & 2              \\
Accounting Act, B.E. 2543                                                                                                   & 1              \\
\begin{tabular}[c]{@{}c@{}}Emergency Decree on Special Purpose Juristic Person for Securitization, B.E. 2540\end{tabular} & 1              \\
Provident Fund Act, B.E. 2530                                                                                               & 1              \\
Emergency Decree on Digital Asset Businesses, B.E. 2561                                                                     & 1              \\
Foreign Business Act, B.E. 2542                                                                                             & 1              \\
Derivatives Act, B.E. 2546                                                                                                  & 1              \\ \bottomrule
\end{tabular}
\caption{False positive distribution on NitiBench-Tax on law code level}
\label{table: retrieval_tax_code_fp}
\end{table}

While the NitiBench-Tax's ground truth labels span only 4 legislation, retrieved false positives originate from 21 different legislation. 
This mirrors the hidden hierarchical information problem observed in NitiBench-CCL, where similar concepts appear in different legislation. 
For instance, both Section 91/15 of the Revenue Code and Section 56 of the Petroleum Act address tax assessor authority, differing only in their conditions of application.

This issue is amplified in NitiBench-Tax because queries directed to the Revenue Department often omit details implicitly covered by the Revenue Code's scope.
Besides previously suggested solutions, such as incorporating hidden information into section content and employing level-based retrieval, restricting retrieval to sections within relevant legislation when queries pertain to a specific domain could potentially alleviate this problem. 
This targeted approach should reduce false positives from unrelated codes.

        



\subsubsection{Incorrect Tax Type Retrieval}
Beyond false positives due to similar content, the retrieval model frequently misidentifies the relevant tax type, especially in complex queries. 
This resembles the \enquote{hard query} error type observed in NitiBench-CCL.

For example, a query about the tax obligations of an employee in Thailand receiving income from both a subsidiary and its parent company (a personal income tax question) should retrieve Sections 41, 48, 50, and 56 of the Revenue Code, which addresses personal income tax, withholding obligations, and calculating tax on foreign income.

However, the model instead retrieves sections related to corporate and export taxes. 
This likely stems from keywords like \enquote{company}, \enquote{corporate}, and \enquote{foreign} influencing the query embedding, shifting its focus away from personal income tax.

\begin{quote}
    \textbf{Question: }  
    A company incorporated under Thai law operates in the textile industry and has investments in related businesses both in Thailand and abroad. The company appoints executives or employees as directors in its foreign subsidiaries. These individuals receive salaries and benefits for their duties in Thailand, paid through their Thai bank accounts. Additionally, they receive director’s fees from the foreign subsidiaries based on market rates or country-specific regulations, with payments made directly to their foreign bank accounts. The company seeks clarification on the following:
    
    1. If the company pays salaries or compensation solely for duties performed for the employer in Thailand, is the company required to include only the income derived from such duties in Thailand for withholding tax calculation under Section 50(1) of the Revenue Code?
    2. If an executive or employee receives director’s fees from a foreign subsidiary and stays in Thailand for more than 180 days within the same tax year, but does not remit this income into Thailand within the same tax year, is the individual exempt from personal income tax for foreign-earned income under Section 41, paragraph two of the Revenue Code?
    
    \textbf{Retrieved Section: }  
    Revenue Code Section 76 bis:  
    A company or juristic partnership established under foreign law that has employees, agents, or representatives conducting business in Thailand, resulting in income or profits in Thailand, shall be deemed to be carrying on business in Thailand. Such persons, whether individuals or entities, shall be considered representatives of the foreign company or partnership and shall be responsible for filing tax returns and paying tax under this section for the relevant income or profits.  
    
    If the responsible person cannot calculate net profits for tax purposes, the provisions on tax assessment under Section 71(1) shall apply accordingly.  
    
    In cases of tax assessment under this section, an appeal against the assessment may be filed.  
\end{quote}

\subsection{Effectiveness of Multi-label Metrics}
\label{subsec: multi_metrics}

To further validate the effectiveness of our proposed multi-label metrics, we compute the correlation between conventional retrieval metrics (Hit Rate and MRR) compared to its multi-label variant.
We use eight retriever model performances (see Table~\ref{table: retrieval_e2e_main}) to measure the correlation between retrieval and the E2E metric.
The result was presented in Table~\ref{tab:correlation}.

According to the result, we can see that our Multi-MRR and Multi-Hit Rate have a higher correlation compared to conventional MRR and hit rate.
These results emphasize the importance of using multi-label metrics in legal QA setups.

\begin{table}[!ht]
\centering
\fontsize{8}{11}\selectfont  
\setlength{\tabcolsep}{3.75pt}   
\renewcommand{\arraystretch}{0.85}  
\begin{tabular}{lccc}
\toprule
               & \textbf{Coverage ($\uparrow$)} & \textbf{Contradiction ($\downarrow$)} & \textbf{E2E F1 ($\uparrow$)} \\ 
\midrule
Hit Rate       & 0.741            & -0.672                 & 0.780          \\
\rowcolor[HTML]{EFEFEF} 
Multi Hit Rate & 0.989            & -0.986                 & 0.984          \\
MRR            & 0.906            & -0.859                 & 0.933         \\
\rowcolor[HTML]{EFEFEF} 
Multi MRR      & 0.989            & -0.973                 & 0.991          \\
\bottomrule
\end{tabular}
\caption{Correlation between conventional and multi-retrieval metrics with evaluation measures using data from 8 retrievers (Table~\ref{table: retrieval_e2e_main}})
\label{tab:correlation}
\end{table}

\subsection{Analysis of Effect of LLMs on E2E and Retrieval Performance}
\label{subsec: llm_error_analysis}

We investigate two metrics to assess the gaps between retrieval recall and E2E recall:
\begin{itemize}
    \item \textbf{Recall Difference}: Formally, defined as the difference between retriever recall and E2E recall. This metric reflects the RAG system's ability to utilize retrieved documents for reasoning and answer generation.
    The lower the recall difference, the better LLM can reach upper bound recall performance of the retriever.
    \item \textbf{Zero Recall, Non-Zero Coverage Ratio (Hallucination Rate)}: The ratio of samples with zero E2E recall but non-zero coverage, which indicates the likelihood of generating answers without relying on correctly retrieved documents, potentially due to parametric knowledge or hallucinations.
\end{itemize}
The results are shown in Table~\ref{table: llm_e2e_recall_diff}.
From the results, \texttt{claude-3.5-sonnet} minimizes the recall difference across both datasets, demonstrating its strong ability to leverage retrieved documents for reasoning. 
This contributes to its high Coverage and low Contradiction scores. 
OpenAI achieves the lowest hallucination ratio on NitiBench-Tax, while Claude and Gemini perform best on NitiBench-CCL. 
As for Typhoon models, the 70b variant performs well in terms of the hallucination ratio in NitiBench-Tax.
However, both Typhoon models exhibit high recall differences on both datasets which brings their abilities to utilize correctly retrieved documents into question. 
The recall difference is significantly higher for NitiBench-Tax than NitiBench-CCL, suggesting that NitiBench-Tax poses greater challenges for applying the retrieved information to reasoning and answer generation.

\begin{table}[!ht]
\centering
\renewcommand{\arraystretch}{1.2}
\resizebox{0.8\textwidth}{!}{%
\begin{tabular}{@{}lcc@{}}
\toprule
\textbf{Model} & \textbf{Recall Difference ($\downarrow$)} & \begin{tabular}[c]{@{}c@{}}\textbf{Zero Recall, Non Zero}\\\textbf{Coverage Ratio (Hallucination Rate) ($\downarrow$)}\end{tabular} \\ 
\midrule
\multicolumn{3}{c}{\textbf{NitiBench-CCL}} \\
\midrule
\texttt{gpt-4o-2024-08-06}          & 0.058 & 0.069 \\
\texttt{claude-3.5-sonnet-20240620} & \textbf{0.036} & \underline{0.060} \\
\texttt{gemini-1.5-pro-002}         & \underline{0.045} & \textbf{0.058} \\
\texttt{typhoon-v2-70b-instruct}    & 0.076 & 0.079 \\
\texttt{typhoon-v2-8b-instruct}     & 0.163 & 0.120 \\

\midrule
\multicolumn{3}{c}{\textbf{NitiBench-Tax}} \\
\midrule
\texttt{gpt-4o-2024-08-06}          & \underline{0.100} & \textbf{0.100} \\
\texttt{claude-3.5-sonnet-20240620} & \textbf{0.095} & 0.160 \\
\texttt{gemini-1.5-pro-002}         & 0.102 & 0.140 \\
\texttt{typhoon-v2-70b-instruct}    & 0.148 & \underline{0.120} \\
\texttt{typhoon-v2-8b-instruct}     & 0.246 & 0.200 \\

\bottomrule
\end{tabular}%
}
\caption{The table shows the performance gap between retrieval recall and citation recall (E2E recall), and the hallucination rate defined by a ratio of the sample that has zero recall (LLM fails to cite correct documents) but achieves non-zero coverage (showing that LLM can answer despite not citing a correct document). Both metrics were measured across both NitiBench-CCl and NitiBench-Tax with different LLM setups.}
\label{table: llm_e2e_recall_diff}
\end{table}

Analyzing samples with high retriever recall but low coverage reveals a recurring pattern: 

\textbf{LLM often omits intermediate reasoning steps, directly applying retrieved laws without sufficient explanation.} For example, a query about exempting income sent from a foreign husband to his wife in Thailand (without a registered marriage) requires multi-step reasoning.

First, the relevant law (Section 42 of the Revenue Code) exempts income received from moral obligation or gifts given during ceremonies or customs. 
The system should then reason whether the money sent to the wife qualifies as a moral obligation. 
Evidence of the unregistered marriage would support this claim and justify the exemption.

However, the LLM often skips this detailed reasoning. 
It prematurely concludes that the money isn't from tradition or moral obligation and is, therefore, taxable.
This overlooks the possibility of the husband's payment being a moral obligation, failing to provide the necessary intermediate reasoning steps.

\begin{quote}
    \textbf{Question: } 
    1. Mrs. K is currently married to a foreigner named Mr. A, but they have not registered their marriage. 
    2. The Revenue Department has received an Exchange of Information from the foreign tax authority (Country B) indicating that Mrs. K, who is a resident of Thailand, received a gift from Mr. A, her boyfriend, in Thai baht, calculated at the average exchange rate of the Bank of Thailand (BOT) as follows:
        2.1 Amount of 1 million baht on March 20, 2017...
    4. Is the money that Mrs. K received in this case considered exempt income under Section 42(28) of the Revenue Code?

    \textbf{Relevant Laws: }
    Revenue Code Section 42
    ...
    (28) Income received from support due to natural duty or from voluntary gifts on ceremonial occasions or in accordance with customs, provided by someone other than parents, descendants, or spouses, but only income not exceeding ten million baht for the entire tax year.
    ...

    \textbf{RAG Analysis: }

    From the given case details, we need to consider the following points:
    
    1. Mrs.K and Mr. A are not legally married, but they are in a relationship.
    
    2. Mrs.K received large sums of money as gifts from Mr. A in 2560, 2561, and 2562.
    
    3. Mrs.K also receives monthly allowances from Mr. A for household expenses.
    
    4. The question is whether these gifts are exempt under Section 42(28) of the Revenue Code.
    
    According to Section 42(28), income received from moral support or gifts given during ceremonies or on occasions according to custom traditions, from persons who are not ancestors, descendants, or spouses, is exempt from income tax, but only for the portion not exceeding 10 million baht throughout the tax year.
    
    In this case:
    1. Mr. A is not legally Mrs.K's spouse, so he falls under the category of \enquote{persons who are not ancestors, descendants, or spouses}.
    2. The gifts received in each year (1 million, 800,000, and 700,000 baht) are all below the 10 million baht threshold.
    3. However, these gifts do not seem to be given \enquote{during ceremonies or on occasions according to custom traditions}. They appear to be regular gifts given in a relationship.

\end{quote}

\textbf{LLM's tendency to incorporate multiple similar, retrieved laws into its answer, retrieved laws into its answer despite being irrelevant.} 
This pattern is often exhibited in NitiBench-CCL. 
We suspect that this could stem from the query's ambiguity and the lack of context surrounding laws with similar content but different applicability conditions.

For example, a query about the tax payment deadline after receiving an assessment notification should retrieve Section 18 bis of the Revenue Code (specifying 7 days). 
However, due to the query's vagueness, the model might also incorporate Section 18 ter of the Revenue Code and Section 44 of the Petroleum Income Tax, both containing similar content but with different conditions.

This issue will likely persist until the system can engage in multi-turn conversations to clarify specific conditions and refine document retrieval.

\begin{quote}
    \textbf{Question: } 
    How many days does a taxpayer have to pay tax after receiving the tax assessment notice?

    \textbf{Relevant Laws: }
    Revenue Code Section 18 Bis

    In cases where necessary to protect the interests of tax collection, the assessing officer has the authority to assess and demand tax from the taxpayer before the due date for filing the return. Once the assessment is made, the amount of tax due must be notified to the taxpayer, and the taxpayer is required to pay the tax within seven days from the date of receiving the assessment notice. In this case, an appeal against the assessment is also allowed. 

    The tax assessed under this provision is considered a credit for the taxpayer in calculating their tax.
    
    When exercising authority under this section, the assessing officer may also order the taxpayer to file a return in the form specified by the Director-General.

    \textbf{RAG Analysis: }

    From the given question, we need to determine the timeframe for tax payment after receiving a tax assessment notice. According to the provided laws, there are different timeframes mentioned for various situations:
    
    1. Section 18 Tri of the Revenue Code states that the taxpayer must pay the tax within 30 days from the date of receiving the assessment notice.
    
    2. Section 44 of the Petroleum Income Tax Act B.E. 2514 also mentions a 30-day period for tax payment after receiving the assessment notice.
    
    3. Section 18 Bis of the Revenue Code mentions a special case where the tax officer can assess and collect tax before the due date of filing a tax return. In this case, the taxpayer must pay within 7 days of receiving the assessment notice.
    
    Since the question doesn't specify a particular type of tax or situation, we should provide the general answer based on the most common timeframe mentioned in the laws. 

\end{quote}

\begin{figure}[!ht]
  \centering
  \begin{subfigure}[b]{0.48\textwidth}
    \centering
    \includegraphics[width=\textwidth]{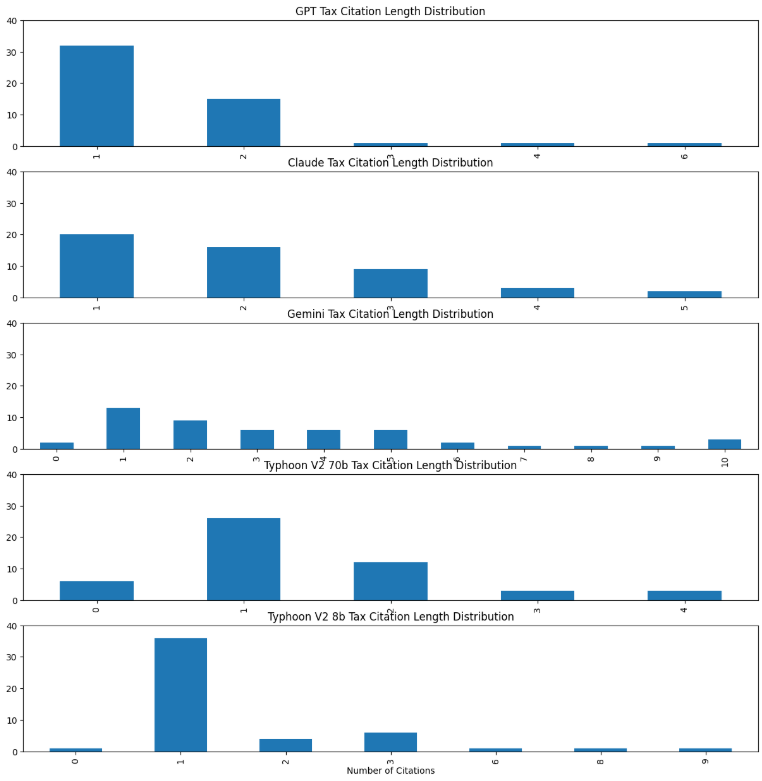}
    \caption{Distribution of the number of citations on NitiBench-Tax}
    \label{fig:tax_citation_len}
  \end{subfigure}
  \hfill
  \begin{subfigure}[b]{0.48\textwidth}
    \centering
    \includegraphics[width=\textwidth]{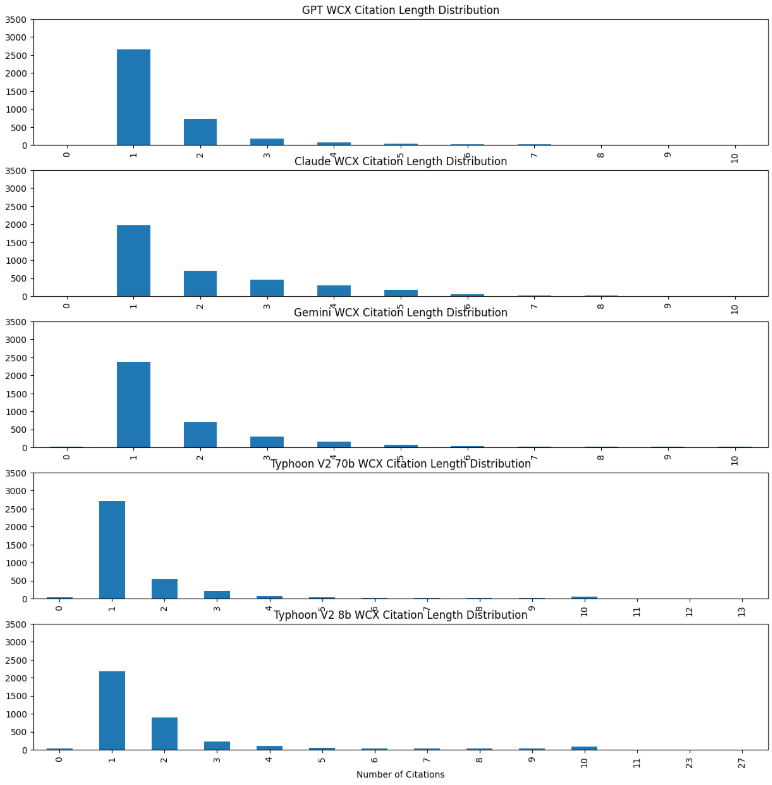}
    \caption{Distribution of the number of citations on NitiBench-CCL}
    \label{fig:wcx_citation_len}
  \end{subfigure}
  \caption{Distribution of the number of citations on both NitiBench-CCL and NitiBenth-Tax on different LLM.}
  \label{fig:citation_len}
\end{figure}

\textbf{Some LLM usually overcited retrieved documents} As shown in Table~\ref{table: llm_e2e_main_tax}, the Gemini model performs significantly worse than others on E2E precision. 
Figure~\ref{fig:citation_len}\subref{fig:tax_citation_len} reveals that Gemini cites considerably more laws than other models despite having a shorter average output token length than Claude. 
A detailed analysis shows that Gemini often cites laws that contain a few relevant keywords but lack overall relevance. 
Conversely, Figure~\ref{fig:citation_len}\subref{fig:wcx_citation_len} shows that Claude cites more laws than other models on average for NitiBench-CCL, likely due to its tendency to incorporate multiple laws for vaguely worded queries, as discussed previously.

\subsection{Analysis of LCLM result}
\label{subsec: lclm_error_analysis}

\begin{figure}[!ht]
\centering
\includegraphics[width=0.7\textwidth]{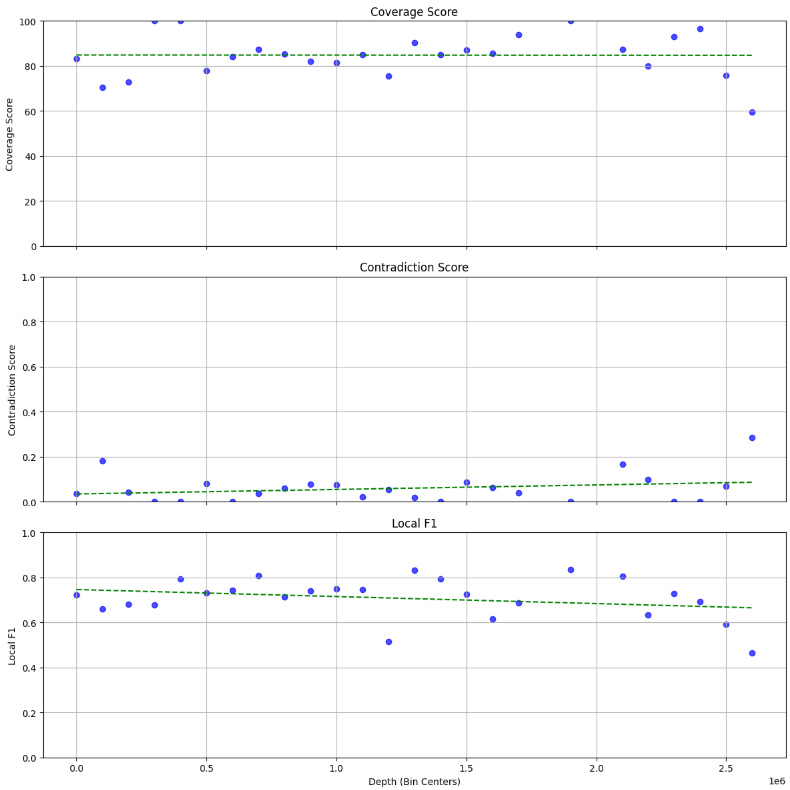}
\caption{Plot of performance grouped by the maximum depth of relevant context in the long context LLM setup.}
\label{fig: depth_performance}
\end{figure}

In this section, the effect of the relevant context position in the overall documents on the performance of the system is analyzed on the sampled NitiBench-CCL with long context Gemini 1.5 Pro ingesting 1.2 million tokens or 2.7 million characters. 
The resulting performance is binned every 100,000 characters by the maximum depth of the relevant laws that need to be retrieved, and the coverage, contradiction, and E2E F1 of each bin are averaged and plotted in Figure~\ref{fig: depth_performance}.

From the resulting plot, there is only a slight decrease in the coverage score and a slightly greater increase in the contradiction score as the depth increases. 
However, there is a significant drop in the E2E F1 score as the depth increases. 
Therefore, it can be concluded that \textbf{the depth of the relevant laws only mildly affects the coverage and contradiction score while its ability to cite applicable laws clearly has a negative impact.}

\subsection{Performance of reasoning model in RAG-based system}
\label{subsec: reasoning_model}
The rise of reasoning LLMs raises a question about their capability to perform legal reasoning in Thai legal QA tasks. 
To see the performance of such a reasoning model against a \enquote{normal} LLM, we evaluate the performance of the RAG system with reasoning LLM as a generator and compare its E2E performance to other systems described in \S\ref{subsubsec: e2e_best_setup}. 
Specifically, we opt to use the OpenAI's \texttt{o1-preview-2024-09-12}\cite{o1} as a generator with hierarchy-aware chunking, Human-Finetuned BGE-M3 as the retriever and not using NitiLink. 
Notably, the o1 model is not provided with any examples since the official reasoning guideline of OpenAI\footnote{\url{https://platform.openai.com/docs/guides/reasoning}} advise that the prompt should be kept as simple as possible. 
Furthermore, the API of o1 does not allow temperature changes; thus, it is set to 1.0. The experiment is conducted on the same subset used in \S\ref{subsec: setup_rq3} to limit the cost of the experiments, and the result is shown in Table~\ref{table: main_exp_reasoning}.

\begin{table}[!ht]
\centering
\renewcommand{\arraystretch}{1.2} 
\resizebox{\textwidth}{!}{%
\begin{tabular}{@{}lcccccccc@{}}
\toprule
\textbf{Setting} & \textbf{Retriever MRR ($\uparrow$)} & \textbf{Retriever Multi-MRR ($\uparrow$)} & \textbf{Retriever Recall ($\uparrow$)} & \textbf{Coverage ($\uparrow$)} & \textbf{Contradiction ($\downarrow$)} & \textbf{E2E Recall ($\uparrow$)} & \textbf{E2E Precision ($\uparrow$)} & \textbf{E2E F1 ($\uparrow$)} \\ 

\midrule
\multicolumn{9}{c}{\textbf{NitiBench-CCL (20\% subsampled)}} \\
\midrule
Parametric                                                    & -     & -     & -     & 60.6 & 0.198 & 0.197 & 0.147 & 0.169  \\
Naïve RAG                                                     & 0.549 & 0.549 & 0.649 & 77.7 & 0.092 & 0.740 & 0.379 & 0.501  \\
Our RAG                                                       & 0.825 & 0.825 & 0.945 & \textbf{90.1} & \textbf{0.028} & \textbf{0.920} & 0.453 & 0.607  \\
LCLM (Gemini)       & -     & -     & -     & 83.2 & 0.063 & 0.765 & \textbf{0.514} & \textbf{0.615}  \\
o1-preview-2024-09-12 & 0.825 & 0.825 & 0.945 & \underline{83.3} & \underline{0.044} & \underline{0.886} & \underline{0.466} & \underline{0.611} \\ 
\cellcolor{lightgray}Golden Context                                                & \cellcolor{lightgray}1.0   & \cellcolor{lightgray}1.0   & \cellcolor{lightgray}1.0   & \cellcolor{lightgray}94.2 & \cellcolor{lightgray}0.025 & \cellcolor{lightgray}0.999 & \cellcolor{lightgray}1.0   & \cellcolor{lightgray}0.999  \\

\midrule
\multicolumn{9}{c}{\textbf{NitiBench-Tax}} \\
\midrule
Parametric                                                    & -     & -     & -     & 46.0 & 0.480 & \textbf{0.458} & \underline{0.629} & \textbf{0.530}  \\
Naïve RAG                                                     & 0.120 & 0.048 & 0.062 & \underline{50.0} & \underline{0.460} & 0.306 & 0.463 & 0.368  \\
Our RAG                                                       & 0.574 & 0.333 & 0.437 & \textbf{51.0} & \textbf{0.440} & 0.389 & 0.554 & \textbf{0.457}  \\
LCLM (Gemini)       & -     & -     & -     & 36.0 & 0.620 & \underline{0.410} & 0.484 & 0.444  \\
o1-preview-2024-09-12 & 0.574 & 0.333 & 0.437 & 43.0 & \underline{0.460} & 0.333 & \textbf{0.640} & 0.438  \\ 
\cellcolor{lightgray}Golden Context                                                & \cellcolor{lightgray}1.0   & \cellcolor{lightgray}1.0   & \cellcolor{lightgray}1.0   & \cellcolor{lightgray}52.0 & \cellcolor{lightgray}0.460 & \cellcolor{lightgray}0.694 & \cellcolor{lightgray}1.000 & \cellcolor{lightgray}0.820  \\

\bottomrule
\end{tabular}%
}
\caption{Experiment results of reasoning model on sampled NitiBench-CCL and NitiBench-Tax. \textbf{For o1-preview, we did \emph{not} adopt few-shot prompt} to kept prompt simple according to \href{https://platform.openai.com/docs/guides/reasoning}{OpenAI guideline}.}
\label{table: main_exp_reasoning}
\end{table}

The reasoning model performs comparably to LCLM and the parametric setting on NitiBench-Tax while matching the performance of Naive RAG on NitiBench-CCL. 
This outcome may stem from the specialized nature of legal reasoning, which could differ from o1's capabilities. 
These results suggest that the current reasoning model is not yet well-suited for Thai legal QA tasks.

\section{Future Works}
\subsection{Limitations of WangchanX-Legal-ThaiCCL and NitiBench-CCL}
\label{subsec: wcx_limitation}

Our proposed dataset construction pipeline for WangchanX-Legal-ThaiCCL training data consists of two distinct processes: a semi-synthetic approach with human quality control (QC) for the training split and a fully human-annotated approach for the test split. 
This design choice helps manage costs, as creating a purely human-annotated training set is expensive. 
However, our test data construction pipeline still has some limitations.

\textbf{Ambiguous Queries from Single-Section Sampling}
This issue arises from the way dataset entries are constructed. 
Specifically, a single section is sampled from one of the 21 available legislation in the test set, and an expert annotator formulates a question and answer based on that section. 
However, annotators often focus solely on the sampled section's content without considering the specific conditions that make it unique compared to other similar sections on the same topic. 
This results in queries that are ambiguous and could be associated with multiple sections simultaneously. 
As discussed in \S\ref{subsec: llm_error_analysis}, this ambiguity leads to situations where the LLM incorporates multiple similar sections as relevant, even when the query was intended to target a single section. 

To address this issue, an additional step could be introduced in the pipeline. 
Instead of showing annotators only the targeted section, we could use an off-the-shelf retriever model or domain knowledge to retrieve similar sections. 
The annotators would then be asked to craft questions that are more specific to the targeted law and ensure that they are not easily confused with retrieved sections.

\textbf{The absence of truly multi-label queries} 
This applies to both the test (NitiBench-CCL) and training sets (of WangchanX-Legal-ThaiCCL). 
In the training set, multi-label ground truth is generated by allowing annotators to select relevant sections from retrieved documents. 
However, since the queries are initially derived from a single section, they are not inherently multi-label. 
As a result, our system evaluation lacks a mechanism to assess performance in scenarios requiring reasoning across multiple laws. 

To partially mitigate this, we constructed NitiBench-Tax, which includes genuinely multi-label queries.
However, it would be beneficial to refine the pipeline further by enabling annotators to create questions based on multiple legal provisions that are naturally grouped using heuristics. 
This adjustment would improve the dataset's ability to evaluate models in scenarios that require reasoning across multiple laws.

\textbf{Queries lack naturalness and differ from how everyday users might phrase questions in a Thai legal QA system} 
There are two potential approaches to address this issue.
One option is to implement an additional step in the system that transforms user queries into a style more aligned with NitiBench-CCL. 
Alternatively, the dataset construction pipeline could be adjusted to encourage annotators to use more informal language, or an LLM could be used to refine the annotators' phrasing to better reflect natural user input.

\subsection{Reasoning Evaluation}
\label{subsec: reasoning_evaluation}

Beyond the Coverage, Contradiction, and Citation scores proposed in our benchmark evaluation—measuring how well the generated answer covers the ground truth, whether it contradicts it, and how accurately it cites relevant sections—another crucial aspect of Legal QA evaluation is legal reasoning. 
Legal reasoning differs from general reasoning as it operates within a structured legal framework, relying on authoritative sources such as legislation and precedents while requiring precise interpretation of legal texts.
Unlike general reasoning, which is flexible and unconstrained, legal reasoning demands strict adherence to legal principles. 
Given its critical role in Legal QA, where the system's reasoning process is as important as the final answer, incorporating legal reasoning evaluation into the assessment framework would provide a more comprehensive measure of system performance.

Several studies have explored reasoning evaluation in LLMs.
ROSCOE \cite{roscoe} proposed a suite of metrics to assess reasoning chains, both with and without a gold standard reference. 
These metrics are categorized into four perspectives:

\begin{enumerate}
    \item \textbf{Semantic Alignment}: Measures how well a reasoning step is supported by the given context, primarily using cosine similarity.
    \item \textbf{Semantic Similarity}: Similar to Semantic Alignment but evaluates the reasoning process as a whole rather than individual steps.
    \item \textbf{Logical Inference}: Assesses logical fallacies between reasoning steps using Natural Language Inference (NLI) models.
    \item \textbf{Language Coherence}: Evaluates the fluency and coherence of the reasoning chain using perplexity scores.
\end{enumerate}

RECEVAL \cite{receval} further refines reasoning evaluation by defining two key qualities of a good reasoning step: 

\begin{itemize}
    \item \textbf{Correctness}: Ensuring that reasoning steps do not contradict themselves or previous steps.
    \item \textbf{Informativeness}: Ensuring that each step introduces new information contributing to the final answer.
\end{itemize}

This approach decomposes reasoning steps into atomic premise and conclusion statements. 
Correctness is measured using NLI models or Pointwise $\nu$-Information (PVI), which assesses entailment probability or the likelihood difference of generating conclusions with and without premises. 
Informativeness is measured by comparing PVI scores of reasoning chains with and without the current step.

While these methods rely on manually defined evaluation criteria, LLM Reasoner \cite{llmreasoners} automates this process by leveraging LLMs to categorize observed reasoning errors into evaluation criteria. 
These criteria are then used as prompts for automatic reasoning chain evaluation. 
This approach eliminates the need for manual criterion construction but requires training data with gold-standard reasoning chains.

Despite these advancements, reasoning evaluation for LLMs remains an ongoing challenge, particularly in the Thai legal domain. 
Key obstacles include defining what constitutes a good legal reasoning process and acquiring datasets that require complex legal reasoning beyond simple lookup-based answers.

\section{Conclusion}
\label{sec: conclusion}
One of the most significant challenges in implementing LLMs for Thai Legal QA systems is the lack of a standardized evaluation process. This issue arises due to the limited availability of Thai legal QA corpora and the absence of robust evaluation metrics. To address this, we introduce a novel benchmark dataset along with a corresponding task and evaluation framework named \textbf{NitiBench}.

Specifically, we construct two datasets: (1) \textbf{NitiBench-CCL} (derived from WangchanX-Legal-ThaiCCL), which covers general QA across 21 Thai financial law codes in its test split and 35 codes in its training split, and (2) \textbf{NitiBench-Tax}, which focuses on specialized QA involving real-world tax cases from the Thai Revenue Department, requiring extensive legal reasoning.

To complement this benchmark, we propose an evaluation framework that includes: (1) \textbf{Multi-label retrieval metrics}, in addition to traditional single-label metrics; (2) \textbf{An E2E task}, evaluating the system’s ability to generate correct answers consistent with ground truth while providing accurate legal citations; and (3) \textbf{E2E evaluation metrics}, measuring \textbf{Coverage} (how well the generated answer aligns with the ground truth), \textbf{Contradiction} (whether the generated answer contradicts the ground truth), and \textbf{Citation} (the accuracy of legal citations provided in the generated answer).

Using the proposed benchmark, we aim to address the research questions introduced in \S\ref{sec:introduction}. Through experiments outlined in \S\ref{subsec: setup_rq1}, we observed that our \textbf{hierarchy-aware} chunking approach slightly but consistently outperforms the system with the naive chunking strategy. This underscores the importance of integrating domain knowledge and understanding document structure when selecting a chunking strategy for optimal performance. Regarding the proposed \textbf{NitiLink}, we found that it notably improves context retrieval but does not enhance E2E performance on NitiBench-Tax queries. This may be attributed to the complexity of NitiBench-Tax queries, which require more detailed reasoning. For the NitiBench-CCL, NitiLink has a limited effect on improving context retrieval, as NitiBench-CCL queries often involve multiple non-hierarchically related sections, which NitiLink cannot resolve. However, those queries where NitiLink does provide an improvement show significant gains in E2E metrics.

In experiments described in \S\ref{subsec: setup_rq2}, we found that all retrieval models struggled with NitiBench-Tax. Among the retrievers, BGE-M3 fine-tuned on the WangchanX-Legal-ThaiCCL training split showed the highest performance. However, fine-tuning did not lead to performance improvements for NitiBench-Tax, and in some cases, it does not elicit performance improvements for NitiBench-CCL itself. Regarding the LLM component, no significant performance differences were observed between models, though Claude 3.5 Sonnet showed marginally better results. All tested LLMs, however, delivered mediocre performance on NitiBench-Tax. This issue is further illustrated in Table~\ref{subsubsec: e2e_best_result}, where even with a golden context provided, the LLMs still struggle to produce meaningful results on NitiBench-Tax queries, which demand complex reasoning. Additionally, the same table highlights limitations in the RAG pipeline on NitiBench-Tax. The performance is constrained by both the retriever model’s inherent limitations and the LLM's restricted ability to effectively leverage the correctly retrieved laws, as evidenced by the disparity between end-to-end recall and retriever recall.

We also evaluated the feasibility of using long-context LLMs (LCLMs) for both E2E tasks and retrieval tasks. LCLMs performed poorly in the E2E setting, showing subpar results. However, in the retrieval task, LCLMs performed reasonably well, although they still lagged behind embedding-based models, particularly when the number of retrieved documents was increased, highlighting the advantages of embedding-based approaches in this context.

Lastly, we address the limitations of our proposed benchmark, particularly with regard to the ambiguity of queries and the multi-label nature of the queries. We suggest potential methods to mitigate these issues in future work. Additionally, we propose the inclusion of legal reasoning evaluation in future assessments, offering a brief literature review on LLM reasoning evaluation to support this recommendation.

\section{Acknowledgment}
\label{sec: acknowledgement}

We would like to express our sincere gratitude to Supavich Punchun for facilitating WCX-CCL data preparation, Apiwat Sukthawornpradit, Watcharit Boonying, and Tawan Tantakull for scraping, preprocessing, and preparing Tax Case Dataset. We also thank all VISAI.AI company members for helping us perform quality control for LLM-as-a-judge metric validation.

We are deeply thankful to the legal expert annotators for their meticulous work in annotating samples, which was essential for validating the LLM-as-a-judge metrics.

Special thanks to Prof. Keerakiat Pratai - Faculty of Law, Thammasat University - for their insightful consultations on Thai legal information and background knowledge, which significantly enriched our research.

We sincerely thank PTT, SCB, and SCBX, the main sponsors of the WangchanX project, for their generous support. Their contributions have been instrumental in advancing research on Thai legal AI.

Finally, we extend our appreciation to the research assistants at VISTEC for their valuable guidance in constructing benchmarks for LLM systems, particularly in retrieval and end-to-end (E2E) metrics.

\bibliographystyle{plain}  
\bibliography{references}

\appendix

\section{Effect of Chunk Size and Chunk Overlap on Chunking Quality}
\label{appendix: chunk_hyper}

Elaborate from \S\ref{subsubsec: chunking_result}, Table~\ref{table: chunking_by_size} presents the performance of various line chunking strategies across different chunk sizes. As expected, larger chunk sizes correlate with more text per chunk, resulting in higher Sections/Chunk and Chunks/Section values. Larger chunks also reduce the likelihood of incomplete section coverage within a single chunk.

\begin{table}[H]
\centering

\resizebox{\textwidth}{!}{%
\renewcommand{\arraystretch}{1.5} 
\begin{tabular}{@{}lccccc@{}}
\toprule
Chunk Size & \multicolumn{1}{l}{Section/Chunk $\rightarrow$1} & \multicolumn{1}{l}{Chunk/Section $\rightarrow$1} & \multicolumn{1}{l}{Fail Chunk Ratio $\downarrow$} & \multicolumn{1}{l}{Fail Section Ratio $\downarrow$} & \multicolumn{1}{l}{Uncovered Section Ratio $\downarrow$} \\ \midrule

212        & \textbf{1.264}                    & \textbf{1.121}                    & 0.8                                  & 0.551                                  & 0.487                                       \\

250        & \underline{1.373}                       & \underline{1.174}                       & 0.768                                & 0.509                                  & 0.425                                       \\

300        & 1.497                             & 1.227                             & 0.725                                & 0.463                                  & 0.356                                       \\

350        & 1.613                             & 1.264                             & 0.684                                & 0.422                                  & 0.304                                       \\

466        & 1.87                              & 1.284                             & 0.597                                & 0.36                                   & 0.207                                       \\

553        & 2.048                             & 1.285                             & \underline{0.528}                          & \underline{0.315}                            & \underline{0.149}                                 \\

600        & 2.16                              & 1.285                             & \textbf{0.5}                         & \textbf{0.301}                         & \textbf{0.13}                               \\ \bottomrule
\end{tabular}
}
\caption{Line chunking strategy result by chunk size}
\label{table: chunking_by_size}
\end{table}

Table~\ref{table: chunking_by_overlap} presents the evaluation of the line chunking strategy with varying overlaps. As overlap increases, the effective chunk size also increases, leading to higher Sections/Chunk and Chunks/Section values. Furthermore, increasing overlap reduces the Fail Chunk Ratio, Fail Section Ratio, and Uncovered Section Ratio.

\begin{table}[H]
\centering

\resizebox{\textwidth}{!}{%
\renewcommand{\arraystretch}{1.5} 
\small 
\begin{tabular}{@{}lccccc@{}}
\toprule
Chunk Overlap & \multicolumn{1}{l}{Section/Chunk $\rightarrow$1} & \multicolumn{1}{l}{Chunk/Section $\rightarrow$1} & \multicolumn{1}{l}{Fail Chunk Ratio $\downarrow$} & \multicolumn{1}{l}{Fail Section Ratio $\downarrow$} & \multicolumn{1}{l}{Uncovered Section Ratio $\downarrow$} \\ \midrule

50            & \textbf{1.627}                    & \textbf{1.153}                    & \textbf{0.654}                       & 0.423                                  & 0.303                                       \\

100           & \underline{1.666}                       & \underline{1.183}                       & \underline{0.659}                          & 0.422                                  & 0.299                                       \\

150           & 1.717                             & 1.254                             & 0.661                                & \underline{0.417}                            & \underline{0.292}                                 \\

200           & 1.747                             & 1.347                             & 0.657                                & \textbf{0.407}                         & \textbf{0.283}                              \\ \bottomrule
\end{tabular}
}
\caption{Line chunking result by chunk overlap}
\label{table: chunking_by_overlap}
\end{table}

Finally, Table~\ref{table: chunking_map_stats} presents statistics on section coverage within each chunk using the selected line chunking strategy (chunk size: 553 characters, overlap: 50 characters).

\begin{table}[H]
\centering
\begin{tabular}{@{}cccccccc@{}}
\toprule
Count & Mean  & Std.  & Min   & 25th Percentile & 50th Percentile & 75th Percentile & Max \\ \midrule
2610  & 1.331 & 0.577 & 1.000 & 1               & 1               & 2               & 7   \\ \bottomrule
\end{tabular}
\caption{Statistics on section coverage within each chunk}
\label{table: chunking_map_stats}
\end{table}

\end{document}